\documentclass{article}

\PassOptionsToPackage{dvipsnames,x11names}{xcolor}
\PassOptionsToPackage{numbers,sort,compress}{natbib}
\usepackage[main, final]{neurips_2026}

\usepackage{microtype}
\usepackage{graphicx}
\usepackage{subcaption}
\usepackage{booktabs} %
\usepackage{tabularx}
\usepackage{longtable}
\usepackage{array}
\usepackage{xcolor}
\usepackage{colortbl}
\usepackage{arydshln} %
\usepackage{threeparttable}
\usepackage{adjustbox}
\usepackage{afterpage}
\usepackage{float}
\usepackage{enumitem}
\usepackage{nameref}
\usepackage{stmaryrd}

\usepackage{hyperref}
   \hypersetup{colorlinks,breaklinks,allcolors=Blue3}
\usepackage{bookmark}
\usepackage{url}

\usepackage{amsmath}
\usepackage{amssymb}
\usepackage{mathtools}
\usepackage{amsthm}

\usepackage[capitalize,noabbrev]{cleveref}

\theoremstyle{plain}

\theoremstyle{definition}

\theoremstyle{remark}

\usepackage{algorithm}
\usepackage{algorithmic}

\usepackage[textsize=tiny]{todonotes}

\usepackage{listings}
\usepackage{listings-rust}
\usepackage{tcolorbox}
\tcbuselibrary{listings}

\usepackage{soul} %
\usepackage{pifont}
\usepackage{bbding}
\usepackage{xspace}

\newcommand{\methodname}{Decoding Time Verification}
\newcommand{\methodabbr}{DTV}
\newcommand{\method}{\methodname{} (\methodabbr{})}

\newcommand{\numCRustSamples}{300}
\newcommand{\numJSTSSamples}{150}
\newcommand{\budgetK}{16}

\definecolor{codegreen}{rgb}{0,0.6,0}
\definecolor{codegray}{rgb}{0.5,0.5,0.5}
\definecolor{codepurple}{rgb}{0.58,0,0.82}
\definecolor{backcolor}{rgb}{0.95,0.95,0.92}
\definecolor{backcolor1}{HTML}{f6f6f6}
\definecolor{backcolor2}{HTML}{fcf4e2}
\definecolor{backcolor3}{HTML}{eff1f5}

\definecolor{rescuegreen}{HTML}{1B7837}
\newcommand{\heatcell}[3]{\cellcolor{rescuegreen!#1}#2\,{\scriptsize(#3)}}

\lstdefinestyle{mystyle}{
    backgroundcolor=\color{backcolor},
    commentstyle=\color{codegreen},
    keywordstyle=\color{magenta},
    stringstyle=\color{codepurple},
    basicstyle=\ttfamily\small,
    breakatwhitespace=false,
    breaklines=true,
    captionpos=b,
    keepspaces=true,
    numbersep=5pt,
    showspaces=false,
    showstringspaces=false,
    showtabs=false,
    tabsize=2,
    lineskip=-1pt
}
\newenvironment{squishenumerate}
  {\begin{list}{\arabic{enumi}.}{%
    \usecounter{enumi}%
    \setlength{\itemsep}{0pt}%
    \setlength{\parsep}{0pt}%
    \setlength{\topsep}{0pt}%
    \setlength{\parskip}{0pt}%
    \setlength{\labelwidth}{.5in}%
    \setlength{\labelsep}{0.05in}%
    \setlength{\leftmargin}{.2in}}}
  {\end{list}}

\newif\iffeedback
\feedbackfalse

\iffeedback
\newcommand{\varun}[1]{{\color{red}(Varun: #1)}}
\newcommand{\varunfix}[1]{{\color{teal}Varun (Fixed): #1)}}
\newcommand{\mihai}[1]{{\sethlcolor{Chartreuse1}\hl{(mihai: #1)}}}
\newcommand{\mihaifix}[1]{{\color{teal}(mihai (fixed): #1)}}
\newcommand{\tz}[1]{{\color{blue}(tz: #1)}}
\newcommand{\tzreply}[1]{{\color{cyan}(tz reply: #1)}}
\newcommand{\tzfix}[1]{{\color{OliveGreen}(tz fixed: #1)}}

\newcommand{\sj}[1]{{\color{purple}(sj: #1)}}
\newcommand{\sjfix}[1]{{\color{teal}(sj (fixed): #1)}}
\newcommand{\kirill}[1]{\textcolor{orange}{$\langle$Kirill: #1$\rangle$}}
\newcommand{\kirillfix}[1]{\textcolor{teal}{$\langle$Kirill (fixed): #1$\rangle$}}

\else
\newcommand{\varun}[1]{}
\newcommand{\varunfix}[1]{}
\newcommand{\mihai}[1]{}
\newcommand{\mihaifix}[1]{}
\newcommand{\tz}[1]{}
\newcommand{\tzreply}[1]{}
\newcommand{\tzfix}[1]{}

\newcommand{\sj}[1]{}
\newcommand{\sjfix}[1]{}
\newcommand{\kirill}[1]{}
\newcommand{\kirillfix}[1]{}
\fi

\makeatletter
\@ifundefined{Comment}{}{%
  \let\oldComment\Comment
  \renewcommand{\Comment}[1]{\textcolor{gray}{\oldComment{#1}}}
}
\makeatother

\title{Verifiers in the Loop: Decoding Time Verification for Code Translation}

\author{
  Tianyang Zhou \\
  University of Illinois Urbana-Champaign \\
  \texttt{tz64@illinois.edu} \\
  \And
  Somesh Jha \\
  University of Wisconsin--Madison and Google \\
  \texttt{jha@cs.wisc.edu} \\
  \And
  Mihai Christodorescu \\
  Google \\
  \texttt{christodorescu@google.com} \\
  \And
  Kirill Levchenko \\
  University of Illinois Urbana-Champaign \\
  \texttt{klevchen@illinois.edu} \\
  \And
  Varun Chandrasekaran \\
  University of Illinois Urbana-Champaign \\
  \texttt{varunc@illinois.edu}
}

\begin{document}

\maketitle
\suppressfloats[t]  %

\newcommand{\rqOneCrustNaiveOneshotPass}{16.3\%}
\newcommand{\rqOneCrustNaiveOneshotNumPass}{49}
\newcommand{\rqOneCrustNaiveOneshotNumTotal}{300}
\newcommand{\rqOneCrustNaiveOneshotAvgK}{1.55}
\newcommand{\rqOneCrustNaiveOneshotTpp}{360}
\newcommand{\rqOneCrustNaiveOneshotTppCiLo}{309}
\newcommand{\rqOneCrustNaiveOneshotTppCiHi}{415}
\newcommand{\rqOneCrustNaiveSrPass}{72.3\%}
\newcommand{\rqOneCrustNaiveSrNumPass}{217}
\newcommand{\rqOneCrustNaiveSrNumTotal}{300}
\newcommand{\rqOneCrustNaiveSrAvgK}{7.18}
\newcommand{\rqOneCrustNaiveSrTpp}{1903}
\newcommand{\rqOneCrustNaiveSrTppCiLo}{1548}
\newcommand{\rqOneCrustNaiveSrTppCiHi}{2339}
\newcommand{\rqOneCrustDtvOneshotPass}{47.0\%}
\newcommand{\rqOneCrustDtvOneshotNumPass}{141}
\newcommand{\rqOneCrustDtvOneshotNumTotal}{300}
\newcommand{\rqOneCrustDtvOneshotAvgK}{2.00}
\newcommand{\rqOneCrustDtvOneshotTpp}{596}
\newcommand{\rqOneCrustDtvOneshotTppCiLo}{499}
\newcommand{\rqOneCrustDtvOneshotTppCiHi}{714}
\newcommand{\rqOneCrustDtvSrPass}{82.0\%}
\newcommand{\rqOneCrustDtvSrNumPass}{246}
\newcommand{\rqOneCrustDtvSrNumTotal}{300}
\newcommand{\rqOneCrustDtvSrAvgK}{5.28}
\newcommand{\rqOneCrustDtvSrTpp}{1654}
\newcommand{\rqOneCrustDtvSrTppCiLo}{1345}
\newcommand{\rqOneCrustDtvSrTppCiHi}{1990}

\newcommand{\rqOneJstsNaiveOneshotPass}{4.7\%}
\newcommand{\rqOneJstsNaiveOneshotNumPass}{7}
\newcommand{\rqOneJstsNaiveOneshotNumTotal}{150}
\newcommand{\rqOneJstsNaiveOneshotAvgK}{1.10}
\newcommand{\rqOneJstsNaiveOneshotTpp}{435}
\newcommand{\rqOneJstsNaiveOneshotTppCiLo}{310}
\newcommand{\rqOneJstsNaiveOneshotTppCiHi}{567}
\newcommand{\rqOneJstsNaiveSrPass}{33.3\%}
\newcommand{\rqOneJstsNaiveSrNumPass}{50}
\newcommand{\rqOneJstsNaiveSrNumTotal}{150}
\newcommand{\rqOneJstsNaiveSrAvgK}{11.48}
\newcommand{\rqOneJstsNaiveSrTpp}{1648}
\newcommand{\rqOneJstsNaiveSrTppCiLo}{1071}
\newcommand{\rqOneJstsNaiveSrTppCiHi}{2497}
\newcommand{\rqOneJstsDtvOneshotPass}{21.3\%}
\newcommand{\rqOneJstsDtvOneshotNumPass}{32}
\newcommand{\rqOneJstsDtvOneshotNumTotal}{150}
\newcommand{\rqOneJstsDtvOneshotAvgK}{2.77}
\newcommand{\rqOneJstsDtvOneshotTpp}{715}
\newcommand{\rqOneJstsDtvOneshotTppCiLo}{547}
\newcommand{\rqOneJstsDtvOneshotTppCiHi}{960}
\newcommand{\rqOneJstsDtvSrPass}{46.0\%}
\newcommand{\rqOneJstsDtvSrNumPass}{69}
\newcommand{\rqOneJstsDtvSrNumTotal}{150}
\newcommand{\rqOneJstsDtvSrAvgK}{10.11}
\newcommand{\rqOneJstsDtvSrTpp}{2103}
\newcommand{\rqOneJstsDtvSrTppCiLo}{1470}
\newcommand{\rqOneJstsDtvSrTppCiHi}{2856}

\newcommand{\rqOneCrustNaiveOneshotVsDtvOneshotDiffPp}{+30.7}
\newcommand{\rqOneCrustNaiveOneshotVsDtvOneshotMcnemarP}{<0.001}
\newcommand{\rqOneCrustNaiveOneshotVsDtvOneshotOnlyA}{13}
\newcommand{\rqOneCrustNaiveOneshotVsDtvOneshotOnlyB}{105}
\newcommand{\rqOneCrustNaiveOneshotVsDtvOneshotN}{300}
\newcommand{\rqOneCrustNaiveSrVsDtvSrDiffPp}{+9.7}
\newcommand{\rqOneCrustNaiveSrVsDtvSrMcnemarP}{<0.001}
\newcommand{\rqOneCrustNaiveSrVsDtvSrOnlyA}{21}
\newcommand{\rqOneCrustNaiveSrVsDtvSrOnlyB}{50}
\newcommand{\rqOneCrustNaiveSrVsDtvSrN}{300}

\newcommand{\rqOneJstsNaiveOneshotVsDtvOneshotDiffPp}{+16.7}
\newcommand{\rqOneJstsNaiveOneshotVsDtvOneshotMcnemarP}{<0.001}
\newcommand{\rqOneJstsNaiveOneshotVsDtvOneshotOnlyA}{1}
\newcommand{\rqOneJstsNaiveOneshotVsDtvOneshotOnlyB}{26}
\newcommand{\rqOneJstsNaiveOneshotVsDtvOneshotN}{150}
\newcommand{\rqOneJstsNaiveSrVsDtvSrDiffPp}{+12.7}
\newcommand{\rqOneJstsNaiveSrVsDtvSrMcnemarP}{=0.001}
\newcommand{\rqOneJstsNaiveSrVsDtvSrOnlyA}{7}
\newcommand{\rqOneJstsNaiveSrVsDtvSrOnlyB}{26}
\newcommand{\rqOneJstsNaiveSrVsDtvSrN}{150}

\newcommand{\rqOneCrustNaiveSrFuncFullPass}{9.3\%}
\newcommand{\rqOneCrustNaiveSrFuncFullNum}{28}
\newcommand{\rqOneCrustNaiveSrFuncNormMean}{0.239}
\newcommand{\rqOneCrustNaiveSrFuncNumCompile}{216}
\newcommand{\rqOneCrustNaiveSrFuncCondFullPass}{13.0\%}
\newcommand{\rqOneCrustNaiveSrFuncCondAvgRate}{33.0\%}
\newcommand{\rqOneCrustDtvSrFuncFullPass}{8.3\%}
\newcommand{\rqOneCrustDtvSrFuncFullNum}{25}
\newcommand{\rqOneCrustDtvSrFuncNormMean}{0.224}
\newcommand{\rqOneCrustDtvSrFuncNumCompile}{245}
\newcommand{\rqOneCrustDtvSrFuncCondFullPass}{10.2\%}
\newcommand{\rqOneCrustDtvSrFuncCondAvgRate}{27.3\%}

\newcommand{\rqOneCrustNaiveSrVsDtvSrFuncDiffPp}{-1.00}
\newcommand{\rqOneCrustNaiveSrVsDtvSrFuncMcnemarP}{=0.648}
\newcommand{\rqOneCrustNaiveSrVsDtvSrFuncOnlyA}{11}
\newcommand{\rqOneCrustNaiveSrVsDtvSrFuncOnlyB}{8}
\newcommand{\rqOneCrustNaiveSrVsDtvSrFuncNormDiff}{-0.015}
\newcommand{\rqOneCrustNaiveSrVsDtvSrFuncNormDiffCiLo}{-0.052}
\newcommand{\rqOneCrustNaiveSrVsDtvSrFuncNormDiffCiHi}{+0.023}
\newcommand{\rqOneCrustNaiveSrVsDtvSrFuncBcN}{195}
\newcommand{\rqOneCrustNaiveSrVsDtvSrFuncBcAFullRate}{12.3\%}
\newcommand{\rqOneCrustNaiveSrVsDtvSrFuncBcBFullRate}{10.8\%}
\newcommand{\rqOneCrustNaiveSrVsDtvSrFuncBcFullDiffPp}{-1.54}
\newcommand{\rqOneCrustNaiveSrVsDtvSrFuncBcFullMcnemarP}{=0.549}
\newcommand{\rqOneCrustNaiveSrVsDtvSrFuncBcFullOnlyA}{7}
\newcommand{\rqOneCrustNaiveSrVsDtvSrFuncBcFullOnlyB}{4}
\newcommand{\rqOneCrustNaiveSrVsDtvSrFuncBcAAvgRate}{32.8\%}
\newcommand{\rqOneCrustNaiveSrVsDtvSrFuncBcBAvgRate}{29.6\%}
\newcommand{\rqOneCrustNaiveSrVsDtvSrFuncBcAvgDiff}{-3.18}
\newcommand{\rqOneCrustNaiveSrVsDtvSrFuncBcAvgDiffCiLo}{-7.49}
\newcommand{\rqOneCrustNaiveSrVsDtvSrFuncBcAvgDiffCiHi}{+1.14}

\newcommand{\rqOneQwenFourBCrustTimeRatio}{0.79}
\newcommand{\rqOneQwenFourBJstsTimeRatio}{1.85}
\newcommand{\rqOneQwenThirtyBCrustTimeRatio}{0.57}
\newcommand{\rqOneQwenThirtyBJstsTimeRatio}{1.26}
\newcommand{\rqOneRustVerifierLatency}{0.054}
\newcommand{\rqOneTypeScriptVerifierLatency}{1.823}
\newcommand{\rqOneParallelTypeScriptLatencyEstimate}{1.253}

\newcommand{\rqOneMainResultsTable}{%
\begin{tabular}{llrrrr}
\toprule
\textbf{Task} & \textbf{Configuration} & \textbf{Pass} & \textbf{Pass rate} & \textbf{Avg.\ $k$} & \textbf{Tokens / pass [95\% CI]} \\
\midrule
C $\to$ Rust ($n=$300) & naive/one-shot & 49/300 & 16.3\% & 1.55 & 360 [309,415] \\
 & naive/self-refine & 217/300 & 72.3\% & 7.18 & 1903 [1548,2339] \\
\noalign{\vskip 1.5pt}\cdashline{2-6}\noalign{\vskip 2.5pt}
 & \methodabbr{}/one-shot & 141/300 & 47.0\% & 2.00 & 596 [499,714] \\
 & \methodabbr{}/self-refine & 246/300 & 82.0\% & 5.28 & 1654 [1345,1990] \\
\midrule
JS $\to$ TS ($n=$150) & naive/one-shot & 7/150 & 4.7\% & 1.10 & 435 [310,567] \\
 & naive/self-refine & 50/150 & 33.3\% & 11.48 & 1648 [1071,2497] \\
\noalign{\vskip 1.5pt}\cdashline{2-6}\noalign{\vskip 2.5pt}
 & \methodabbr{}/one-shot & 32/150 & 21.3\% & 2.77 & 715 [547,960] \\
 & \methodabbr{}/self-refine & 69/150 & 46.0\% & 10.11 & 2103 [1470,2856] \\
\bottomrule
\end{tabular}
}

\newcommand{\rqOnePairTable}{%
\begin{tabular}{lllrrrrr}
\toprule
\textbf{Task} & \textbf{Baseline (A)} & \textbf{\methodabbr{} (B)} & \textbf{$n$} & \textbf{Only A} & \textbf{Only B} & \textbf{Diff (pp)} & \textbf{McNemar $p$} \\
\midrule
C $\to$ Rust & naive/one-shot & \methodabbr{}/one-shot & 300 & 13 & 105 & +30.7 & $<0.001$ \\
 & naive/self-refine & \methodabbr{}/self-refine & 300 & 21 & 50 & +9.7 & $<0.001$ \\
\midrule
JS $\to$ TS & naive/one-shot & \methodabbr{}/one-shot & 150 & 1 & 26 & +16.7 & $<0.001$ \\
 & naive/self-refine & \methodabbr{}/self-refine & 150 & 7 & 26 & +12.7 & $0.001$ \\
\bottomrule
\end{tabular}
}

\newcommand{\rqOneFunctionalTable}{%
\begin{tabular}{lrrrrr}
\toprule
\textbf{Configuration} & \textbf{Compile} & \textbf{Full success} & \textbf{Norm.\ score} & \textbf{Full $\mid$ compile} & \textbf{Avg.\ test rate $\mid$ compile} \\
\midrule
naive/self-refine & 217/300 (72.3\%) & 28/300 (9.33\%) & 0.239 & 28/216 (12.96\%) & 32.99\% \\
\methodabbr{}/self-refine & 246/300 (82.0\%) & 25/300 (8.33\%) & 0.224 & 25/245 (10.20\%) & 27.27\% \\
\bottomrule
\end{tabular}
}

\newcommand{\rqOneFunctionalPairTable}{%
\begin{tabular}{llrrrrrl}
\toprule
\textbf{A} & \textbf{B} & \textbf{$n$} & \textbf{Only A} & \textbf{Only B} & \textbf{Diff (pp)} & \textbf{McNemar $p$} & \textbf{Norm.\ diff [95\% CI]} \\
\midrule
naive/self-refine & \methodabbr{}/self-refine & 300 & 11 & 8 & -1.00 & $0.648$ & -0.015 [-0.052,+0.023] \\
\bottomrule
\end{tabular}
}

\newcommand{\rqOneFunctionalBothCompileTable}{%
\begin{tabular}{llrrrrrrl}
\toprule
\textbf{A} & \textbf{B} & \textbf{$n$} & \textbf{A full} & \textbf{B full} & \textbf{Full diff (pp)} & \textbf{McNemar $p$} & \textbf{Avg.\ rate diff} & \textbf{[95\% CI]} \\
\midrule
naive/self-refine & \methodabbr{}/self-refine & 195 & 12.31\% & 10.77\% & -1.54 & $0.549$ & -3.18pp & [-7.49,+1.14] \\
\bottomrule
\end{tabular}
}

\newcommand{\rqOneRuntimeTable}{%
\begin{tabular}{llrrrrrr}
\toprule
Model & Task & $n$ & \multicolumn{2}{c}{Pass rate (\%)} & \multicolumn{2}{c}{Seconds/case} & Time ratio \\
 & & & na\"ive & \methodabbr{} & na\"ive & \methodabbr{} & DTV/na\"ive \\
\midrule
Qwen3-4B & C-to-Rust & 300 & 72.3\% & 82.0\% & 245.1 & 192.8 & 0.79 \\
Qwen3-4B & JS-to-TS & 150 & 33.3\% & 46.0\% & 588.7 & 1,091.4 & 1.85 \\
Qwen3-30B-A3B & C-to-Rust & 100 & 92\% & 97\% & 173.3 & 98.1 & 0.57 \\
Qwen3-30B-A3B & JS-to-TS & 100 & 57\% & 66\% & 807.9 & 1,017.9 & 1.26 \\
\bottomrule
\end{tabular}%
}

\newcommand{\rqOneVerifierCallsTable}{%
\begin{tabular}{llrrr}
\toprule
Model & Task & \multicolumn{2}{c}{Outer rounds/case} & DTV inner calls/case \\
 & & na\"ive & \methodabbr{} & \\
\midrule
Qwen3-4B & C-to-Rust & 5.49 & 3.08 & 208.5 \\
Qwen3-4B & JS-to-TS & 9.27 & 8.23 & 259.6 \\
Qwen3-30B-A3B & C-to-Rust & 2.34 & 1.34 & 82.11 \\
Qwen3-30B-A3B & JS-to-TS & 6.82 & 5.02 & 363.83 \\
\bottomrule
\end{tabular}%
}

\newcommand{\rqOneCrustNaiveOneshotVsDtvOneshotBpN}{36}
\newcommand{\rqOneCrustNaiveOneshotVsDtvOneshotBpDtvBetter}{11}
\newcommand{\rqOneCrustNaiveOneshotVsDtvOneshotBpDtvBetterPct}{30.6\%}
\newcommand{\rqOneCrustNaiveOneshotVsDtvOneshotBpNaiveBetter}{18}
\newcommand{\rqOneCrustNaiveOneshotVsDtvOneshotBpNaiveBetterPct}{50.0\%}
\newcommand{\rqOneCrustNaiveOneshotVsDtvOneshotBpTied}{7}
\newcommand{\rqOneCrustNaiveOneshotVsDtvOneshotBpTiedPct}{19.4\%}
\newcommand{\rqOneCrustNaiveOneshotVsDtvOneshotBpMedianK}{+0.00}
\newcommand{\rqOneCrustNaiveOneshotVsDtvOneshotBpMeanK}{+0.12}
\newcommand{\rqOneCrustNaiveOneshotVsDtvOneshotBpMedianTok}{+0}
\newcommand{\rqOneCrustNaiveOneshotVsDtvOneshotBpMeanTok}{+34}
\newcommand{\rqOneCrustNaiveOneshotVsDtvOneshotBpSignP}{=0.265}
\newcommand{\rqOneCrustNaiveOneshotVsDtvOneshotBpTailTopN}{3}
\newcommand{\rqOneCrustNaiveOneshotVsDtvOneshotBpTailDtvGoodSum}{-145}
\newcommand{\rqOneCrustNaiveOneshotVsDtvOneshotBpTailDtvBadSum}{+924}
\newcommand{\rqOneCrustNaiveOneshotVsDtvOneshotBpTailRatio}{6.37}
\newcommand{\rqOneCrustNaiveOneshotVsDtvOneshotBpTrimN}{30}
\newcommand{\rqOneCrustNaiveOneshotVsDtvOneshotBpTrimMeanTok}{+14}
\newcommand{\rqOneCrustNaiveSrVsDtvSrBpN}{196}
\newcommand{\rqOneCrustNaiveSrVsDtvSrBpDtvBetter}{133}
\newcommand{\rqOneCrustNaiveSrVsDtvSrBpDtvBetterPct}{67.9\%}
\newcommand{\rqOneCrustNaiveSrVsDtvSrBpNaiveBetter}{56}
\newcommand{\rqOneCrustNaiveSrVsDtvSrBpNaiveBetterPct}{28.6\%}
\newcommand{\rqOneCrustNaiveSrVsDtvSrBpTied}{7}
\newcommand{\rqOneCrustNaiveSrVsDtvSrBpTiedPct}{3.6\%}
\newcommand{\rqOneCrustNaiveSrVsDtvSrBpMedianK}{-0.83}
\newcommand{\rqOneCrustNaiveSrVsDtvSrBpMeanK}{-0.82}
\newcommand{\rqOneCrustNaiveSrVsDtvSrBpMedianTok}{-208}
\newcommand{\rqOneCrustNaiveSrVsDtvSrBpMeanTok}{-519}
\newcommand{\rqOneCrustNaiveSrVsDtvSrBpSignP}{<0.001}
\newcommand{\rqOneCrustNaiveSrVsDtvSrBpTailTopN}{19}
\newcommand{\rqOneCrustNaiveSrVsDtvSrBpTailDtvGoodSum}{-98903}
\newcommand{\rqOneCrustNaiveSrVsDtvSrBpTailDtvBadSum}{+44106}
\newcommand{\rqOneCrustNaiveSrVsDtvSrBpTailRatio}{0.45}
\newcommand{\rqOneCrustNaiveSrVsDtvSrBpTrimN}{158}
\newcommand{\rqOneCrustNaiveSrVsDtvSrBpTrimMeanTok}{-297}

\newcommand{\rqOneJstsNaiveOneshotVsDtvOneshotBpN}{6}
\newcommand{\rqOneJstsNaiveOneshotVsDtvOneshotBpDtvBetter}{1}
\newcommand{\rqOneJstsNaiveOneshotVsDtvOneshotBpDtvBetterPct}{16.7\%}
\newcommand{\rqOneJstsNaiveOneshotVsDtvOneshotBpNaiveBetter}{0}
\newcommand{\rqOneJstsNaiveOneshotVsDtvOneshotBpNaiveBetterPct}{0.0\%}
\newcommand{\rqOneJstsNaiveOneshotVsDtvOneshotBpTied}{5}
\newcommand{\rqOneJstsNaiveOneshotVsDtvOneshotBpTiedPct}{83.3\%}
\newcommand{\rqOneJstsNaiveOneshotVsDtvOneshotBpMedianK}{+0.00}
\newcommand{\rqOneJstsNaiveOneshotVsDtvOneshotBpMeanK}{-0.00}
\newcommand{\rqOneJstsNaiveOneshotVsDtvOneshotBpMedianTok}{+0}
\newcommand{\rqOneJstsNaiveOneshotVsDtvOneshotBpMeanTok}{+0}
\newcommand{\rqOneJstsNaiveOneshotVsDtvOneshotBpSignP}{=1.000}
\newcommand{\rqOneJstsNaiveOneshotVsDtvOneshotBpTailTopN}{1}
\newcommand{\rqOneJstsNaiveOneshotVsDtvOneshotBpTailDtvGoodSum}{-3}
\newcommand{\rqOneJstsNaiveOneshotVsDtvOneshotBpTailDtvBadSum}{+0}
\newcommand{\rqOneJstsNaiveOneshotVsDtvOneshotBpTailRatio}{0.00}
\newcommand{\rqOneJstsNaiveOneshotVsDtvOneshotBpTrimN}{4}
\newcommand{\rqOneJstsNaiveOneshotVsDtvOneshotBpTrimMeanTok}{+0}
\newcommand{\rqOneJstsNaiveSrVsDtvSrBpN}{43}
\newcommand{\rqOneJstsNaiveSrVsDtvSrBpDtvBetter}{24}
\newcommand{\rqOneJstsNaiveSrVsDtvSrBpDtvBetterPct}{55.8\%}
\newcommand{\rqOneJstsNaiveSrVsDtvSrBpNaiveBetter}{14}
\newcommand{\rqOneJstsNaiveSrVsDtvSrBpNaiveBetterPct}{32.6\%}
\newcommand{\rqOneJstsNaiveSrVsDtvSrBpTied}{5}
\newcommand{\rqOneJstsNaiveSrVsDtvSrBpTiedPct}{11.6\%}
\newcommand{\rqOneJstsNaiveSrVsDtvSrBpMedianK}{-0.27}
\newcommand{\rqOneJstsNaiveSrVsDtvSrBpMeanK}{+0.45}
\newcommand{\rqOneJstsNaiveSrVsDtvSrBpMedianTok}{-61}
\newcommand{\rqOneJstsNaiveSrVsDtvSrBpMeanTok}{+220}
\newcommand{\rqOneJstsNaiveSrVsDtvSrBpSignP}{=0.143}
\newcommand{\rqOneJstsNaiveSrVsDtvSrBpTailTopN}{4}
\newcommand{\rqOneJstsNaiveSrVsDtvSrBpTailDtvGoodSum}{-7604}
\newcommand{\rqOneJstsNaiveSrVsDtvSrBpTailDtvBadSum}{+20875}
\newcommand{\rqOneJstsNaiveSrVsDtvSrBpTailRatio}{2.75}
\newcommand{\rqOneJstsNaiveSrVsDtvSrBpTrimN}{35}
\newcommand{\rqOneJstsNaiveSrVsDtvSrBpTrimMeanTok}{-109}

\newcommand{\rqOneBothPassTokenTable}{%
\begin{tabular}{lrrrrrrr}
\toprule
\textbf{Task} & $n$ & \textbf{\methodabbr{} $<$ naive} & \textbf{\methodabbr{} $=$ naive} & \textbf{\methodabbr{} $>$ naive} & \textbf{Median $\Delta$tok} & \textbf{Mean $\Delta$tok} & \textbf{Trimmed mean} \\
\midrule
C $\to$ Rust & 196 & 133 (68\%) & 7 (4\%) & 56 (29\%) & -208 & -519 & -297 \\
JS $\to$ TS & 43 & 24 (56\%) & 5 (12\%) & 14 (33\%) & -61 & +220 & -109 \\
\bottomrule
\end{tabular}
}

\newcommand{\rqBaselineCrustNaiveSrPass}{76.0\%}
\newcommand{\rqBaselineCrustNaiveSrNumPass}{152}
\newcommand{\rqBaselineCrustNaiveSrTokPerCase}{5266}
\newcommand{\rqBaselineCrustNaiveSrTokPerPass}{1798}
\newcommand{\rqBaselineCrustDtvSrPass}{84.5\%}
\newcommand{\rqBaselineCrustDtvSrNumPass}{169}
\newcommand{\rqBaselineCrustDtvSrTokPerCase}{3489}
\newcommand{\rqBaselineCrustDtvSrTokPerPass}{1805}
\newcommand{\rqBaselineCrustNaiveBonOnePass}{15.5\%}
\newcommand{\rqBaselineCrustNaiveBonOneNumPass}{31}
\newcommand{\rqBaselineCrustNaiveBonOneTokPerCase}{737}
\newcommand{\rqBaselineCrustNaiveBonOneTokPerPass}{377}
\newcommand{\rqBaselineCrustNaiveBonTwoPass}{21.0\%}
\newcommand{\rqBaselineCrustNaiveBonTwoNumPass}{42}
\newcommand{\rqBaselineCrustNaiveBonTwoTokPerCase}{1437}
\newcommand{\rqBaselineCrustNaiveBonTwoTokPerPass}{476}
\newcommand{\rqBaselineCrustNaiveBonFourPass}{29.0\%}
\newcommand{\rqBaselineCrustNaiveBonFourNumPass}{58}
\newcommand{\rqBaselineCrustNaiveBonFourTokPerCase}{2772}
\newcommand{\rqBaselineCrustNaiveBonFourTokPerPass}{928}
\newcommand{\rqBaselineCrustNaiveBonEightPass}{36.5\%}
\newcommand{\rqBaselineCrustNaiveBonEightNumPass}{73}
\newcommand{\rqBaselineCrustNaiveBonEightTokPerCase}{5154}
\newcommand{\rqBaselineCrustNaiveBonEightTokPerPass}{1880}
\newcommand{\rqBaselineCrustNaiveBonSixteenPass}{42.0\%}
\newcommand{\rqBaselineCrustNaiveBonSixteenNumPass}{84}
\newcommand{\rqBaselineCrustNaiveBonSixteenTokPerCase}{9574}
\newcommand{\rqBaselineCrustNaiveBonSixteenTokPerPass}{2573}
\newcommand{\rqBaselineCrustNaiveBonThirtyTwoPass}{44.5\%}
\newcommand{\rqBaselineCrustNaiveBonThirtyTwoNumPass}{89}
\newcommand{\rqBaselineCrustNaiveBonThirtyTwoTokPerCase}{17866}
\newcommand{\rqBaselineCrustNaiveBonThirtyTwoTokPerPass}{3445}
\newcommand{\rqBaselineCrustSstarNERrPass}{75.5\%}
\newcommand{\rqBaselineCrustSstarNERrNumPass}{151}
\newcommand{\rqBaselineCrustSstarNERrTokPerCase}{16563}
\newcommand{\rqBaselineCrustSstarNERrTokPerPass}{11551}

\newcommand{\rqBaselineJstsNaiveSrPass}{37.0\%}
\newcommand{\rqBaselineJstsNaiveSrNumPass}{37}
\newcommand{\rqBaselineJstsNaiveSrTokPerCase}{15105}
\newcommand{\rqBaselineJstsNaiveSrTokPerPass}{1740}
\newcommand{\rqBaselineJstsDtvSrPass}{50.0\%}
\newcommand{\rqBaselineJstsDtvSrNumPass}{50}
\newcommand{\rqBaselineJstsDtvSrTokPerCase}{13815}
\newcommand{\rqBaselineJstsDtvSrTokPerPass}{2185}
\newcommand{\rqBaselineJstsNaiveBonOnePass}{7.0\%}
\newcommand{\rqBaselineJstsNaiveBonOneNumPass}{7}
\newcommand{\rqBaselineJstsNaiveBonOneTokPerCase}{1186}
\newcommand{\rqBaselineJstsNaiveBonOneTokPerPass}{435}
\newcommand{\rqBaselineJstsNaiveBonTwoPass}{7.0\%}
\newcommand{\rqBaselineJstsNaiveBonTwoNumPass}{7}
\newcommand{\rqBaselineJstsNaiveBonTwoTokPerCase}{2377}
\newcommand{\rqBaselineJstsNaiveBonTwoTokPerPass}{435}
\newcommand{\rqBaselineJstsNaiveBonFourPass}{7.0\%}
\newcommand{\rqBaselineJstsNaiveBonFourNumPass}{7}
\newcommand{\rqBaselineJstsNaiveBonFourTokPerCase}{4727}
\newcommand{\rqBaselineJstsNaiveBonFourTokPerPass}{435}
\newcommand{\rqBaselineJstsNaiveBonEightPass}{7.0\%}
\newcommand{\rqBaselineJstsNaiveBonEightNumPass}{7}
\newcommand{\rqBaselineJstsNaiveBonEightTokPerCase}{9509}
\newcommand{\rqBaselineJstsNaiveBonEightTokPerPass}{435}
\newcommand{\rqBaselineJstsNaiveBonSixteenPass}{7.0\%}
\newcommand{\rqBaselineJstsNaiveBonSixteenNumPass}{7}
\newcommand{\rqBaselineJstsNaiveBonSixteenTokPerCase}{18871}
\newcommand{\rqBaselineJstsNaiveBonSixteenTokPerPass}{435}
\newcommand{\rqBaselineJstsNaiveBonThirtyTwoPass}{7.0\%}
\newcommand{\rqBaselineJstsNaiveBonThirtyTwoNumPass}{7}
\newcommand{\rqBaselineJstsNaiveBonThirtyTwoTokPerCase}{37664}
\newcommand{\rqBaselineJstsNaiveBonThirtyTwoTokPerPass}{435}
\newcommand{\rqBaselineJstsSstarNERrPass}{33.0\%}
\newcommand{\rqBaselineJstsSstarNERrNumPass}{33}
\newcommand{\rqBaselineJstsSstarNERrTokPerCase}{27388}
\newcommand{\rqBaselineJstsSstarNERrTokPerPass}{9751}

\newcommand{\rqBaselineCrustSstarNERrVsDtvSrDiffPp}{+9.0}
\newcommand{\rqBaselineCrustSstarNERrVsDtvSrMcnemarP}{=0.008}
\newcommand{\rqBaselineCrustNaiveBonThirtyTwoVsDtvSrDiffPp}{+40.0}
\newcommand{\rqBaselineCrustNaiveBonThirtyTwoVsDtvSrMcnemarP}{<0.001}
\newcommand{\rqBaselineJstsSstarNERrVsDtvSrDiffPp}{+17.0}
\newcommand{\rqBaselineJstsSstarNERrVsDtvSrMcnemarP}{<0.001}
\newcommand{\rqBaselineJstsNaiveBonThirtyTwoVsDtvSrDiffPp}{+43.0}
\newcommand{\rqBaselineJstsNaiveBonThirtyTwoVsDtvSrMcnemarP}{<0.001}

\newcommand{\rqTwoQwenName}{Qwen3}
\newcommand{\rqTwoGemmaName}{Gemma4}

\newcommand{\rqTwoQwenCrustNaivePass}{76.0\%}
\newcommand{\rqTwoQwenCrustNaiveNumPass}{152}
\newcommand{\rqTwoQwenCrustNaiveNumTotal}{200}
\newcommand{\rqTwoQwenCrustNaiveAvgK}{6.77}
\newcommand{\rqTwoQwenCrustNaiveTpp}{1798}
\newcommand{\rqTwoQwenCrustDtvPass}{84.5\%}
\newcommand{\rqTwoQwenCrustDtvNumPass}{169}
\newcommand{\rqTwoQwenCrustDtvNumTotal}{200}
\newcommand{\rqTwoQwenCrustDtvAvgK}{5.17}
\newcommand{\rqTwoQwenCrustDtvTpp}{1805}
\newcommand{\rqTwoQwenCrustGain}{+8.5~pp}
\newcommand{\rqTwoQwenCrustGainAbs}{8.5}
\newcommand{\rqTwoQwenCrustKDelta}{-23.6\%}
\newcommand{\rqTwoQwenCrustKDeltaAbs}{23.6\%}
\newcommand{\rqTwoQwenCrustMcnemarP}{=0.019}
\newcommand{\rqTwoQwenCrustMcnemarOnlyNaive}{15}
\newcommand{\rqTwoQwenCrustMcnemarOnlyDtv}{32}
\newcommand{\rqTwoQwenCrustMcnemarN}{200}
\newcommand{\rqTwoQwenCrustPairedN}{137}
\newcommand{\rqTwoQwenCrustPairedNaiveK}{3.54}
\newcommand{\rqTwoQwenCrustPairedDtvK}{2.95}
\newcommand{\rqTwoQwenCrustPairedKDelta}{-16.5\%}

\newcommand{\rqTwoGemmaCrustNaivePass}{93.5\%}
\newcommand{\rqTwoGemmaCrustNaiveNumPass}{187}
\newcommand{\rqTwoGemmaCrustNaiveNumTotal}{200}
\newcommand{\rqTwoGemmaCrustNaiveAvgK}{5.91}
\newcommand{\rqTwoGemmaCrustNaiveTpp}{3327}
\newcommand{\rqTwoGemmaCrustDtvPass}{86.0\%}
\newcommand{\rqTwoGemmaCrustDtvNumPass}{172}
\newcommand{\rqTwoGemmaCrustDtvNumTotal}{200}
\newcommand{\rqTwoGemmaCrustDtvAvgK}{5.99}
\newcommand{\rqTwoGemmaCrustDtvTpp}{3024}
\newcommand{\rqTwoGemmaCrustGain}{-7.5~pp}
\newcommand{\rqTwoGemmaCrustGainAbs}{7.5}
\newcommand{\rqTwoGemmaCrustKDelta}{+1.2\%}
\newcommand{\rqTwoGemmaCrustKDeltaAbs}{1.2\%}
\newcommand{\rqTwoGemmaCrustMcnemarP}{=0.003}
\newcommand{\rqTwoGemmaCrustMcnemarOnlyNaive}{19}
\newcommand{\rqTwoGemmaCrustMcnemarOnlyDtv}{4}
\newcommand{\rqTwoGemmaCrustMcnemarN}{200}
\newcommand{\rqTwoGemmaCrustPairedN}{168}
\newcommand{\rqTwoGemmaCrustPairedNaiveK}{5.10}
\newcommand{\rqTwoGemmaCrustPairedDtvK}{4.66}
\newcommand{\rqTwoGemmaCrustPairedKDelta}{-8.7\%}

\newcommand{\rqTwoQwenJstsNaivePass}{37.0\%}
\newcommand{\rqTwoQwenJstsNaiveNumPass}{37}
\newcommand{\rqTwoQwenJstsNaiveNumTotal}{100}
\newcommand{\rqTwoQwenJstsNaiveAvgK}{10.96}
\newcommand{\rqTwoQwenJstsNaiveTpp}{1740}
\newcommand{\rqTwoQwenJstsDtvPass}{50.0\%}
\newcommand{\rqTwoQwenJstsDtvNumPass}{50}
\newcommand{\rqTwoQwenJstsDtvNumTotal}{100}
\newcommand{\rqTwoQwenJstsDtvAvgK}{9.70}
\newcommand{\rqTwoQwenJstsDtvTpp}{2185}
\newcommand{\rqTwoQwenJstsGain}{+13.0~pp}
\newcommand{\rqTwoQwenJstsGainAbs}{13.0}
\newcommand{\rqTwoQwenJstsKDelta}{-11.5\%}
\newcommand{\rqTwoQwenJstsKDeltaAbs}{11.5\%}
\newcommand{\rqTwoQwenJstsMcnemarP}{=0.015}
\newcommand{\rqTwoQwenJstsMcnemarOnlyNaive}{6}
\newcommand{\rqTwoQwenJstsMcnemarOnlyDtv}{19}
\newcommand{\rqTwoQwenJstsMcnemarN}{100}
\newcommand{\rqTwoQwenJstsPairedN}{31}
\newcommand{\rqTwoQwenJstsPairedNaiveK}{2.54}
\newcommand{\rqTwoQwenJstsPairedDtvK}{3.39}
\newcommand{\rqTwoQwenJstsPairedKDelta}{+33.3\%}

\newcommand{\rqTwoGemmaJstsNaivePass}{50.0\%}
\newcommand{\rqTwoGemmaJstsNaiveNumPass}{50}
\newcommand{\rqTwoGemmaJstsNaiveNumTotal}{100}
\newcommand{\rqTwoGemmaJstsNaiveAvgK}{9.87}
\newcommand{\rqTwoGemmaJstsNaiveTpp}{3515}
\newcommand{\rqTwoGemmaJstsDtvPass}{58.0\%}
\newcommand{\rqTwoGemmaJstsDtvNumPass}{58}
\newcommand{\rqTwoGemmaJstsDtvNumTotal}{100}
\newcommand{\rqTwoGemmaJstsDtvAvgK}{8.43}
\newcommand{\rqTwoGemmaJstsDtvTpp}{3692}
\newcommand{\rqTwoGemmaJstsGain}{+8.0~pp}
\newcommand{\rqTwoGemmaJstsGainAbs}{8.0}
\newcommand{\rqTwoGemmaJstsKDelta}{-14.5\%}
\newcommand{\rqTwoGemmaJstsKDeltaAbs}{14.5\%}
\newcommand{\rqTwoGemmaJstsMcnemarP}{=0.200}
\newcommand{\rqTwoGemmaJstsMcnemarOnlyNaive}{11}
\newcommand{\rqTwoGemmaJstsMcnemarOnlyDtv}{19}
\newcommand{\rqTwoGemmaJstsMcnemarN}{100}
\newcommand{\rqTwoGemmaJstsPairedN}{39}
\newcommand{\rqTwoGemmaJstsPairedNaiveK}{3.01}
\newcommand{\rqTwoGemmaJstsPairedDtvK}{2.49}
\newcommand{\rqTwoGemmaJstsPairedKDelta}{-17.2\%}

\newcommand{\rqTwoQwenCrustNaiveFuncFull}{11.50\%}
\newcommand{\rqTwoQwenCrustDtvFuncFull}{9.50\%}
\newcommand{\rqTwoQwenCrustNaiveFuncFullNum}{23/200}
\newcommand{\rqTwoQwenCrustDtvFuncFullNum}{19/200}
\newcommand{\rqTwoQwenCrustNaiveFuncCondFullPass}{15.23\%}
\newcommand{\rqTwoQwenCrustDtvFuncCondFullPass}{11.31\%}
\newcommand{\rqTwoQwenCrustFuncDelta}{-2.0~pp}
\newcommand{\rqTwoQwenCrustFuncMcnemarP}{=0.424}
\newcommand{\rqTwoQwenCrustFuncOnlyNaive}{9}
\newcommand{\rqTwoQwenCrustFuncOnlyDtv}{5}

\newcommand{\rqTwoGemmaCrustNaiveFuncFull}{15.00\%}
\newcommand{\rqTwoGemmaCrustDtvFuncFull}{15.00\%}
\newcommand{\rqTwoGemmaCrustNaiveFuncFullNum}{30/200}
\newcommand{\rqTwoGemmaCrustDtvFuncFullNum}{30/200}
\newcommand{\rqTwoGemmaCrustNaiveFuncCondFullPass}{16.22\%}
\newcommand{\rqTwoGemmaCrustDtvFuncCondFullPass}{17.65\%}
\newcommand{\rqTwoGemmaCrustFuncDelta}{+0.0~pp}
\newcommand{\rqTwoGemmaCrustFuncMcnemarP}{=1.000}
\newcommand{\rqTwoGemmaCrustFuncOnlyNaive}{8}
\newcommand{\rqTwoGemmaCrustFuncOnlyDtv}{8}

\newcommand{\rqTwoQwenThirtyBCrustNaivePass}{92\%}
\newcommand{\rqTwoQwenThirtyBCrustNaiveBudgetTwoPass}{70\%}
\newcommand{\rqTwoQwenThirtyBCrustDtvPass}{97\%}
\newcommand{\rqTwoQwenThirtyBCrustDtvBudgetTwoPass}{87\%}
\newcommand{\rqTwoQwenThirtyBJstsNaivePass}{57\%}
\newcommand{\rqTwoQwenThirtyBJstsNaiveBudgetTwoPass}{18\%}
\newcommand{\rqTwoQwenThirtyBJstsDtvPass}{66\%}
\newcommand{\rqTwoQwenThirtyBJstsDtvBudgetTwoPass}{45\%}
\newcommand{\rqTwoQwenThirtyBCrustNaiveOneshotPass}{67\%}
\newcommand{\rqTwoQwenThirtyBJstsNaiveOneshotPass}{18\%}
\newcommand{\rqTwoQwenThirtyBCrustDtvOneshotPass}{89\%}
\newcommand{\rqTwoQwenThirtyBJstsDtvOneshotPass}{48\%}
\newcommand{\rqTwoQwenThirtyBCrustNaiveOneshotK}{1.13}
\newcommand{\rqTwoQwenThirtyBCrustDtvOneshotK}{1.25}
\newcommand{\rqTwoQwenThirtyBJstsNaiveOneshotK}{1.13}
\newcommand{\rqTwoQwenThirtyBJstsDtvOneshotK}{1.62}

\newcommand{\rqTwoMainResultsTable}{%
\begin{tabular}{llrrrr}
\toprule
\textbf{Cell} & \textbf{Strategy} & \textbf{Pass} & \textbf{Pass rate} & \textbf{Avg.\ $k$} & \textbf{Tokens / pass} \\
\midrule
Qwen3-4B $\times$ C $\to$ Rust ($n=$200) & naive/self-refine & 152/200 & 76.0\% & 6.77 & 1798 \\
 & \methodabbr{}/self-refine & 169/200 & 84.5\% & 5.17 & 1805 \\
\midrule
Gemma-4-E4B $\times$ C $\to$ Rust ($n=$200) & naive/self-refine & 187/200 & 93.5\% & 5.91 & 3327 \\
 & \methodabbr{}/self-refine & 172/200 & 86.0\% & 5.99 & 3024 \\
\midrule
Qwen3-4B $\times$ JS $\to$ TS ($n=$100) & naive/self-refine & 37/100 & 37.0\% & 10.96 & 1740 \\
 & \methodabbr{}/self-refine & 50/100 & 50.0\% & 9.70 & 2185 \\
\midrule
Gemma-4-E4B $\times$ JS $\to$ TS ($n=$100) & naive/self-refine & 50/100 & 50.0\% & 9.87 & 3515 \\
 & \methodabbr{}/self-refine & 58/100 & 58.0\% & 8.43 & 3692 \\
\bottomrule
\end{tabular}
}

\newcommand{\rqTwoPairTable}{%
\begin{tabular}{lrrrrl}
\toprule
\textbf{Cell} & \textbf{$n$} & \textbf{Only naive} & \textbf{Only \methodabbr{}} & \textbf{Diff (pp)} & \textbf{McNemar $p$} \\
\midrule
Qwen3-4B $\times$ C $\to$ Rust & 200 & 15 & 32 & +8.5 & $0.019$ \\
Gemma-4-E4B $\times$ C $\to$ Rust & 200 & 19 & 4 & -7.5 & $0.003$ \\
Qwen3-4B $\times$ JS $\to$ TS & 100 & 6 & 19 & +13.0 & $0.015$ \\
Gemma-4-E4B $\times$ JS $\to$ TS & 100 & 11 & 19 & +8.0 & $0.200$ \\
\bottomrule
\end{tabular}
}

\newcommand{\rqTwoPairedKTable}{%
\begin{tabular}{lrrrr}
\toprule
\textbf{Cell} & \textbf{Both-pass $n$} & \textbf{naive avg.\ $k$} & \textbf{\methodabbr{} avg.\ $k$} & \textbf{$\Delta k$ (\%)} \\
\midrule
Qwen3-4B $\times$ C $\to$ Rust & 137 & 3.54 & 2.95 & -16.5\% \\
Gemma-4-E4B $\times$ C $\to$ Rust & 168 & 5.10 & 4.66 & -8.7\% \\
Qwen3-4B $\times$ JS $\to$ TS & 31 & 2.54 & 3.39 & +33.3\% \\
Gemma-4-E4B $\times$ JS $\to$ TS & 39 & 3.01 & 2.49 & -17.2\% \\
\bottomrule
\end{tabular}
}

\newcommand{\rqTwoCrustFunctionalTable}{%
\begin{tabular}{llrrrl}
\toprule
\textbf{Cell} & \textbf{Strategy} & \textbf{Compile} & \textbf{Full success} & \textbf{Full $\mid$ compile} & \textbf{McNemar $p$ (full)} \\
\midrule
Qwen3-4B $\times$ C $\to$ Rust & naive/self-refine & 151/200 & 23/200 (11.50\%) & 23/151 (15.23\%) & $0.424$ \\
 & \methodabbr{}/self-refine & 168/200 & 19/200 (9.50\%) & 19/168 (11.31\%) &  \\
\midrule
Gemma-4-E4B $\times$ C $\to$ Rust & naive/self-refine & 185/200 & 30/200 (15.00\%) & 30/185 (16.22\%) & $1.000$ \\
 & \methodabbr{}/self-refine & 170/200 & 30/200 (15.00\%) & 30/170 (17.65\%) &  \\
\bottomrule
\end{tabular}
}

\newcommand{\rqTwoQwenThirtyBOneshotTable}{%
\begin{tabular}{llrr}
\toprule
Task & Configuration & Pass / cases & Mean $k$ \\
\midrule
C to Rust & na\"ive/one-shot & 67/100 & 1.13 \\
C to Rust & \methodabbr{}/one-shot & 89/100 & 1.25 \\
JS to TS & na\"ive/one-shot & 18/100 & 1.13 \\
JS to TS & \methodabbr{}/one-shot & 48/100 & 1.62 \\
\bottomrule
\end{tabular}%
}

\newcommand{\rqThreeCrustDtvSrPass}{85.0\%}
\newcommand{\rqThreeCrustDtvSrNumPass}{85}
\newcommand{\rqThreeCrustDtvSrNumTotal}{100}
\newcommand{\rqThreeCrustDtvSrInnerOnePass}{44.0\%}
\newcommand{\rqThreeCrustDtvSrInnerOneNum}{44}
\newcommand{\rqThreeCrustDtvSrFuncPass}{10.0\%}
\newcommand{\rqThreeCrustDtvSrFuncNum}{10}
\newcommand{\rqThreeCrustDtvSrAvgTok}{3580}
\newcommand{\rqThreeCrustDtvSrAvgOuterRounds}{3.1}
\newcommand{\rqThreeCrustDtvSrTokensPerPass}{1694}
\newcommand{\rqThreeCrustDtvSrTokensPerPassCiLo}{1164}
\newcommand{\rqThreeCrustDtvSrTokensPerPassCiHi}{2405}
\newcommand{\rqThreeCrustNoFeedbackPass}{74.0\%}
\newcommand{\rqThreeCrustNoFeedbackNumPass}{74}
\newcommand{\rqThreeCrustNoFeedbackNumTotal}{100}
\newcommand{\rqThreeCrustNoFeedbackInnerOnePass}{21.0\%}
\newcommand{\rqThreeCrustNoFeedbackInnerOneNum}{21}
\newcommand{\rqThreeCrustNoFeedbackFuncPass}{8.0\%}
\newcommand{\rqThreeCrustNoFeedbackFuncNum}{8}
\newcommand{\rqThreeCrustNoFeedbackAvgTok}{4138}
\newcommand{\rqThreeCrustNoFeedbackAvgOuterRounds}{6.0}
\newcommand{\rqThreeCrustNoFeedbackTokensPerPass}{1412}
\newcommand{\rqThreeCrustNoFeedbackTokensPerPassCiLo}{1157}
\newcommand{\rqThreeCrustNoFeedbackTokensPerPassCiHi}{1691}
\newcommand{\rqThreeCrustNoEscalationPass}{79.0\%}
\newcommand{\rqThreeCrustNoEscalationNumPass}{79}
\newcommand{\rqThreeCrustNoEscalationNumTotal}{100}
\newcommand{\rqThreeCrustNoEscalationInnerOnePass}{44.0\%}
\newcommand{\rqThreeCrustNoEscalationInnerOneNum}{44}
\newcommand{\rqThreeCrustNoEscalationFuncPass}{5.0\%}
\newcommand{\rqThreeCrustNoEscalationFuncNum}{5}
\newcommand{\rqThreeCrustNoEscalationAvgTok}{3623}
\newcommand{\rqThreeCrustNoEscalationAvgOuterRounds}{2.3}
\newcommand{\rqThreeCrustNoEscalationTokensPerPass}{1437}
\newcommand{\rqThreeCrustNoEscalationTokensPerPassCiLo}{1025}
\newcommand{\rqThreeCrustNoEscalationTokensPerPassCiHi}{2000}
\newcommand{\rqThreeCrustDetectAbortPass}{70.0\%}
\newcommand{\rqThreeCrustDetectAbortNumPass}{70}
\newcommand{\rqThreeCrustDetectAbortNumTotal}{100}
\newcommand{\rqThreeCrustDetectAbortInnerOnePass}{14.0\%}
\newcommand{\rqThreeCrustDetectAbortInnerOneNum}{14}
\newcommand{\rqThreeCrustDetectAbortFuncPass}{8.0\%}
\newcommand{\rqThreeCrustDetectAbortFuncNum}{8}
\newcommand{\rqThreeCrustDetectAbortAvgTok}{4417}
\newcommand{\rqThreeCrustDetectAbortAvgOuterRounds}{13.5}
\newcommand{\rqThreeCrustDetectAbortTokensPerPass}{1173}
\newcommand{\rqThreeCrustDetectAbortTokensPerPassCiLo}{937}
\newcommand{\rqThreeCrustDetectAbortTokensPerPassCiHi}{1416}

\newcommand{\rqThreeCrustDtvSrVsNoFeedbackDiffPp}{-11.0}
\newcommand{\rqThreeCrustDtvSrVsNoFeedbackMcnemarP}{=0.035}
\newcommand{\rqThreeCrustDtvSrVsNoFeedbackOnlyA}{17}
\newcommand{\rqThreeCrustDtvSrVsNoFeedbackOnlyB}{6}
\newcommand{\rqThreeCrustDtvSrVsNoFeedbackN}{100}
\newcommand{\rqThreeCrustDtvSrVsNoFeedbackInnerOneDiffPp}{-23.0}
\newcommand{\rqThreeCrustDtvSrVsNoFeedbackInnerOneMcnemarP}{<0.001}
\newcommand{\rqThreeCrustDtvSrVsNoFeedbackFuncDiffPp}{-2.0}
\newcommand{\rqThreeCrustDtvSrVsNoFeedbackFuncMcnemarP}{=0.754}
\newcommand{\rqThreeCrustDtvSrVsNoFeedbackAvgTokDiff}{+558}
\newcommand{\rqThreeCrustDtvSrVsNoEscalationDiffPp}{-6.0}
\newcommand{\rqThreeCrustDtvSrVsNoEscalationMcnemarP}{=0.263}
\newcommand{\rqThreeCrustDtvSrVsNoEscalationOnlyA}{13}
\newcommand{\rqThreeCrustDtvSrVsNoEscalationOnlyB}{7}
\newcommand{\rqThreeCrustDtvSrVsNoEscalationN}{100}
\newcommand{\rqThreeCrustDtvSrVsNoEscalationInnerOneDiffPp}{+0.0}
\newcommand{\rqThreeCrustDtvSrVsNoEscalationInnerOneMcnemarP}{=1.000}
\newcommand{\rqThreeCrustDtvSrVsNoEscalationFuncDiffPp}{-5.0}
\newcommand{\rqThreeCrustDtvSrVsNoEscalationFuncMcnemarP}{=0.180}
\newcommand{\rqThreeCrustDtvSrVsNoEscalationAvgTokDiff}{+43}
\newcommand{\rqThreeCrustDtvSrVsDetectAbortDiffPp}{-15.0}
\newcommand{\rqThreeCrustDtvSrVsDetectAbortMcnemarP}{=0.003}
\newcommand{\rqThreeCrustDtvSrVsDetectAbortOnlyA}{19}
\newcommand{\rqThreeCrustDtvSrVsDetectAbortOnlyB}{4}
\newcommand{\rqThreeCrustDtvSrVsDetectAbortN}{100}
\newcommand{\rqThreeCrustDtvSrVsDetectAbortInnerOneDiffPp}{-30.0}
\newcommand{\rqThreeCrustDtvSrVsDetectAbortInnerOneMcnemarP}{<0.001}
\newcommand{\rqThreeCrustDtvSrVsDetectAbortFuncDiffPp}{-2.0}
\newcommand{\rqThreeCrustDtvSrVsDetectAbortFuncMcnemarP}{=0.688}
\newcommand{\rqThreeCrustDtvSrVsDetectAbortAvgTokDiff}{+837}

\newcommand{\rqThreeCrustNoFeedbackVsNoEscalationDiffPp}{+5.0}
\newcommand{\rqThreeCrustNoFeedbackVsNoEscalationMcnemarP}{=0.383}
\newcommand{\rqThreeCrustNoFeedbackVsNoEscalationOnlyA}{8}
\newcommand{\rqThreeCrustNoFeedbackVsNoEscalationOnlyB}{13}
\newcommand{\rqThreeCrustNoFeedbackVsDetectAbortDiffPp}{-4.0}
\newcommand{\rqThreeCrustNoFeedbackVsDetectAbortMcnemarP}{=0.454}
\newcommand{\rqThreeCrustNoFeedbackVsDetectAbortOnlyA}{10}
\newcommand{\rqThreeCrustNoFeedbackVsDetectAbortOnlyB}{6}
\newcommand{\rqThreeCrustNoEscalationVsDetectAbortDiffPp}{-9.0}
\newcommand{\rqThreeCrustNoEscalationVsDetectAbortMcnemarP}{=0.136}
\newcommand{\rqThreeCrustNoEscalationVsDetectAbortOnlyA}{19}
\newcommand{\rqThreeCrustNoEscalationVsDetectAbortOnlyB}{10}

\newcommand{\rqThreeAblationTable}{%
\begin{tabular}{lrrrr}
\toprule
\textbf{Configuration} & \textbf{Pass rate} & \textbf{$\Delta$ pp} & \textbf{Avg.\ tokens} & \textbf{$\Delta$ tokens} \\
\midrule
\methodabbr{}-full & 85.0\% & -- & 3580 & -- \\
\noalign{\vskip 1.5pt}\cdashline{1-5}\noalign{\vskip 2.5pt}
\methodabbr{}-no-feedback & 74.0\% & -11.0 & 4138 & +558 \\
\methodabbr{}-no-escalation & 79.0\% & -6.0 & 3623 & +43 \\
\methodabbr{}-detect-and-abort & 70.0\% & -15.0 & 4417 & +837 \\
\bottomrule
\end{tabular}
}

\newcommand{\rqThreeMainResultsTable}{%
\begin{tabular}{lrrr}
\toprule
\textbf{Configuration} & \textbf{Pass rate} & \textbf{Avg.\ tokens} & \textbf{Tokens / pass [95\% CI]} \\
\midrule
\methodabbr{}-full & 85/100 (85.0\%) & 3580 & 1694 [1164,2405] \\
\methodabbr{}-no-feedback & 74/100 (74.0\%) & 4138 & 1412 [1157,1691] \\
\methodabbr{}-no-escalation & 79/100 (79.0\%) & 3623 & 1437 [1025,2000] \\
\methodabbr{}-detect-and-abort & 70/100 (70.0\%) & 4417 & 1173 [937,1416] \\
\bottomrule
\end{tabular}
}

\newcommand{\rqThreePairTable}{%
\begin{tabular}{llrrrrl}
\toprule
\textbf{Comparison} & \textbf{Metric} & \textbf{Only-A} & \textbf{Only-B} & \textbf{$n$} & \textbf{Diff (pp)} & \textbf{McNemar $p$} \\
\midrule
\methodabbr{}-full vs \methodabbr{}-no-feedback & Compile (HEADLINE) & 17 & 6 & 100 & -11.0 & $0.035$ \\
\methodabbr{}-full vs \methodabbr{}-no-feedback & Inner-1shot & 29 & 6 & 100 & -23.0 & $<0.001$ \\
\methodabbr{}-full vs \methodabbr{}-no-feedback & Functional (guardrail) & 6 & 4 & 100 & -2.0 & $0.754$ \\
\midrule
\methodabbr{}-full vs \methodabbr{}-no-escalation & Compile (HEADLINE) & 13 & 7 & 100 & -6.0 & $0.263$ \\
\methodabbr{}-full vs \methodabbr{}-no-escalation & Inner-1shot & 17 & 17 & 100 & +0.0 & $1.000$ \\
\methodabbr{}-full vs \methodabbr{}-no-escalation & Functional (guardrail) & 7 & 2 & 100 & -5.0 & $0.180$ \\
\midrule
\methodabbr{}-full vs \methodabbr{}-detect-and-abort & Compile (HEADLINE) & 19 & 4 & 100 & -15.0 & $0.003$ \\
\methodabbr{}-full vs \methodabbr{}-detect-and-abort & Inner-1shot & 34 & 4 & 100 & -30.0 & $<0.001$ \\
\methodabbr{}-full vs \methodabbr{}-detect-and-abort & Functional (guardrail) & 4 & 2 & 100 & -2.0 & $0.688$ \\
\bottomrule
\end{tabular}
}

\newcommand{\rqThreeRescueTable}{%
\begin{tabular}{lrrrr}
\toprule
\textbf{Configuration} & \textbf{Inner-1shot} & \textbf{Final compile} & \textbf{Rescued} & \textbf{Rescue rate} \\
\midrule
\methodabbr{}-full & 44/100 (44.0\%) & 85/100 (85.0\%) & 41/56 & 73.2\% \\
\methodabbr{}-no-feedback & 21/100 (21.0\%) & 74/100 (74.0\%) & 53/79 & 67.1\% \\
\methodabbr{}-no-escalation & 44/100 (44.0\%) & 79/100 (79.0\%) & 35/56 & 62.5\% \\
\methodabbr{}-detect-and-abort & 14/100 (14.0\%) & 70/100 (70.0\%) & 56/86 & 65.1\% \\
\bottomrule
\end{tabular}
}

\newcommand{\rqFourCrustOneshotUniverseN}{230}
\newcommand{\rqFourCrustAgreementEligibleN}{159}
\newcommand{\rqFourCrustAgreementExact}{88.1\%}
\newcommand{\rqFourCrustAgreementLoose}{91.2\%}
\newcommand{\rqFourCrustOptALocalShare}{12}
\newcommand{\rqFourCrustOptALocalDenom}{12}
\newcommand{\rqFourCrustOptALocalRescued}{6}
\newcommand{\rqFourCrustOptALocalRate}{50.0\%}
\newcommand{\rqFourCrustOptAMixedShare}{171}
\newcommand{\rqFourCrustOptAMixedDenom}{168}
\newcommand{\rqFourCrustOptAMixedRescued}{73}
\newcommand{\rqFourCrustOptAMixedRate}{43.5\%}
\newcommand{\rqFourCrustOptANonlocalShare}{28}
\newcommand{\rqFourCrustOptANonlocalDenom}{27}
\newcommand{\rqFourCrustOptANonlocalRescued}{8}
\newcommand{\rqFourCrustOptANonlocalRate}{29.6\%}
\newcommand{\rqFourCrustOptAUnknownShare}{19}
\newcommand{\rqFourCrustOptAUnknownDenom}{19}
\newcommand{\rqFourCrustOptAUnknownRescued}{11}
\newcommand{\rqFourCrustOptAUnknownRate}{57.9\%}
\newcommand{\rqFourCrustOptAUniverseN}{230}
\newcommand{\rqFourCrustOptBLocalShare}{2}
\newcommand{\rqFourCrustOptBLocalDenom}{2}
\newcommand{\rqFourCrustOptBLocalRescued}{2}
\newcommand{\rqFourCrustOptBLocalRate}{100.0\%}
\newcommand{\rqFourCrustOptBMixedShare}{131}
\newcommand{\rqFourCrustOptBMixedDenom}{129}
\newcommand{\rqFourCrustOptBMixedRescued}{65}
\newcommand{\rqFourCrustOptBMixedRate}{50.4\%}
\newcommand{\rqFourCrustOptBNonlocalShare}{10}
\newcommand{\rqFourCrustOptBNonlocalDenom}{10}
\newcommand{\rqFourCrustOptBNonlocalRescued}{4}
\newcommand{\rqFourCrustOptBNonlocalRate}{40.0\%}
\newcommand{\rqFourCrustOptBUnknownShare}{16}
\newcommand{\rqFourCrustOptBUnknownDenom}{16}
\newcommand{\rqFourCrustOptBUnknownRescued}{12}
\newcommand{\rqFourCrustOptBUnknownRate}{75.0\%}
\newcommand{\rqFourCrustOptBUniverseN}{159}

\newcommand{\rqFourCrustRescuedN}{105}
\newcommand{\rqFourCrustBOneN}{20}
\newcommand{\rqFourCrustBOnePct}{19.0\%}
\newcommand{\rqFourCrustBTwoN}{85}
\newcommand{\rqFourCrustBTwoPct}{81.0\%}
\newcommand{\rqFourCrustSharedRatioN}{57}
\newcommand{\rqFourCrustSharedRatioMean}{1.19}
\newcommand{\rqFourCrustSharedRatioMedian}{1.00}
\newcommand{\rqFourCrustSharedRatioQOne}{1.00}
\newcommand{\rqFourCrustSharedRatioQThree}{1.50}
\newcommand{\rqFourCrustBothFailN}{146}
\newcommand{\rqFourCrustBothFailSameCodeN}{59}
\newcommand{\rqFourCrustBothFailSameCodePct}{40.4\%}

\newcommand{\rqFourCrustSrNTotal}{300}
\newcommand{\rqFourCrustSrNaivePass}{72.3\%}
\newcommand{\rqFourCrustSrDtvPass}{82.0\%}
\newcommand{\rqFourCrustSrBothPassN}{196}
\newcommand{\rqFourCrustSrDtvOnlyN}{50}
\newcommand{\rqFourCrustSrNaiveOnlyN}{21}
\newcommand{\rqFourCrustSrBothFailN}{33}
\newcommand{\rqFourCrustSrNaiveMeanR}{2.87}
\newcommand{\rqFourCrustSrDtvMeanR}{1.65}
\newcommand{\rqFourCrustSrNaiveMedianR}{2}
\newcommand{\rqFourCrustSrDtvMedianR}{1}
\newcommand{\rqFourCrustSrPairedRatioMean}{2.06}
\newcommand{\rqFourCrustSrPairedRatioMedian}{2.00}
\newcommand{\rqFourCrustSrPairedRatioQOne}{1.00}
\newcommand{\rqFourCrustSrPairedRatioQThree}{3.00}
\newcommand{\rqFourCrustSrPairedRatioCiLo}{1.88}
\newcommand{\rqFourCrustSrPairedRatioCiHi}{2.27}
\newcommand{\rqFourCrustSrPairedDiffMean}{+1.22}
\newcommand{\rqFourCrustSrPairedDiffMedian}{+1}
\newcommand{\rqFourCrustSrPairedDiffCiLo}{+0.94}
\newcommand{\rqFourCrustSrPairedDiffCiHi}{+1.52}
\newcommand{\rqFourCrustSrEarlier}{129}
\newcommand{\rqFourCrustSrSame}{46}
\newcommand{\rqFourCrustSrLater}{21}
\newcommand{\rqFourCrustSrSignP}{<10^{-19}}

\newcommand{\rqFourCrustSrRescuedTotalN}{50}
\newcommand{\rqFourCrustSrRescuedClassifiedN}{43}
\newcommand{\rqFourCrustSrRescuedMissingN}{7}
\newcommand{\rqFourCrustOneshotRescuedClassifiedN}{98}
\newcommand{\rqFourCrustSrRescuedLocalN}{3}
\newcommand{\rqFourCrustSrRescuedLocalPct}{7.0\%}
\newcommand{\rqFourCrustOneshotRescuedLocalN}{6}
\newcommand{\rqFourCrustOneshotRescuedLocalPct}{6.1\%}
\newcommand{\rqFourCrustSrRescuedMixedN}{32}
\newcommand{\rqFourCrustSrRescuedMixedPct}{74.4\%}
\newcommand{\rqFourCrustOneshotRescuedMixedN}{73}
\newcommand{\rqFourCrustOneshotRescuedMixedPct}{74.5\%}
\newcommand{\rqFourCrustSrRescuedNonlocalN}{7}
\newcommand{\rqFourCrustSrRescuedNonlocalPct}{16.3\%}
\newcommand{\rqFourCrustOneshotRescuedNonlocalN}{8}
\newcommand{\rqFourCrustOneshotRescuedNonlocalPct}{8.2\%}
\newcommand{\rqFourCrustSrRescuedUnknownN}{1}
\newcommand{\rqFourCrustSrRescuedUnknownPct}{2.3\%}
\newcommand{\rqFourCrustOneshotRescuedUnknownN}{11}
\newcommand{\rqFourCrustOneshotRescuedUnknownPct}{11.2\%}

\newcommand{\rqFourJstsOneshotUniverseN}{141}
\newcommand{\rqFourJstsAgreementEligibleN}{44}
\newcommand{\rqFourJstsAgreementExact}{84.1\%}
\newcommand{\rqFourJstsAgreementLoose}{95.5\%}
\newcommand{\rqFourJstsOptALocalShare}{61}
\newcommand{\rqFourJstsOptALocalDenom}{58}
\newcommand{\rqFourJstsOptALocalRescued}{17}
\newcommand{\rqFourJstsOptALocalRate}{29.3\%}
\newcommand{\rqFourJstsOptAMixedShare}{70}
\newcommand{\rqFourJstsOptAMixedDenom}{67}
\newcommand{\rqFourJstsOptAMixedRescued}{8}
\newcommand{\rqFourJstsOptAMixedRate}{11.9\%}
\newcommand{\rqFourJstsOptANonlocalShare}{5}
\newcommand{\rqFourJstsOptANonlocalDenom}{2}
\newcommand{\rqFourJstsOptANonlocalRescued}{0}
\newcommand{\rqFourJstsOptANonlocalRate}{0.0\%}
\newcommand{\rqFourJstsOptAUnknownShare}{5}
\newcommand{\rqFourJstsOptAUnknownDenom}{5}
\newcommand{\rqFourJstsOptAUnknownRescued}{1}
\newcommand{\rqFourJstsOptAUnknownRate}{20.0\%}
\newcommand{\rqFourJstsOptAUniverseN}{141}
\newcommand{\rqFourJstsOptBLocalShare}{26}
\newcommand{\rqFourJstsOptBLocalDenom}{25}
\newcommand{\rqFourJstsOptBLocalRescued}{12}
\newcommand{\rqFourJstsOptBLocalRate}{48.0\%}
\newcommand{\rqFourJstsOptBMixedShare}{16}
\newcommand{\rqFourJstsOptBMixedDenom}{16}
\newcommand{\rqFourJstsOptBMixedRescued}{6}
\newcommand{\rqFourJstsOptBMixedRate}{37.5\%}
\newcommand{\rqFourJstsOptBNonlocalShare}{0}
\newcommand{\rqFourJstsOptBNonlocalDenom}{0}
\newcommand{\rqFourJstsOptBNonlocalRescued}{0}
\newcommand{\rqFourJstsOptBNonlocalRate}{n/a}
\newcommand{\rqFourJstsOptBUnknownShare}{2}
\newcommand{\rqFourJstsOptBUnknownDenom}{2}
\newcommand{\rqFourJstsOptBUnknownRescued}{0}
\newcommand{\rqFourJstsOptBUnknownRate}{0.0\%}
\newcommand{\rqFourJstsOptBUniverseN}{44}

\newcommand{\rqFourJstsRescuedN}{26}
\newcommand{\rqFourJstsBOneN}{0}
\newcommand{\rqFourJstsBOnePct}{0.0\%}
\newcommand{\rqFourJstsBTwoN}{26}
\newcommand{\rqFourJstsBTwoPct}{100.0\%}
\newcommand{\rqFourJstsSharedRatioN}{26}
\newcommand{\rqFourJstsSharedRatioMean}{0.71}
\newcommand{\rqFourJstsSharedRatioMedian}{0.50}
\newcommand{\rqFourJstsSharedRatioQOne}{0.23}
\newcommand{\rqFourJstsSharedRatioQThree}{1.00}
\newcommand{\rqFourJstsBothFailN}{117}
\newcommand{\rqFourJstsBothFailSameCodeN}{101}
\newcommand{\rqFourJstsBothFailSameCodePct}{86.3\%}

\newcommand{\rqFourJstsSrNTotal}{150}
\newcommand{\rqFourJstsSrNaivePass}{33.3\%}
\newcommand{\rqFourJstsSrDtvPass}{46.0\%}
\newcommand{\rqFourJstsSrBothPassN}{43}
\newcommand{\rqFourJstsSrDtvOnlyN}{26}
\newcommand{\rqFourJstsSrNaiveOnlyN}{7}
\newcommand{\rqFourJstsSrBothFailN}{74}
\newcommand{\rqFourJstsSrNaiveMeanR}{2.28}
\newcommand{\rqFourJstsSrDtvMeanR}{1.65}
\newcommand{\rqFourJstsSrNaiveMedianR}{2}
\newcommand{\rqFourJstsSrDtvMedianR}{1}
\newcommand{\rqFourJstsSrPairedRatioMean}{1.61}
\newcommand{\rqFourJstsSrPairedRatioMedian}{1.50}
\newcommand{\rqFourJstsSrPairedRatioQOne}{1.00}
\newcommand{\rqFourJstsSrPairedRatioQThree}{2.00}
\newcommand{\rqFourJstsSrPairedRatioCiLo}{1.38}
\newcommand{\rqFourJstsSrPairedRatioCiHi}{1.88}
\newcommand{\rqFourJstsSrPairedDiffMean}{+0.63}
\newcommand{\rqFourJstsSrPairedDiffMedian}{+1}
\newcommand{\rqFourJstsSrPairedDiffCiLo}{+0.35}
\newcommand{\rqFourJstsSrPairedDiffCiHi}{+0.93}
\newcommand{\rqFourJstsSrEarlier}{24}
\newcommand{\rqFourJstsSrSame}{15}
\newcommand{\rqFourJstsSrLater}{4}
\newcommand{\rqFourJstsSrSignP}{<0.001}

\newcommand{\rqFourJstsSrRescuedTotalN}{26}
\newcommand{\rqFourJstsSrRescuedClassifiedN}{26}
\newcommand{\rqFourJstsSrRescuedMissingN}{0}
\newcommand{\rqFourJstsOneshotRescuedClassifiedN}{26}
\newcommand{\rqFourJstsSrRescuedLocalN}{9}
\newcommand{\rqFourJstsSrRescuedLocalPct}{34.6\%}
\newcommand{\rqFourJstsOneshotRescuedLocalN}{17}
\newcommand{\rqFourJstsOneshotRescuedLocalPct}{65.4\%}
\newcommand{\rqFourJstsSrRescuedMixedN}{16}
\newcommand{\rqFourJstsSrRescuedMixedPct}{61.5\%}
\newcommand{\rqFourJstsOneshotRescuedMixedN}{8}
\newcommand{\rqFourJstsOneshotRescuedMixedPct}{30.8\%}
\newcommand{\rqFourJstsSrRescuedNonlocalN}{0}
\newcommand{\rqFourJstsSrRescuedNonlocalPct}{0.0\%}
\newcommand{\rqFourJstsOneshotRescuedNonlocalN}{0}
\newcommand{\rqFourJstsOneshotRescuedNonlocalPct}{0.0\%}
\newcommand{\rqFourJstsSrRescuedUnknownN}{1}
\newcommand{\rqFourJstsSrRescuedUnknownPct}{3.8\%}
\newcommand{\rqFourJstsOneshotRescuedUnknownN}{1}
\newcommand{\rqFourJstsOneshotRescuedUnknownPct}{3.8\%}

\newcommand{\rqFourFixshapeTableMain}{%
\begin{tabular}{lcccc}
\toprule
\textbf{Task} & \textbf{Local} & \textbf{Mixed} & \textbf{Nonlocal} & \textbf{$\Delta_{\text{L-N}}$} \\
\midrule
C $\to$ Rust & \heatcell{50}{50.0\%}{12} & \heatcell{43}{43.5\%}{168} & \heatcell{30}{29.6\%}{27} & $\mathbf{+20.4}$~pp \\
JS $\to$ TS & \heatcell{29}{29.3\%}{58} & \heatcell{12}{11.9\%}{67} & \heatcell{0}{\phantom{0}0.0\%}{2$^{*}$} & $\mathbf{+29.3}$~pp \\
\bottomrule
\end{tabular}
}

\newcommand{\rqFourFixshapeTableFull}{%
\begin{tabular}{lcccc}
\toprule
\textbf{Fix-shape} & \textbf{C $\to$ Rust A} & \textbf{C $\to$ Rust B} & \textbf{JS $\to$ TS A} & \textbf{JS $\to$ TS B} \\
\midrule
Local & 50.0\% (12) & 100.0\% (2*) & 29.3\% (58) & 48.0\% (25) \\
Mixed & 43.5\% (168) & 50.4\% (129) & 11.9\% (67) & 37.5\% (16) \\
Nonlocal & 29.6\% (27) & 40.0\% (10) & 0.0\% (2*) & -- (0) \\
Unknown & 57.9\% (19) & 75.0\% (16) & 20.0\% (5) & 0.0\% (2*) \\
\bottomrule
\end{tabular}
}

\newcommand{\rqFourPercodeTable}{%
\begin{tabular}{llrrr}
\toprule
\textbf{Task} & \textbf{Code} & \textbf{Naive-fail $n$} & \textbf{\methodabbr{} rescued} & \textbf{Rescue rate} \\
\midrule
C $\to$ Rust & \texttt{E0599} & 79 & 43 & 54.4\% \\
 & \texttt{E0308} & 66 & 22 & 33.3\% \\
 & \texttt{E0277} & 61 & 24 & 39.3\% \\
 & \texttt{E0425} & 49 & 15 & 30.6\% \\
 & \texttt{E0433} & 14 & 4 & 28.6\% \\
 & \texttt{E0384} & 8 & 4 & 50.0\% \\
\midrule
JS $\to$ TS & \texttt{@typescript-eslint/typedef} & 118 & 23 & 19.5\% \\
 & \texttt{@typescript-eslint/explicit-function-return-type} & 58 & 7 & 12.1\% \\
 & \texttt{TS2339} & 43 & 1 & 2.3\% \\
 & \texttt{TS2554} & 28 & 1 & 3.6\% \\
 & \texttt{TS1192} & 21 & 0 & 0.0\% \\
 & \texttt{TS1259} & 16 & 0 & 0.0\% \\
\bottomrule
\end{tabular}
}

\newcommand{\rqFourPairedRoundsTable}{%
\begin{tabular}{lrrrrrrl}
\toprule
\textbf{Task} & \textbf{$n$} & \textbf{med ratio} & \textbf{mean ratio [95\% CI]} & \textbf{e/s/l} & \textbf{$\Delta$ mean} & \textbf{$\Delta$ [95\% CI]} & \textbf{Sign $p$} \\
\midrule
C $\to$ Rust & 196 & 2.00 & 2.06 [1.88, 2.27] & 129/46/21 & +1.22 & [+0.94, +1.52] & $<10^{-19}$ \\
JS $\to$ TS & 43 & 1.50 & 1.61 [1.38, 1.88] & 24/15/4 & +0.63 & [+0.35, +0.93] & $<0.001$ \\
\bottomrule
\end{tabular}
}

\newcommand{\rqFourFixshapeTransferTable}{%
\begin{tabular}{lccccc}
\toprule
\textbf{Task} & \textbf{Subset} & \textbf{LOCAL} & \textbf{MIXED} & \textbf{NONLOCAL} & \textbf{UNKNOWN} \\
\midrule
C $\to$ Rust & SR rescued ($n=$43) & 7.0\% (3) & 74.4\% (32) & 16.3\% (7) & 2.3\% (1) \\
 & One-shot rescued ($n=$98) & 6.1\% (6) & 74.5\% (73) & 8.2\% (8) & 11.2\% (11) \\
\midrule
JS $\to$ TS & SR rescued ($n=$26) & 34.6\% (9) & 61.5\% (16) & 0.0\% (0) & 3.8\% (1) \\
 & One-shot rescued ($n=$26) & 65.4\% (17) & 30.8\% (8) & 0.0\% (0) & 3.8\% (1) \\
\bottomrule
\end{tabular}
}

\newcommand{\regenerationSignalCaseCount}{81}
\newcommand{\regenerationSignalDiagnosticCount}{36}
\newcommand{\regenerationSignalDiagnosticPass}{44.4\%}
\newcommand{\regenerationSignalDecayCount}{22}
\newcommand{\regenerationSignalDecayPass}{27.2\%}
\newcommand{\regenerationSignalDiff}{17.3}
\newcommand{\regenerationSignalMcnemarP}{0.0066}
\newcommand{\regenerationSignalCiLo}{6.2}
\newcommand{\regenerationSignalCiHi}{28.4}

\newcommand{\rollbackLocalizationUpstreamCount}{45}
\newcommand{\rollbackLocalizationCompilingCaseCount}{67}
\newcommand{\rollbackLocalizationUpstreamPct}{67.2\%}
\newcommand{\rollbackLocalizationEntropyExact}{2}
\newcommand{\rollbackLocalizationEntropyExposes}{17}
\newcommand{\rollbackLocalizationEntropyLate}{28}
\newcommand{\rollbackLocalizationEntropyDistance}{+3}

\newcommand{\strictRecheckQwenFourBNaivePass}{20.7\%}
\newcommand{\strictRecheckQwenFourBDtvPass}{26.7\%}
\newcommand{\strictRecheckGemmaNaivePass}{30.0\%}
\newcommand{\strictRecheckGemmaDtvPass}{36.0\%}
\newcommand{\strictRecheckQwenThirtyBNaivePass}{34.0\%}
\newcommand{\strictRecheckQwenThirtyBDtvPass}{42.0\%}

\newcommand{\regenerationSignalPassTable}{%
\begin{tabular}{lrr}
\toprule
Regeneration signal & Pass / cases & Pass rate \\
\midrule
\methodabbr{} diagnostic & 36/81 & 44.4\% \\
ROCODE path decay & 22/81 & 27.2\% \\
\bottomrule
\end{tabular}%
}

\newcommand{\regenerationSignalPairsTable}{%
\begin{tabular}{llr}
\toprule
\methodabbr{} diagnostic & ROCODE path decay & Cases \\
\midrule
Pass & Pass & 17 \\
Pass & Fail & 19 \\
Fail & Pass & 5 \\
Fail & Fail & 40 \\
\bottomrule
\end{tabular}%
}

\newcommand{\rollbackLocalizationEntropyTable}{%
\begin{tabular}{lrrrr}
\toprule
Tokens included & Exact line & Exposes edit & Too late & Median $L_H-L_e$ \\
\midrule
All generated tokens & 2/45 & 17/45 & 28/45 & +3 \\
Code only & 2/45 & 17/45 & 28/45 & +3 \\
\bottomrule
\end{tabular}%
}

\newcommand{\strictRecheckTable}{%
\begin{tabular}{lrrrrr}
\toprule
Model & $n$ & \multicolumn{2}{c}{Shared final criterion} & \multicolumn{2}{c}{Final + strict} \\
 & & na\"ive & \methodabbr{} & na\"ive & \methodabbr{} \\
\midrule
Qwen3-4B & 150 & 33.3\% & 46.0\% & 20.7\% & 26.7\% \\
Gemma-4-E4B & 100 & 50.0\% & 58.0\% & 30.0\% & 36.0\% \\
Qwen3-30B-A3B & 100 & 57.0\% & 66.0\% & 34.0\% & 42.0\% \\
\bottomrule
\end{tabular}%
}

\begin{abstract}
Test-time scaling is an important mechanism for improving large language models, especially on tasks with deterministic verifiers. Code translation is a canonical example: the source program constrains valid outputs, while compilers, type checkers, and behavioral checks provide exact pass/fail feedback. Outcome-only approaches apply these verifiers only after generation, which is inefficient because early errors corrupt the autoregressive context and are rarely corrected later.
We introduce \emph{\method{}}, a framework that treats structural boundaries as meta steps for verifier-guided decoding. \methodabbr{} interleaves generation with verifier calls under a state-machine controller that commits verified prefixes, using structural-boundary checks and structure-aware rollback to prevent error propagation while reducing wasted tokens.
We evaluate \methodabbr{} on C-to-Rust and JavaScript-to-TypeScript translation. Using Qwen3-4B as the primary generator under matched token budgets, \methodabbr{} improves pass rates from \rqOneCrustNaiveSrPass{} to \rqOneCrustDtvSrPass{} on C-to-Rust and from \rqOneJstsNaiveSrPass{} to \rqOneJstsDtvSrPass{} on JavaScript-to-TypeScript relative to matched self-refinement baselines, while using fewer tokens per case; the same trend largely transfers to Gemma-4-E4B and Qwen3-30B-A3B. In the evaluated cost-matched grid, \methodabbr{} achieves a more favorable tradeoff between pass rate and generated-token cost than post-hoc verification or sampling-based scaling. These results show that verifier-guided decoding is an effective use of inference-time compute for code translation.
\end{abstract}

\section{Introduction}
\label{sec:introduction}

\begin{figure}[t]
\centering
\includegraphics[width=0.8\textwidth]{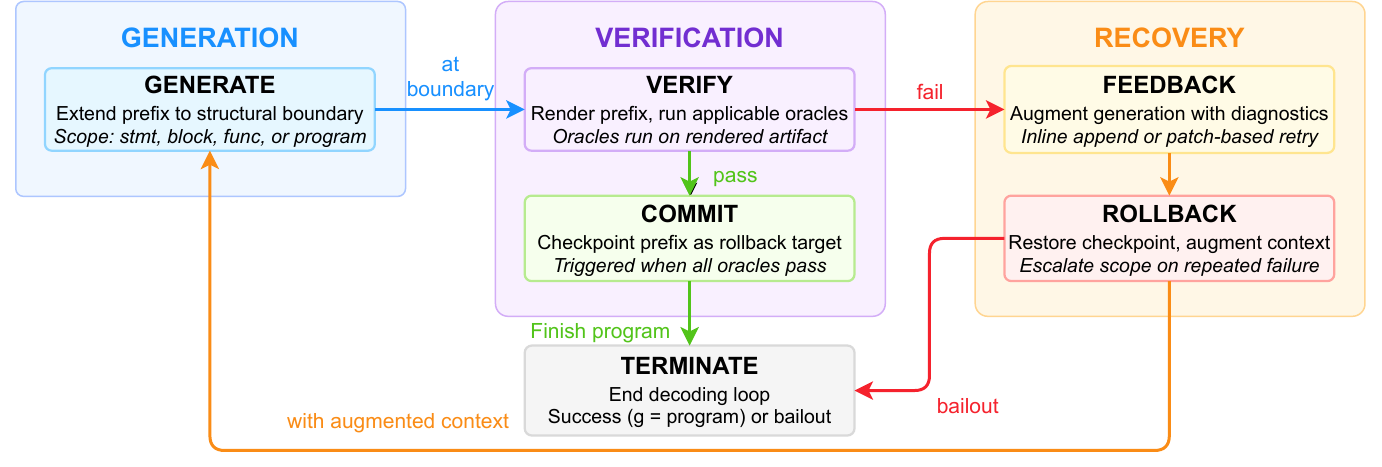}
\caption{\small %
\textbf{GENERATION} extends the target prefix to the next structural boundary (statement, block, function, or program). \textbf{VERIFICATION} renders the prefix and runs applicable oracles, committing on pass. \textbf{RECOVERY} parses diagnostics on fail, rolls back to the last checkpoint with augmented context, and escalates scope on repeated failure. The loop terminates on program-level success or budget bailout.\vspace{-7mm}}
\label{fig:architecture}
\end{figure}

Large language models (LLMs) have become capable general-purpose code generators, but improving performance on difficult inputs is increasingly pursued at inference time rather than solely through larger pretraining runs. A growing line of work studies \emph{test-time scaling}: allocating additional inference compute through repeated sampling, search, self-refinement, or process-level guidance to improve the performance of a frozen pretrained model~\citep{snellScalingLLMTestTime2025, muennighoffS1SimpleTesttime2025a, jiSurveyTestTimeCompute2025, lightmanLetsVerifyStep2024, openaiOpenAIO1System2024, liCompetitionlevelCodeGeneration2022}.

Most existing test-time scaling methods are designed for open-ended domains where correctness is difficult to verify exactly. Consequently, they guide generation using probabilistic signals derived from the model itself or learned scorers~\citep{brownLargeLanguageMonkeys2024, yaoTreeThoughtsDeliberate2023a}, such as token likelihoods, preference models, or intermediate reasoning heuristics. While these signals can improve reasoning quality in open-ended settings, coding tasks impose discrete semantic and syntactic correctness constraints, making probabilistic guidance only weakly aligned with true success: models may assign high probability to plausible-looking yet incorrect programs, while learned reward models for code are expensive to train and brittle across languages and libraries~\citep{spiessCalibrationCorrectnessLanguage2025, khalifaProcessRewardModels2025}.
At the same time, coding tasks expose deterministic external verifiers unavailable in most domains. Compilers, type checkers, and behavioral tests provide exact pass/fail feedback together with structured diagnostics, creating an opportunity for stronger inference-time guidance. Outcome-only approaches use these signals post-hoc to filter, rerank, or repair completed samples~\citep{niLEVERLearningVerify2023, chenCodeTCodeGeneration2022, chenTeachingLargeLanguage2024}. This cannot influence the trajectory of generation itself: once a verifier-detectable error enters the autoregressive prefix, later decoding proceeds under a corrupted conditioning context, and models tend to continue the trajectory rather than recover~\citep{aroraWhyExposureBias2022}.

One way to overcome this limitation is to let verifier signals participate directly in the autoregressive loop, governing which prefixes are admitted as conditioning context rather than only which completed samples are retained. ROCODE~\citep{jiang2025rocode}, EG-CFG~\citep{lavonExecutionGuidedLinebyLine2025}, and VerMCTS~\citep{brandfonbrenerVerMCTSSynthesizingMultiStep2024} establish this in-loop premise through backtracking, execution feedback, and partial-program search. Doing so raises three coupled questions: \emph{when} to invoke a verifier, since verifiers require structurally complete artifacts rather than arbitrary token prefixes; \emph{how} to roll back when verification fails, balancing context preservation against error excision; and \emph{what} to do with diagnostics beyond retry with the same context.

Building on this in-loop premise, we introduce \textit{\methodname{}} (\methodabbr{}), a decoding framework that interleaves generation from a pretrained code model with deterministic verifier calls under an explicit state-machine controller\footnote{Source code is available at: \url{https://github.com/qsdrqs/dtv-translation}}. As illustrated in Figure~\ref{fig:architecture}, \methodabbr{} generates target code up to the next structural boundary (e.g., statement, block, or function), renders the current prefix into a verifier-consumable artifact, and runs the applicable verifiers. If all checks pass, the prefix is committed; else, the controller rolls back to an earlier checkpoint, updates a feedback state using verifier diagnostics, and retries generation. Unlike prior in-loop verification systems that fix a particular combination of scope, verifier type, and search mechanism~\citep{lavonExecutionGuidedLinebyLine2025, brandfonbrenerVerMCTSSynthesizingMultiStep2024, aggarwalAlphaVerusBootstrappingFormally2024a}, \methodabbr{} separates the controller, verification scopes, and verifier suite as configurable components.
It combines three key mechanisms. First, \emph{structural-boundary verification} couples rendering and a unanimous-pass commit condition to maintain a \emph{legal prefix invariant}: committed prefixes can be extended into programs that pass all oracles. Second, \emph{structure-aware rollback with escalation} rolls back to the smallest scope, escalating to larger scopes when the same diagnostic persists rather than retrying more. Third, \emph{feedback via prompt augmentation} summarizes verifier diagnostics and injects them back into the model context so retries proceed under updated information rather than blind resampling. Together, these mechanisms prevent verifier-detectable errors from propagating through the autoregressive context by ensuring generation continues only from committed verified prefixes. While we instantiate \methodabbr{} on code translation, the framework in principle applies to any task with deterministic process-level verifiers and a structural renderer, such as code generation tasks with external knowledge validators.
In translation, the source and retained prefix constrain retries, while earlier decisions may first fail at later uses. Diagnostic feedback and expanding repair regions address both (\S~\ref{sec:method}).

We evaluate \methodabbr{} on two translation settings with complementary verifier structures. \emph{C-to-Rust} combines compiler checks with differential testing against the output in the end, while \emph{JavaScript-to-TypeScript} relies on static type and lint verification for the type annotation task. Using Qwen3-4B~\citep{qwenteamQwen3TechnicalReport2025} as the primary generator under matched generation budgets, \methodabbr{} with self-refinement improves compile-pass rate from \rqOneCrustNaiveSrPass{} to \rqOneCrustDtvSrPass{} on C-to-Rust and from \rqOneJstsNaiveSrPass{} to \rqOneJstsDtvSrPass{} on JavaScript-to-TypeScript, while using fewer tokens overall (\S~\ref{sec:rq1}); the same gains largely transfer to Gemma-4-E4B~\citep{farabetGemma4Byte2026} and Qwen3-30B-A3B (Appendix~\ref{app:rq2}).
These gains are not solely due to outer-loop scaling: even without outer-loop retry, in-loop verification alone improves one-shot pass rate by \rqOneCrustNaiveOneshotVsDtvOneshotDiffPp{}~pp and \rqOneJstsNaiveOneshotVsDtvOneshotDiffPp{}~pp over direct decoding. In the cost-matched grid, \methodabbr{} also achieves a more favorable tradeoff between pass rate and generated-token cost than other test-time scaling methods, including best-of-$N$ and the compile-feedback-based S\textsuperscript{*} baseline~\citep{liTestTimeScaling2025a}. Ablations show that all three key design components of \methodabbr{} contribute, with in-loop recovery providing the largest benefit (\S~\ref{sec:rq3}). Additional mechanism analyses, runtime measurements, and strict rechecking extend this evidence (Section~\ref{sec:rq1} and Appendix~\ref{app:repair_guidance}).

\section{\methodname{}}
\label{sec:dtv}

As discussed in \S~\ref{sec:introduction}, post-hoc verification applies verifier feedback only after generation is complete, surfacing errors too late to guide decoding. We instead verify \emph{during} decoding using deterministic program verifiers (e.g., compilers, type checkers, and tests) as oracles. Verifier calls are aligned with structural boundaries such as completed statements, blocks, or functions, where verdicts become meaningful and rollback can target the scope at which an error was introduced. This forms an \emph{inner loop} of generation that complements traditional \emph{outer-loop} strategies (e.g., self-refinement with verifier feedback) operating only on completed programs. Outer-loop verification is recovered as a special case by invoking oracles only at full-program scope.%

\subsection{Overview}
\label{sec:dtv_overview}

\method{} instantiates this idea as a deterministic state-machine controller. At each structural boundary, the controller selects from a fixed action set: \textsc{Generate} (extend the token prefix), \textsc{Verify} (render the prefix and run oracles), \textsc{Commit} (checkpoint the current prefix), \textsc{Rollback} (restore an earlier checkpoint), \textsc{Feedback} (construct a retry prompt from diagnostics), and \textsc{Terminate} (end decoding); Figure~\ref{fig:architecture} shows the overall control flow. While the framework is task-agnostic, we focus on code translation because it admits an abundance of deterministic oracles.%

\noindent{\bf Translation setting.}
Code translation aims for the generated program to exhibit the same observable behavior as the source. Given a source program $S$, a translation model $M$ generates a target token sequence $Y = (y_1, \dots, y_m)$. At each step $i$, $M$ defines a conditional distribution $p_M(y_i \mid y_{<i}, S)$, where $Y_{<i} = (y_1, \dots, y_{i-1})$ is the prefix and $Y_{\le i}$ is its length-$i$ extension.
We treat $M$ as a fixed pretrained black-box generator and assume access to a suite of deterministic oracles whose verdicts capture correctness of the translation result.

\noindent{\bf Key abstractions.}
We now formalize the three abstractions needed to state our algorithm: %

\emph{1. Scope.} Let $\mathcal{G}$ be a finite totally ordered set of scopes ordered by granularity, with $\mathcal{G} = \{\texttt{stmt}, \texttt{block}, \texttt{func}, \texttt{program}\}$ in our instantiations and $\texttt{stmt} \preceq \texttt{block} \preceq \texttt{func} \preceq \texttt{program}$ denoting nested enclosure (a statement inside a block, a block inside a function, and so on). Each scope $g \in \mathcal{G}$ also marks a class of \emph{structural boundaries}: positions in the target token sequence at which an instance of $g$ ends (e.g., a function's closing brace, a statement's semicolon). Oracles can be invoked only at these boundaries, since intermediate prefixes are typically syntactically incomplete.%

\emph{2. Renderer and artifacts.} Partial prefixes are typically syntactically incomplete (unclosed braces, unfinished statements), and different oracles expect different inputs (a compilable snippet for a compiler, an executable for a test runner). We therefore define a deterministic \emph{renderer}
\begin{equation}
\mathsf{Render}: (S, Y_{\le i}) \mapsto A_i \in \mathcal{A}
\label{eq:render}
\end{equation}
mapping a boundary-terminated prefix to an \emph{artifact} an oracle can process, where $\mathcal{A}$ is the artifact space equipped with a scope map $g: \mathcal{A} \to \mathcal{G}$. $\mathsf{Render}$ is defined only at \emph{structural boundaries}: prefixes ending at the closing token of some $g \in \mathcal{G}$.

\emph{3. Oracle interface.} Let $\mathcal{O} = \{O_1, \dots, O_J\}$ be the suite of verification oracles. Each oracle $O_j$ has an \emph{applicability predicate} $F_j: \mathcal{A} \to \{0, 1\}$ indicating whether it can be meaningfully invoked on $A$, and returns a deterministic verdict and diagnostics:
\begin{equation}
O_j(S, A) = (q_j, d_j), \quad q_j \in \{0, 1, \bot\},
\label{eq:oracle}
\end{equation}
where $1$, $0$, $\bot$ denote pass, fail, and not applicable respectively; $q_j = \bot$ iff $F_j(A) = 0$.
In our instantiations $F_j$ is typically a scope threshold $F_j(A) = \mathbf{1}[g(A) \succeq g_j^{\min}]$, but $F_j$ may depend on other features of $A$ (e.g., a differential-testing oracle requires the target function to have test inputs).%

\noindent{\bf Objective.}
A decoding strategy $\pi$ interleaves token generation under $M$ with oracle invocations on intermediate artifacts, and may rollback and regenerate. Let $A_m := \mathsf{Render}(S, Y_{\le m})$ be the terminal artifact, with terminal quality $Q(S, A_m) = 1$ iff all applicable evaluation oracles pass on $A_m$. Given a generation budget $B_{\mathrm{gen}}$ that hard-caps next-token model calls (counting tokens later discarded by rollback), \methodabbr{} targets
\begin{equation}
\max_{\pi} \quad \mathbb{E}\big[ Q(S, A_m) \big] \quad
\text{s.t.} \quad N_{\mathrm{gen}}(\pi) \le B_{\mathrm{gen}},
\label{eq:objective}
\end{equation}
where the expectation is over randomness in sampling from $p_M$ and any stochasticity in $\pi$. \S~\ref{sec:method} gives our concrete, deterministic instantiation of $\pi$.

\subsection{Core Mechanisms}
\label{sec:method}

We target the objective Eq.~\eqref{eq:objective} through three mechanisms: (i) \emph{structural-boundary verification} (\S~\ref{sec:dtv_overview}), which maintains the legal prefix invariant through boundary-aligned rendering and a unanimous-pass commit condition; (ii) \emph{structure-aware rollback with escalation}, which rolls back to the smallest scope and escalates to larger scopes (e.g., from a statement to its enclosing block or function) when the same diagnostic persists; and (iii) \emph{feedback via prompt augmentation}, which surfaces unresolved oracle failures back to the model so subsequent generation conditions on what went wrong.
Algorithm~\ref{alg:dtv} (Appendix~\ref{app:actions}) describes these mechanisms as a single loop: at each iteration the controller extends the prefix to the next structural boundary, runs applicable oracles on the rendered artifact, and either commits (on unanimous pass) or invokes feedback-augmented rollback at a chosen scope (on failure). The inner loop ends on candidate completion, on \emph{bailout} (repeated diagnostic-anchor visits reach the recovery limit and the current artifact is returned to the outer loop), or generation budget exhaustion. The common final gate judges the returned artifact.

\noindent{\bf Meta steps.}
Between structural boundaries, \methodabbr{} delegates token generation to the base sampler; at each boundary it pauses to invoke oracles and act on their verdicts. We call each pause a \emph{meta step} (indexed $k = 1, \dots, K$): let $b_k$ denote the prefix length at meta step $k$ (after any rollbacks), $A_k := \mathsf{Render}(S, Y_{\le b_k})$ the artifact, and $g_k := g(A_k)$ its scope. Oracles $O_j$ with $F_j(A_k) = 1$ are applicable at meta step $k$. Every committed checkpoint must satisfy the \emph{legal prefix invariant}: the committed prefix can be extended into a program that passes all oracles, so that rollback targets are always viable starting points. Enforcement is two-part: the renderer turns the current prefix into an oracle-consumable artifact, and Algorithm~\ref{alg:dtv} commits only when all oracles applicable at the current scope pass. The controller below decides which action to take at each meta step.%

\noindent{\bf Controller.}
The policy $\pi$ is instantiated as a deterministic state-machine controller. At each meta step $k$, it consumes an observation $\sigma_k$ and emits an operation $\omega_k$ that drives the next iteration. The observation aggregates the current prefix, the latest oracle outputs $o_k = \{(q_j, d_j) : F_j(A_k) = 1\}$, the feedback state $\mathcal{F}_k$, bounded retry and visit counters, and the remaining token budget. The op $\omega_k = (a_k, \ell_k, \mu_k, \beta_k)$ chooses one of six actions $a_k \in \{\textsc{Generate}, \textsc{Verify}, \textsc{Commit}, \textsc{Rollback}, \textsc{Feedback}, \textsc{Terminate}\}$, with optional fields: a rollback scope $\ell_k \in \mathcal{G}$, a retry mode $\mu_k$, and a bailout flag $\beta_k$. Between meta steps, token generation follows the base sampler unchanged (greedy, top-$k$, top-$p$). \methodabbr{} is a modular framework with reusable controller interfaces and task/model adapters (Appendix~\ref{app:impl}). Appendix~\ref{app:actions} gives precise semantics for each action.

\noindent{\bf Feedback.}
Rollback alone amounts to resampling from the same conditional distribution: although sampling diversity (e.g., varying temperature) can yield retries that differ from earlier attempts, exploring for the specific fix without diagnostic guidance is inefficient. %
The feedback loop therefore has two parts.%

\emph{1. Feedback state.} The feedback state $\mathcal{F}_k$ is the controller's working record of oracle failures not yet resolved, with each entry holding the diagnostics needed to guide repair (error message and source anchor). It feeds both the prompt augmentation step and the rollback/bailout policy. After observing $o_k$, the state is updated:
\begin{equation}
\mathcal{F}_{k+1} = \mathsf{Update}(\mathcal{F}_k, o_k).
\label{eq:update_feedback}
\end{equation}
$\mathsf{Update}$ is a deterministic state update that adds new failures, drops failures resolved by passing verdicts, and bounds the state's size.

\emph{2. Prompt augmentation.} We communicate $\mathcal{F}_{k+1}$ back to the model by augmenting the conditioning context. Let $\phi_{k+1} := \mathsf{Encode}(\mathcal{F}_{k+1})$ be a textual block summarizing the active diagnostics, and $\widetilde{S}_{k+1} := \mathsf{Augment}(S, \phi_{k+1})$ the augmented conditioning context. Between meta steps, $\widetilde{S}_{k+1}$ replaces $S$ in the conditioning context, so for any prefix $Y_{<i}$ the next-token distribution shifts from $p_M(\cdot \mid Y_{<i}, S)$ to $p_M(\cdot \mid Y_{<i}, \widetilde{S}_{k+1})$. Rollback-with-feedback is therefore \emph{not} blind resampling: decoding resumes under updated feedback while $S$ remains fixed. Two concrete retry modes (inline continuation, patch-based repair) are described below.

\noindent{\bf Rollback, escalation, and bailout.}
When verification fails, the controller rolls back the prefix at a scope $\ell \in \mathcal{G}$ chosen by a fixed escalation rule. Rollback starts at statement scope; when repeated retries yield the same diagnostic, the controller escalates to the enclosing block, or to the function if no block is open. Coarser scopes expose a larger region of code in which the model can revise its generation while preserving the retained context.

For example, consider a translation that must retain \texttt{message} for the later output:
\begin{lstlisting}[language=Rust,basicstyle=\ttfamily\footnotesize,aboveskip=3pt,belowskip=3pt]
fn consume(_: String) {}
fn show(message: String) {
    consume(message);       // Ownership moves here.
    println!("{message}");  // E0382 is reported here.
}
\end{lstlisting}
Cloning \texttt{message} at the later \texttt{println!} fails because ownership has already moved. Changing the earlier call to \texttt{consume(message.clone())} preserves \texttt{message} for the later output. Statement-only rollback cannot modify the earlier call, motivating scope escalation.

To avoid unbounded retries on unfixable errors, \methodabbr{} introduces two safeguards. \emph{Escalation}: the controller widens the rollback scope after repeated failures at finer scopes. For example, a Rust mutability error whose root cause lies several statements upstream may require escalation to block scope so the binding region itself can be regenerated (Appendix~\ref{app:rq3_traces}). \emph{Bailout}: when diagnostic-anchor visits reach a fixed threshold, the controller terminates the inner loop and returns the current artifact together with outstanding diagnostics to the outer loop.

In our implementation, a streaming-parser renderer maintains checkpoints over both the token prefix and structural parsing state. After rollback, retries proceed in one of two modes: \emph{inline continuation}, which appends diagnostics as comments and resumes decoding in the same turn, or \emph{patch-based repair}, which requests a minimal patch in a new turn (Appendix~\ref{app:feedback_examples} shows the prompt format and an example exchange for each mode). Fixed counters govern scope escalation, retry-mode changes, and bailout (Appendix~\ref{app:impl}).

\section{Experimental Setup}
\label{sec:experiments}

\noindent\textbf{Tasks and datasets.}
We evaluate \methodabbr{} on two complementary translation tasks:
\begin{squishenumerate}
\item \emph{C to Rust} (unsafe-to-safe systems translation): we sample \numCRustSamples{} C programs from CodeNet~\citep{puriCodeNetLargeScaleAI2021}, with test inputs for differential testing generated via AFL++ fuzzing with LLM-based seeding; we use \texttt{rustc} as oracle and differential testing for post-hoc behavioral equivalence.
\item \emph{JavaScript to TypeScript} (dynamic-to-static typing): we sample \numJSTSSamples{} self-bundled programs from TypeWeaver~\citep{yeeMachineLearningModels2023} that do not type-check as plain JavaScript, using \texttt{tsc} and ESLint's \texttt{@typescript-eslint/typedef} rule as oracles; executable tests are unavailable.
\end{squishenumerate}
\vspace{-2mm}
For comparisons with other baselines and the Gemma cross-model analysis, we use cost-matched subsets of $200$ (C to Rust) and $100$ (JS to TS), drawn from the same pools. Qwen3-30B-A3B and the mechanism comparisons use the first $100$ cases per relevant task. Per-task \methodabbr{} instantiation, sampling pipelines, and oracle filtering choices are in Appendix~\ref{app:experimental_setup}.

\noindent\textbf{Models.}
\methodabbr{} requires direct control over decoding (boundary stops, rollback, feedback injection); closed-weight inference APIs do not expose this control surface, so we use open-weight models throughout. Our primary translation model is Qwen3-4B-Instruct~\citep{qwenteamQwen3TechnicalReport2025}; we additionally use Gemma-4-E4B-it~\citep{farabetGemma4Byte2026} to test generalization across model families and Qwen3-30B-A3B-Instruct to evaluate a stronger generator. \methodabbr{} and the baselines share the same model and decoding configuration in every comparison; full hyperparameters are in Appendix~\ref{app:models}.

\noindent\textbf{Baselines and budget.}
We decompose test-time decoding strategies along two orthogonal axes. The \emph{inner loop} is either \textbf{na\"ive} (the outcome-only configuration of \S~\ref{sec:method}: complete generation in one pass with no in-loop verification) or \textbf{\methodabbr{}} (in-loop verification, rollback, and feedback; \S~\ref{sec:method}). The \emph{outer loop} is either \textbf{one-shot} (a single attempt), \textbf{self-refine} (sequential regeneration with diagnostic feedback when available), or \textbf{best-of-$N$ (BoN)} ($N$ independent attempts, returning the first oracle-passing candidate). We evaluate five inner$\times$outer combinations---na\"ive/one-shot, na\"ive/self-refine, na\"ive/BoN, \methodabbr{}/one-shot, \methodabbr{}/self-refine---plus S\textsuperscript{*}~\citep{liTestTimeScaling2025a}\footnote{Configured with independent attempts $N{=}8$, rounds $R{=}3$, and compile-error-based selection, deviating from the original defaults~\citep{liTestTimeScaling2025a} to fit the per-case token budget and fair comparison (clarified in Appendix~\ref{app:rq1_s_star_deviations}).} as a process-verified reference baseline combining parallel sampling with compile-driven self-debug.
All self-refine configurations share the same per-case generation budget of $\budgetK\times$ source tokens, counting all output tokens within a case.

\noindent\textbf{Metrics.}
The primary metric is the post-hoc compilation pass rate (\texttt{rustc} for C to Rust; \texttt{tsc} with strict mode disabled and ESLint zero-\texttt{typedef} for JS to TS). All methods share the unfiltered final gate; selected diagnostic filtering applies only to \methodabbr{}'s intermediate search (Appendix~\ref{app:experimental_setup}). On C to Rust we additionally report functional success rate via differential testing to verify that compile-pass gains do not mask semantic regressions (Appendix~\ref{app:rq1_functional}). We report the average per-case output tokens as the cost metric, supplemented by elapsed time and verifier calls. A shared \texttt{tsc --strict} recheck evaluates the saved JS-to-TS outputs under a stronger criterion (Appendix~\ref{app:rq1_strict}).

\section{Results}
\label{sec:results}

Through our evaluation, we aim to answer the following questions:
\begin{itemize}[nosep,leftmargin=*]
\item \textbf{RQ1.} Under equal per-case token budgets, does (i) \methodabbr{} perform better than baselines on pass rate versus token cost on both tasks, and (ii) the gain transfer to different generator model families?
\item \textbf{RQ2.} Do \methodabbr{}'s key mechanisms all contribute positively to its performance?
\item \textbf{RQ3.} (i) On which cases does \methodabbr{} outperform na\"ive, (ii) why does it do so, and (iii) how does this picture hold under multi-round self-refine?
\end{itemize}

\subsection{Comparing \methodabbr{} with baselines on pass rate and token efficiency}
\label{sec:rq1}

Under a fixed per-case generation budget of $\budgetK\times$ source tokens, \methodabbr{}/self-refine Pareto-dominates na\"ive/self-refine and lies in the upper-left region of the cost-vs-pass-rate frontier on both tasks (cumulative view in Figure~\ref{fig:rq1_scaling}; full-set Pareto view in Figure~\ref{fig:rq1_pareto_full} in Appendix~\ref{app:rq1_full_set}). It reaches \rqOneCrustDtvSrPass{} on C to Rust and \rqOneJstsDtvSrPass{} on JS to TS, exceeding na\"ive/self-refine by \rqOneCrustNaiveSrVsDtvSrDiffPp{}~pp and \rqOneJstsNaiveSrVsDtvSrDiffPp{}~pp.

\begin{figure}[!ht]
  \centering
  \includegraphics[width=\linewidth]{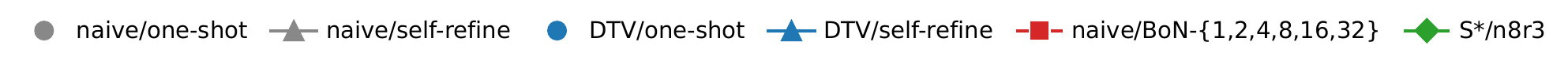}\\[-0.4em]
  \begin{subfigure}[t]{0.49\linewidth}
    \centering
    \includegraphics[width=\linewidth]{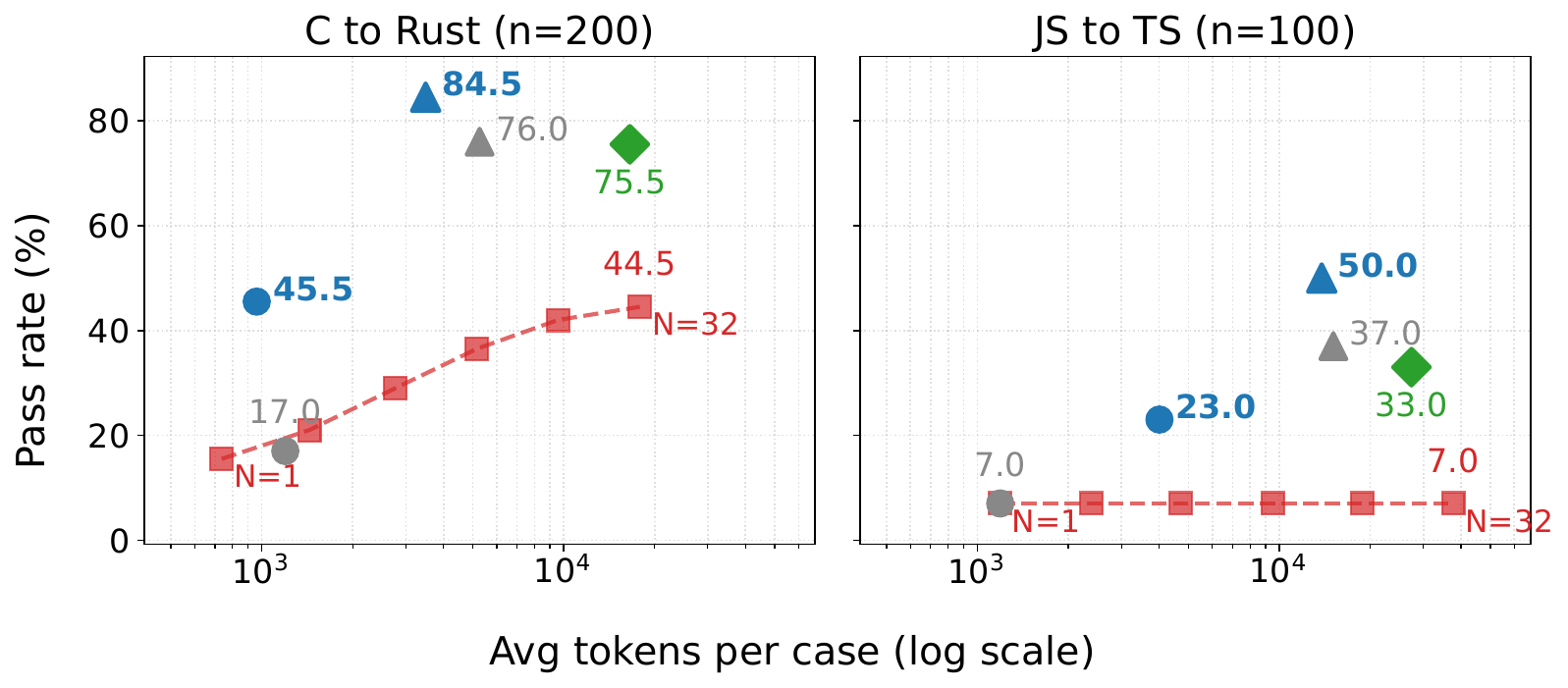}
    \caption{Per-case token cost (log scale) vs.\ compile pass rate, comparing against na\"ive, best-of-$N$  and S\textsuperscript{*}}
    \label{fig:rq1_baseline}
  \end{subfigure}
  \begin{subfigure}[t]{0.49\linewidth}
    \centering
    \includegraphics[width=\linewidth]{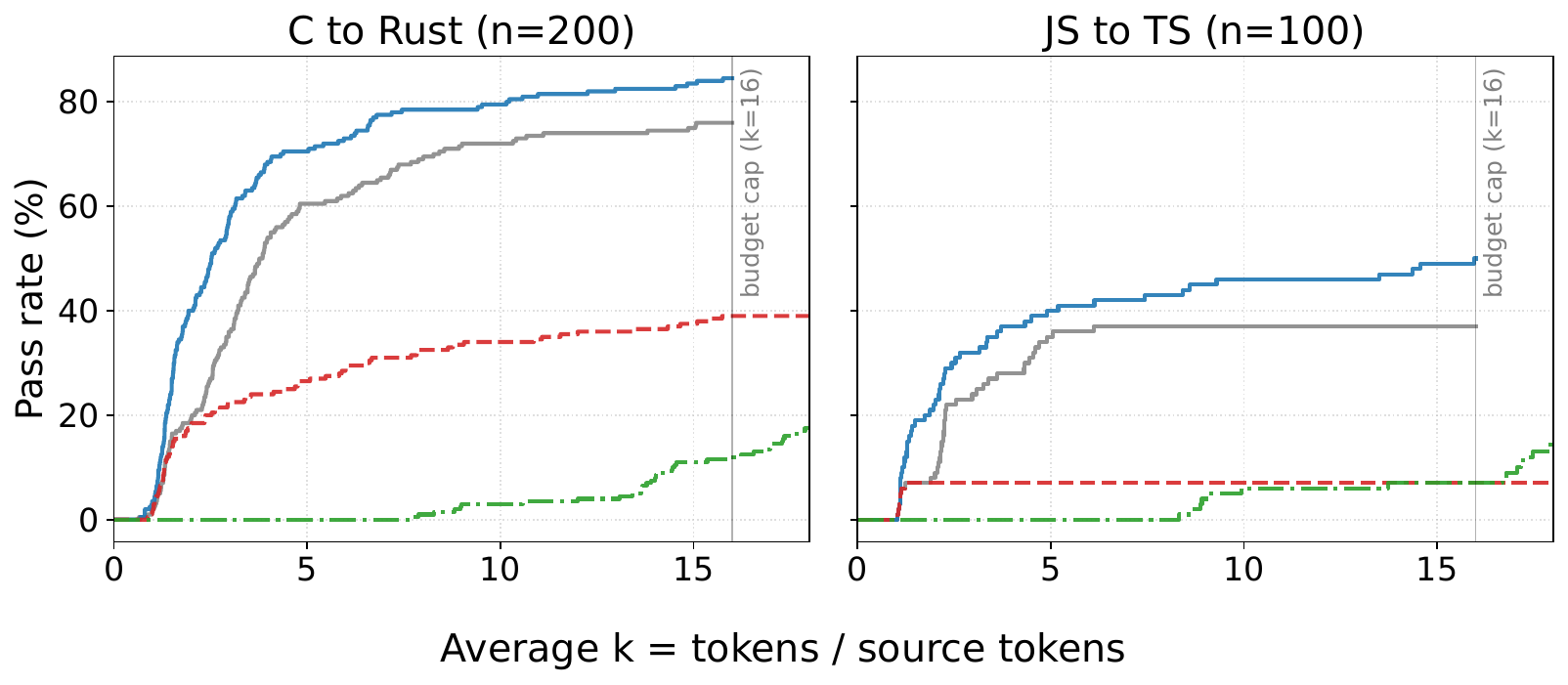}
    \caption{Cumulative pass rate vs.\ per-case token budget $k$}
    \label{fig:rq1_scaling}
  \end{subfigure}
  \caption{\methodabbr{}/self-refine Pareto-dominates na\"ive/self-refine and sits on the upper-left of the cost-pass-rate frontier on both tasks, above the BoN-$N$ and S\textsuperscript{*} frontiers at matched per-case token cost.}
  \label{fig:rq1}
\end{figure}

\methodabbr{}/self-refine is also more token-efficient than na\"ive/self-refine while attaining a higher pass rate: on C to Rust the two configurations average $k=\rqOneCrustDtvSrAvgK$ and $k=\rqOneCrustNaiveSrAvgK$ tokens per case respectively.

\noindent{\bf Cumulative scaling of \methodabbr{} vs.\ na\"ive.} The cumulative pass-rate curves over the per-case token budget $k$ (Figure~\ref{fig:rq1_scaling}) reveal two structural differences. First, \methodabbr{} exhibits an early-rise effect: its step curve begins climbing at substantially smaller $k$ than na\"ive/self-refine on both tasks, because in-loop verification and repair admit successful programs well before na\"ive's first complete sample is generated. Second, on JS to TS, the two regimes diverge in the long-run: na\"ive/self-refine plateaus once additional tokens cease to yield further passes, while \methodabbr{}/self-refine continues to rise. Once a generation has settled on an under-typed structure, further refinement attempts tend not to recover from it; in-loop type-annotation feedback, by contrast, redirects generation away from these basins.

\noindent{\bf Scaling through in-loop verification.} With a single attempt and no outer-loop retry, \methodabbr{} raises the pass rate from \rqOneCrustNaiveOneshotPass{} to \rqOneCrustDtvOneshotPass{} on C to Rust and from \rqOneJstsNaiveOneshotPass{} to \rqOneJstsDtvOneshotPass{} on JS to TS (Figure~\ref{fig:rq1_pareto_full} in Appendix~\ref{app:rq1_full_set}). Its in-attempt rollbacks and retries cost extra tokens, but the pass rate grows $2.9\times$ and $4.6\times$ while average tokens per case ($k$, generated tokens per source token) grow only $1.3\times$ and $2.5\times$ (Table~\ref{tab:rq1_main} in Appendix~\ref{app:rq1_main_table}). Adding self-refinement raises the pass rate further, to \rqOneCrustDtvSrPass{} and \rqOneJstsDtvSrPass{}, so \methodabbr{} complements rather than replaces outer-loop test-time scaling.

This compile-pass improvement is not accompanied by a regression in behavioral correctness: on C to Rust, \methodabbr{}/self-refine attains a full-functional success rate of \rqOneCrustDtvSrFuncFullPass{} compared with \rqOneCrustNaiveSrFuncFullPass{} for na\"ive/self-refine, a difference indistinguishable from zero under the paired test (Appendix~\ref{app:rq1_functional}). \methodabbr{}'s in-loop oracle on C to Rust is \texttt{rustc} alone, with differential testing reserved for post-hoc evaluation, so the controller receives no behavioral-equivalence signal during decoding; this guardrail finds no evidence of eroded behavioral correctness while optimizing compile pass.

The JS-to-TS gain is robust to a stricter final criterion: adding \texttt{tsc --strict} preserves \methodabbr{}'s pass-rate advantage across all evaluated models (Appendix~\ref{app:rq1_strict}).

\textbf{Cross-model robustness.} The same in-loop verification mechanism transfers to a second model family. We re-evaluate \methodabbr{}/self-refine against na\"ive/self-refine using \rqTwoGemmaName{} as the generator on both tasks (Appendix~\ref{app:rq2}). \methodabbr{} exceeds na\"ive at the budget cap on three of the four model-task pairs and reduces normalized cost on three. The exception is C to Rust on \rqTwoGemmaName{}, where na\"ive already reaches \rqTwoGemmaCrustNaivePass{} at the cap; under tight budgets ($k\le 7$), \methodabbr{}/self-refine still leads na\"ive on this cell with a peak advantage of about $+10$~pp at $k\approx 4$, and the curves cross near $k\approx 7$ where na\"ive's outer-loop retry over residual hard cases yields more marginal passes than \methodabbr{}'s verify-and-rollback.

The benefit extends to the stronger Qwen3-30B-A3B generator: final pass rates rise from \rqTwoQwenThirtyBCrustNaivePass{} to \rqTwoQwenThirtyBCrustDtvPass{} (C to Rust) and \rqTwoQwenThirtyBJstsNaivePass{} to \rqTwoQwenThirtyBJstsDtvPass{} (JS to TS), with gains already in the one-shot setting (Appendix~\ref{app:rq2_large}).

\paragraph{Comparing with other baselines.}

We additionally compare \methodabbr{}/self-refine against best-of-$N$ (BoN, $N{\in}\{1,2,4,8,16,32\}$) and S\textsuperscript{*} at matched per-case budget; \methodabbr{}/self-refine remains on the upper-left frontier against both on both tasks.
On C to Rust, BoN rises from \rqBaselineCrustNaiveBonOnePass{} at $N{=}1$ to \rqBaselineCrustNaiveBonThirtyTwoPass{} at $N{=}32$, but no $N$ reaches \methodabbr{}/self-refine's frontier: small $N$ ($N{\le}4$) is cheaper yet caps below \rqBaselineCrustNaiveBonFourPass{}; large $N$ ($N{\ge}8$) costs more yet trails by at least \rqBaselineCrustNaiveBonThirtyTwoVsDtvSrDiffPp{}~pp (Figure~\ref{fig:rq1_baseline}). On JS to TS, BoN stays at \rqBaselineJstsNaiveBonOnePass{} for all $N$ as cost grows, suggesting resampling repeats the same under-typed structure.

S\textsuperscript{*} reaches \rqBaselineCrustSstarNERrPass{} on C to Rust and \rqBaselineJstsSstarNERrPass{} on JS to TS, versus \methodabbr{}/self-refine's \rqBaselineCrustDtvSrPass{} and \rqBaselineJstsDtvSrPass{}, while consuming roughly $\fpeval{round(\rqBaselineCrustSstarNERrTokPerCase / \rqBaselineCrustDtvSrTokPerCase, 1)}\times$ \methodabbr{}'s per-case tokens on C to Rust and $\fpeval{round(\rqBaselineJstsSstarNERrTokPerCase / \rqBaselineJstsDtvSrTokPerCase, 1)}\times$ on JS to TS. \methodabbr{}'s advantage stems from verification timing: S\textsuperscript{*} compiles only after each completed attempt, whereas \methodabbr{} verifies at structural boundaries within a single attempt, converting the same signal into early localized rollback rather than full restart. The full baseline grid is in Appendix~\ref{app:rq1_baseline_table}.

\noindent{\bf Wall-clock tradeoff.} \methodabbr{} saves time with \texttt{rustc} but adds overhead with serial TypeScript/ESLint. From Qwen3-4B to Qwen3-30B-A3B, DTV/na\"ive time ratios improve from \rqOneQwenFourBCrustTimeRatio{} to \rqOneQwenThirtyBCrustTimeRatio{} and from \rqOneQwenFourBJstsTimeRatio{} to \rqOneQwenThirtyBJstsTimeRatio{}, respectively. This smaller relative overhead favors \methodabbr{} as generation becomes expensive relative to a fixed verifier stack. Table~\ref{tab:runtime} and Appendix~\ref{app:rq1_runtime} give timings and call counts.

\noindent \underline{\bf Takeaway.} In pass rate versus generated tokens, \methodabbr{}/self-refine Pareto-dominates na\"ive/self-refine and lies above all baseline frontiers on both tasks, with the gain transferring across model families.

\subsection{Ablation study on \methodabbr{}'s three key mechanisms}
\label{sec:rq3}

\begin{table}[H]
  \centering
  \caption{Ablations on \rqThreeCrustDtvSrNumTotal{} C-to-Rust cases. All four configurations share the self-refine outer loop and the same per-case token budget. $\Delta$ columns report the ablation minus \methodabbr{}-full.}
  \label{tab:rq3_main}
  \small
  \rqThreeAblationTable
\end{table}

We ablate each of \methodabbr{}'s three key mechanisms (\S~\ref{sec:method}) in turn, holding the remainder of \methodabbr{} (including the outer self-refine loop) fixed:
\vspace{-2mm}
\begin{itemize}[nosep,leftmargin=*]
  \item \emph{\methodabbr{}-no-feedback} (ablates feedback via prompt augmentation): rollback without injecting diagnostics during inner generation.
  \item \emph{\methodabbr{}-no-escalation} (ablates structure-aware rollback with escalation): always roll back to statement scope, never widening to block or function.
  \item \emph{\methodabbr{}-detect-and-abort} (disables in-loop recovery): terminate the inner loop on the first verification failure, deferring recovery to outer self-refine.
\end{itemize}
\vspace{-2mm}

\methodabbr{} with all three mechanisms enabled (hereafter \methodabbr{}-full) attains the highest pass rate among the four configurations, and disabling any of the three reduces pass rate (Table~\ref{tab:rq3_main}). \methodabbr{}-full reaches \rqThreeCrustDtvSrPass{} on the first \rqThreeCrustDtvSrNumTotal{} C-to-Rust cases. Removing inner-loop diagnostic feedback drops pass rate to \rqThreeCrustNoFeedbackPass{} (\rqThreeCrustDtvSrVsNoFeedbackDiffPp{}~pp), restricting rollback to statement scope drops it to \rqThreeCrustNoEscalationPass{} (\rqThreeCrustDtvSrVsNoEscalationDiffPp{}~pp), and disabling structure-aware inner recovery (detect-and-abort) drops it to \rqThreeCrustDetectAbortPass{} (\rqThreeCrustDtvSrVsDetectAbortDiffPp{}~pp).

\noindent{\bf Within an equal per-case token budget.} \methodabbr{}-full converts more of the budget into passing translations than the ablations: \rqThreeCrustDtvSrAvgTok{} avg tokens per case versus \rqThreeCrustNoFeedbackAvgTok{}, \rqThreeCrustNoEscalationAvgTok{}, and \rqThreeCrustDetectAbortAvgTok{} for the three ablations. The detect-and-abort ablation repeatedly delegates recovery to the outer self-refine loop: \rqThreeCrustDetectAbortAvgOuterRounds{} outer rounds versus \rqThreeCrustDtvSrAvgOuterRounds{} for \methodabbr{}-full, roughly $4{\times}$. \methodabbr{}-full achieves this by spending more tokens on each successful translation than the ablations: \rqThreeCrustDtvSrTokensPerPass{} per pass versus \rqThreeCrustNoFeedbackTokensPerPass{}, \rqThreeCrustNoEscalationTokensPerPass{}, and \rqThreeCrustDetectAbortTokensPerPass{} (detailed in Appendix~\ref{app:rq3_main_table}). The extra per-success tokens fund the in-loop verification and structured-recovery work that converts otherwise-failing cases into passes.

\noindent{\bf Differential-testing pass rates.} \methodabbr{}-full attains \rqThreeCrustDtvSrFuncPass{}, while no-feedback, no-escalation, and detect-and-abort reach only \rqThreeCrustNoFeedbackFuncPass{}, \rqThreeCrustNoEscalationFuncPass{}, and \rqThreeCrustDetectAbortFuncPass{} respectively. \methodabbr{}-full thus leads on this behavioral criterion as well. Appendix~\ref{app:rq3_paired} reports paired McNemar tests on this \rqThreeCrustDtvSrNumTotal{}-case ablation cohort.

\noindent{\bf The inner-1shot pass rate.} This decomposes each mechanism's effect between inner \methodabbr{} and outer self-refine. Removing diagnostic feedback changes inner-1shot by \rqThreeCrustDtvSrVsNoFeedbackInnerOneDiffPp{}~pp: without an explicit signal about what failed, the inner loop's resampling tends to reproduce the same error pattern, deferring recovery to outer self-refine. Disabling structured inner recovery (detect-and-abort) changes inner-1shot by \rqThreeCrustDtvSrVsDetectAbortInnerOneDiffPp{}~pp, the largest first-pass drop: every inner verification failure discards in-loop state and forces outer self-refine to retry from scratch with only post-hoc diagnostic context. Restricting rollback to statement scope leaves inner-1shot unchanged at \rqThreeCrustNoEscalationInnerOnePass{} but limits recovery on cases where a diagnostic detected at one statement requires repair at an earlier committed statement (e.g., rustc's ``cannot assign to immutable variable'' detected at a later mutation whose root cause is an earlier \texttt{let} binding, which statement-scope rollback cannot reach). \methodabbr{}-full's scope escalation does occasionally over-rollback and discard correctly-committed prefix in exchange; the net effect is \rqThreeCrustDtvSrVsNoEscalationDiffPp{}~pp final pass rate for no-escalation relative to full. Detailed analysis is in Appendix~\ref{app:rq3_traces}.

\noindent \underline{\bf Takeaway.} All three of \methodabbr{}'s mechanisms contribute, and removing any one reduces pass rate. In-loop recovery contributes the largest gain compared to diagnostic feedback and scope escalation.

\subsection{Deep analysis of \methodabbr{}'s gain over the na\"ive baseline}
\label{sec:rq4}

We take \methodabbr{}'s gain apart through three sub-questions: (i) on which cases \methodabbr{} outperforms na\"ive, (ii) why it does so, and (iii) whether the same picture holds under multi-round self-refine. The analysis uses the \numCRustSamples{} C-to-Rust and \numJSTSSamples{} JS-to-TS samples from \S~\ref{sec:rq1}.

\noindent{\bf What kind of error \methodabbr{} helps?}
To localize where \methodabbr{}'s gain comes from in the one-shot regime, we measure \methodabbr{}'s rescue rate (the fraction of na\"ive-failed cases that \methodabbr{} solves) separately for each type of fix the case requires. Lacking a ground-truth oracle for fix location, we use na\"ive's self-refine behavior as a proxy: when na\"ive's first generation (R1) fails, its second generation (R2) attempts to repair R1, and we take the lines R2 edits in R1 as an approximation of the necessary fix location. We label each case's \emph{fix shape}, the relation between R2's edited lines and R1's flagged error lines, as LOCAL (every edit lands on a flagged error line), NONLOCAL (none does), MIXED (partial overlap), or UNKNOWN (R2 rewrites more than 50\% of R1, excluded as uninformative); a more rigorous variant using na\"ive's final passing round $R_{\textrm{final}}$ in place of R2 yields the same conclusion on a smaller set of cases (Appendix~\ref{app:rq4_classifier}).
On both tasks, \methodabbr{}'s rescue rate decreases monotonically as the necessary fix moves further from the failing line, with the LOCAL-to-NONLOCAL gap reaching 20.4pp on C to Rust and 29.3pp on JS to TS (Table~\ref{tab:rq4_fixshape}; full breakdown and small-bin qualification in Appendix~\ref{app:rq4_fixshape}). An edit away from the diagnostic line may still lie within an enclosing block or function, which motivates scope escalation (\S~\ref{sec:rq3}); broader nonlocal restructuring remains harder.

\begin{figure}[!ht]
  \centering
  \begin{minipage}[b]{0.55\linewidth}
    \centering
    \resizebox{\linewidth}{!}{\rqFourFixshapeTableMain}
    \captionof{table}{\methodabbr{}/one-shot rescue rate on na\"ive R1-failure cases by fix shape (R1$\to$R2 diff). Cells show rate with sample number $n$; color tracks rate; $*$ marks $n<5$. Rescue rate decreases monotonically LOCAL$\to$NONLOCAL on both tasks. Full breakdown in Table~\ref{tab:rq4_fixshape_full} in Appendix~\ref{app:rq4_fixshape}.}
    \label{tab:rq4_fixshape}
  \end{minipage}\hfill
  \begin{minipage}[b]{0.43\linewidth}
    \centering
    \includegraphics[width=\linewidth]{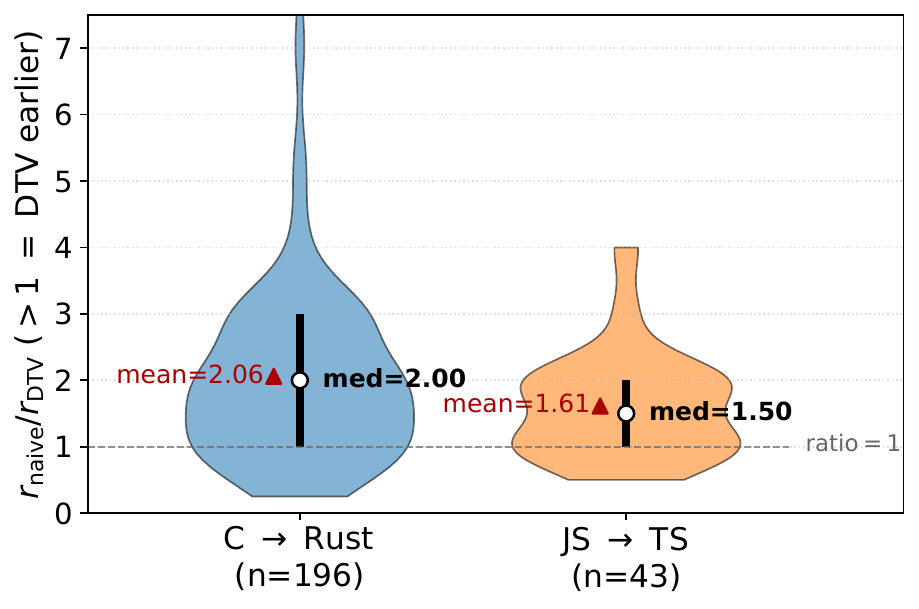}
    \captionof{figure}{Paired round-count ratios ($r_{\text{na\"ive}}/r_{\mathrm{\methodabbr{}}}$) on the both-pass subset of self-refine. Markers: median (white circle), mean (red triangle).}
    \label{fig:rq4_paired_rounds}
  \end{minipage}
\end{figure}

\noindent{\bf Why \methodabbr{} outperforms post-hoc verification?}
\methodabbr{} and na\"ive differ in when they invoke verification: \methodabbr{} intervenes
during generation, while na\"ive verifies only the completed program. This
timing matters because autoregressive language models condition each
token's distribution on the preceding prefix~\citep{aroraWhyExposureBias2022}: an
error introduced early biases subsequent tokens toward extending the
error rather than recovering from it, so intervening early should reduce
error accumulation compared to post-hoc verification.
To check whether
\methodabbr{}'s interventions reduce error counts, we restrict to error codes that
appear in both \methodabbr{}'s inner-loop trace and na\"ive's one-shot loop, and
on each rescued case compute the per-case shared-code edit-concentration
ratio: distinct \methodabbr{} inner-loop diagnostic emissions on the shared codes
divided by distinct na\"ive R1 emissions on the same codes
(Appendix~\ref{app:rq4_mechanism}); a ratio below 1 means \methodabbr{} emits
fewer per-shared-code diagnostics than na\"ive. On JS to TS the ratio is
well below 1 (median \rqFourJstsSharedRatioMedian{}, mean
\rqFourJstsSharedRatioMean{}, $n=\rqFourJstsSharedRatioN$): a missing
or wrong type annotation propagates to every downstream use site, so
catching it early prevents many derivative errors. On C to Rust the
ratio is near parity (median \rqFourCrustSharedRatioMedian{}, mean
\rqFourCrustSharedRatioMean{}, $n=\rqFourCrustSharedRatioN$); the
suppression effect is weaker because rustc's main error families
(borrow check, lifetime, type mismatch) do not compound across use
sites the way TypeScript type errors do.

\noindent{\bf How \methodabbr{} saves self-refine tokens?}
On cases both methods eventually pass (Figure~\ref{fig:rq4_paired_rounds}), \methodabbr{}'s earlier intervention translates into fewer self-refine rounds (median paired ratio of na\"ive's rounds to \methodabbr{}'s: \rqFourCrustSrPairedRatioMedian{} on C to Rust, \rqFourJstsSrPairedRatioMedian{} on JS to TS), and the round savings carry through to tokens on C to Rust (\methodabbr{} uses fewer tokens on \rqOneCrustNaiveSrVsDtvSrBpDtvBetterPct{} of cases, median paired difference \rqOneCrustNaiveSrVsDtvSrBpMedianTok{} tokens). On JS to TS the median still favors \methodabbr{} (\rqOneJstsNaiveSrVsDtvSrBpMedianTok{} tokens), but the mean reverses (\rqOneJstsNaiveSrVsDtvSrBpMeanTok{} tokens), driven by a small tail of nonlocal-fix cases (e.g., typedef errors whose correct annotation depends on usage elsewhere) where the inner loop re-emits the same diagnostic and rolls back without converging. A detailed breakdown is in Appendix~\ref{app:rq4_bothpass_tokens}.

\noindent \underline{\bf Takeaway.} \methodabbr{}'s gain concentrates on cases with local fixes while weakens on nonlocal ones, where early intervention prevents derivative errors and reduces self-refine rounds.

\section{Related Work}
\label{sec:related-work}

\methodabbr{} sits within test-time compute scaling, in the process-level supervision sub-family that emits step-level signals during generation \citep{snellScalingLLMTestTime2025, jiSurveyTestTimeCompute2025, brownLargeLanguageMonkeys2024, openaiOpenAIO1System2024, muennighoffS1SimpleTesttime2025a}; its distinguishing feature is that signals come from deterministic verifiers rather than learned scorers or repeated sampling.
Appendix~\ref{app:related-work} provides a complete discussion; Figure~\ref{fig:paradigms} in Appendix~\ref{app:paradigms} contrasts \methodabbr{} against the two dominant prior paradigms. We summarize the closest threads here.

\noindent{\bf Process supervision.}
Process reward models attach learned step-level scores and outperform outcome-only verifiers on multi-step reasoning \citep{lightmanLetsVerifyStep2024, uesatoSolvingMathWord2022, cobbeTrainingVerifiersSolve2021, wangMathShepherdVerifyReinforce2024, luoImproveMathematicalReasoning2024, khalifaProcessRewardModels2025}; SemGuard~\citep{wang2025semguard} similarly applies a learned line-level semantic evaluator during code generation and rolls back flagged lines.
\methodabbr{} produces step-level signals from \emph{deterministic oracles} instead, yielding exact rather than probabilistic verdicts, removing the need for annotation or training, and providing structured diagnostics to guide repair.

\noindent{\bf Verifier-guided generation.}
Outcome-only methods apply verifiers to completed programs for filtering, ranking, or single-shot repair \citep{liCompetitionlevelCodeGeneration2022, chenCodeTCodeGeneration2022, niLEVERLearningVerify2023, chenTeachingLargeLanguage2024, liTestTimeScaling2025a}.
Recent work on probabilistic programs of thought reuses token-level uncertainty to obtain cheap candidate variants before post-hoc verification~\citep{gargProbabilisticProgramsThought2026}.
In-loop methods place verifiers \emph{inside} generation: line-level Python execution feedback \citep{lavonExecutionGuidedLinebyLine2025}, function-level MCTS over Dafny/Coq partial programs \citep{brandfonbrenerVerMCTSSynthesizingMultiStep2024}, and verifier-driven tree-search refinement on top of a self-improving training loop for Dafny-to-Verus \citep{aggarwalAlphaVerusBootstrappingFormally2024a}.
\methodabbr{} differs in three respects: adaptive structural-boundary granularity (statement to program) rather than a fixed scope, hard state-machine control rather than stochastic search, and multi-oracle composition (compiler, type checker, linter).

ROCODE~\citep{jiang2025rocode} uses program-analysis-triggered backtracking, diagnostic-line or maximum-entropy-statement localization, and failed-path probability decay. \methodabbr{} instead combines nested structural repair regions with diagnostic-conditioned regeneration, preserving unaffected context. Appendix~\ref{app:repair_guidance} examines the roles of diagnostic guidance and editable repair scope.
Concurrent work, Hydra~\citep{du2026hydra}, couples an incremental compiler checker with checkpoint rollback to reduce the cost of reaching statically valid C/C++ code, while \methodabbr{} requires no verifier modification and raises pass rate under a fixed token budget.

\noindent{\bf Constrained decoding.}
Token-level recognizers (grammars, regexes, type automata) enforce hard syntactic constraints during sampling \citep{scholakPICARDParsingIncrementally2021, poesiaSynchromeshReliableCode2022, willardEfficientGuidedGeneration2023, beurer-kellnerPromptingProgrammingQuery2023, gengGrammarConstrainedDecodingStructured2023, agrawalMonitorGuidedDecodingCode2023, weiCopilotingCopilotsFusing2023, mundlerTypeConstrainedCodeGeneration2025, banerjeeCRANEReasoningConstrained2025, princisTreeCoderSystematicExploration2025}.
They cannot invoke semantic verifiers (full type checking, differential testing) on intermediate program scopes, since these require structurally complete artifacts.
\methodabbr{} is complementary: constrained decoding can enforce syntactic invariants while \methodabbr{} adds semantic verification on top.

\section{Conclusion}
\label{sec:conclusion}

Outcome-only test-time scaling for code applies deterministic verifiers (compilers, type checkers, tests) post-hoc, after errors have compounded the whole generated code. We introduce \emph{\methodname{}} (\methodabbr{}), a decoding framework that interleaves generation with verifier calls, combining structural-boundary verification, structure-aware rollback, and feedback via prompt augmentation. With Qwen3-4B, \methodabbr{}/self-refine raises compile pass rate from \rqOneCrustNaiveSrPass{} to \rqOneCrustDtvSrPass{} on C to Rust and from \rqOneJstsNaiveSrPass{} to \rqOneJstsDtvSrPass{} on JS to TS using fewer tokens per case than na\"ive/self-refine, Pareto-dominating na\"ive/self-refine and sitting above the best-of-$N$ and S\textsuperscript{*} frontiers at matched token cost; gains transfer to Gemma-4-E4B on three of four pairs. Additional experiments with Qwen3-30B-A3B show gains on both tasks, and shared strict rechecking preserves the JS-to-TS advantage. Key challenges remain: adapting recovery decisions to the remaining budget, extending the approach beyond translation, and reducing the implementation effort for new models and tasks (Appendix~\ref{sec:limitations}). We hope \methodabbr{} motivates future work on decoding-time control with deterministic oracles.

\begin{ack}
We thank the anonymous reviewers and the area chair for their valuable feedback on this paper.
This material is based upon work supported by the Defense Advanced Research Projects Agency (DARPA) under Agreement No. HR00112590131.
Approved for public release; distribution is unlimited.
This work used the Delta and DeltaAI systems at the National Center for Supercomputing Applications (awards OAC 2005572 and OAC 2320345) through allocation CIS251314 from the Advanced Cyberinfrastructure Coordination Ecosystem: Services \& Support (ACCESS) program~\citep{boerner2023access}, which is supported by U.S. National Science Foundation grants \#2138259, \#2138286, \#2138307, \#2137603, and \#2138296.
The authors declare no competing interests.
\end{ack}

\newpage
\bibliographystyle{unsrtnat}
\bibliography{references,DTV}

\newpage
\appendix
{\large \begin{center} {\bf APPENDIX} \end{center}}
\section{Limitations}
\label{sec:limitations}

\methodabbr{} rests on assumptions that bound its applicability and incurs non-trivial per-task and per-model implementation effort, both of which we document here.

\paragraph{Oracle requirements.}
\methodabbr{} requires oracles to produce \emph{local, actionable} diagnostics with source-line attribution (Section~\ref{sec:method}); oracles returning only a global pass/fail verdict can drive the commit decision but cannot guide structure-aware rollback or feedback. Our in-loop oracle suite consists only of compilation-style verifiers (\texttt{rustc} on C to Rust, \texttt{tsc} and ESLint on JS to TS); slower or weaker-signal oracles such as cross-language differential testing (Table~\ref{tab:instantiation}) are reserved for post-hoc evaluation, and whether \methodabbr{}'s in-loop mechanism remains effective when driven by such oracles is not addressed empirically.

\paragraph{Non-local semantic errors.}
Structure-aware rollback is most effective when a failure can be fixed within a bounded region. An upstream edit may still be reachable through block/function escalation, as the analyses in Section~\ref{sec:rq3} illustrate, but cross-module type inconsistencies or architectural redesign can require broader changes. The JS-to-TS inner oracle filters TS2322, TS2339, TS2551, and TS2345 when they are not actionable under partial context; all remain in the shared final gate. Gains persist under additional strict type checking (Appendix~\ref{app:rq1_strict}), while the absolute pass rates show that annotation coverage alone does not establish full type correctness.

\paragraph{Implementation and evaluation scope.}
\methodabbr{} requires language-specific rendering and context rules, verifier drivers and diagnostic filters, and model-family decoding adapters. Evaluation covers Qwen3-4B, Gemma-4-E4B, and the 30.5B-total-parameter Qwen3-30B-A3B MoE on two translation tasks. Gains with the stronger model expand the tested range, but do not establish generality across model scales, dense architectures, or non-translation domains. Near-saturation baseline performance can also limit the benefit of in-loop recovery, as seen on Gemma C to Rust.

\paragraph{Fixed recovery policy.}
The current policy uses fixed retry and bailout thresholds (Appendix~\ref{app:impl}). The ablations support recovery and scope widening, but do not show that these thresholds are optimal. Conditioning escalation and bailout on remaining budget and diagnostic type is a natural extension. The observed-edit analyses use successful-program diffs as localization proxies rather than identifying necessary or unique repairs.

\paragraph{Verifier wall-clock cost.}
Token matching controls generated-token expenditure rather than total runtime. Our measurements include verifier cost: C-to-Rust runs save wall-clock time, while the serial TypeScript/ESLint stack adds overhead (Table~\ref{tab:runtime}). Relative overhead decreases in the stronger-model experiment, consistent with a more favorable balance when generation is expensive relative to verification. Toolchain startup, code size, and CPU resources can change this balance, so token-Pareto improvements need not imply wall-clock improvements for every verifier stack.

\section{Broader Impacts}
\label{sec:broader_impacts}

\methodabbr{} is a decoding-time algorithm for code translation under deterministic
verifiers; it does not train new models or release new generative capabilities.
Its most direct positive impact is lowering the cost of translating legacy code
into safer target languages, for example C to Rust, where memory-safety bugs
account for a substantial fraction of reported security vulnerabilities, and
JavaScript to TypeScript, where static typing catches a class of latent defects
at development time. By converting outcome verifiers into process-level
guidance, \methodabbr{} also reduces the number of generation tokens spent on rejected
candidates relative to outcome-only decoding, which lowers compute and energy
cost per successful translation.

The most direct negative impact is the risk of misplaced confidence in
oracle-passing translations. \methodabbr{}'s \textsc{Commit} decision certifies only
that the committed prefix passes the configured verifier suite (compilation,
type checking, and differential testing on a finite set of inputs). It does
not certify behavioral equivalence on all inputs, freedom from undefined
behavior, concurrency correctness, or fitness for any particular
safety-critical use. Users who deploy \methodabbr{}-translated code without independent
review, especially in security-sensitive or safety-critical settings, may be
exposed to defects outside the oracles' coverage. We recommend treating \methodabbr{}
outputs as a starting point for human review and co-designing the verifier
suite with the deployment risk profile (e.g., adding fuzzing, sanitizers, or
formal verification for high-assurance settings).

\section{Verification Paradigms}
\label{app:paradigms}

\begin{figure}[t]
\centering
\includegraphics[width=\textwidth]{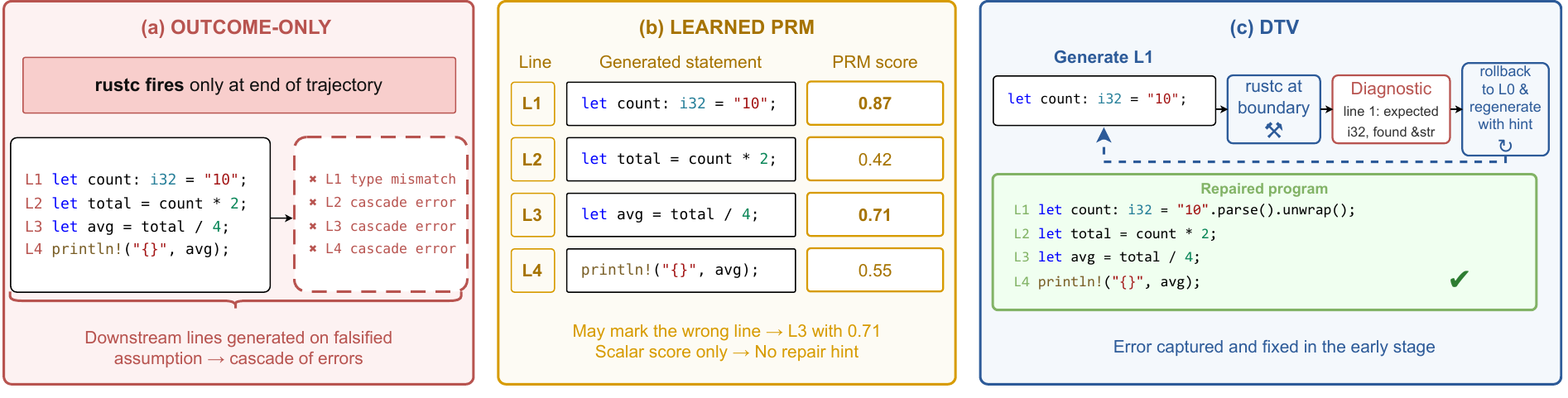}
\caption{Three verification paradigms on an illustrative C-to-Rust translation. \textbf{(a)} Outcome-only verifies after completion; \textbf{(b)} a learned PRM produces scalar ranking signals; \textbf{(c)} \methodabbr{} verifies at structural boundaries, rolls back, and regenerates with diagnostic feedback.}
\label{fig:paradigms}
\end{figure}

Figure~\ref{fig:paradigms} situates \methodabbr{} against the two paradigms that dominate prior work on verifier-guided generation. Outcome-only verification (panel a) applies verifiers post-hoc to completed samples, so a failing verdict discards an entire program-length sequence and the model has already extended downstream code on top of the falsified prefix. Learned process supervision (panel b) attaches probabilistic scores to intermediate steps, but a high score cannot certify that a partial translation can be extended to pass the compiler, type checker, and differential tests, and training such scorers requires verifier-, language-, and source-target-specific supervision. \methodabbr{} (panel c) integrates deterministic verifiers directly into the autoregressive loop, combining structural-boundary verification, structure-aware rollback, and feedback re-injection. Appendix~\ref{app:related-work} discusses the closest prior work in each paradigm.

\section{Related Work (Extended)}
\label{app:related-work}

This appendix expands the compressed discussion in \S\ref{sec:related-work} with a per-system breakdown.

\paragraph{Test-time compute scaling.}
Test-time compute scaling has emerged as a complement to parameter scaling: extra inference compute can substitute for larger models.
Existing approaches differ along two axes: the \emph{signal source} used to allocate compute (repeated independent sampling~\citep{brownLargeLanguageMonkeys2024}, compute-optimal allocation against learned process reward models~\citep{snellScalingLLMTestTime2025}, deliberate search over reasoning traces~\citep{yaoTreeThoughtsDeliberate2023a}, and end-to-end long-reasoning systems~\citep{openaiOpenAIO1System2024,muennighoffS1SimpleTesttime2025a}; surveys catalog these strategies~\citep{jiSurveyTestTimeCompute2025}), and the \emph{decoding integration} of those signals (outcome-only after a complete program, or process-level inside generation).
\methodabbr{} is itself a test-time compute scaling method, in the \emph{process-level supervision} sub-family, with the distinguishing feature that signals come from deterministic verifiers (compilers, type checkers, tests) integrated at structural boundaries during decoding. Probabilistic programs of thought (PPoT)~\citep{gargProbabilisticProgramsThought2026} are also motivated by fixed-budget inference: they reuse next-token probabilities from completed LLM samples to draw cheap local variants before post-hoc verification, whereas \methodabbr{} changes when verifier diagnostics affect the search by committing, rolling back, and repairing prefixes.
The rest of this section compares \methodabbr{} against the closest prior work in each adjacent corner: learned process scorers, execution-guided and verifier-guided generation (post-hoc and in-loop), and constrained decoding.

\paragraph{Process supervision via learned scorers.}
Process reward models (PRMs) train a learned scorer on step-level annotations and outperform outcome-only verifiers on multi-step reasoning~\citep{lightmanLetsVerifyStep2024,uesatoSolvingMathWord2022,cobbeTrainingVerifiersSolve2021}.
Recent work reduces annotation cost via automated step supervision~\citep{wangMathShepherdVerifyReinforce2024,luoImproveMathematicalReasoning2024}, while \citet{khalifaProcessRewardModels2025} replace discriminative PRMs with chain-of-thought verifiers. Reinforced Agent~\citep{taReinforcedAgentInferenceTime2026} similarly moves feedback before tool execution, its signal is a reviewer LLM over complete tool-call actions rather than a deterministic verifier over rendered code artifacts.
\methodabbr{} produces step-level signals as well, but obtains them from \emph{deterministic oracles} rather than learned scorers.
This shifts the design space along three axes: signals are \emph{exact} rather than \emph{probabilistic}, enabling hard \textsc{Commit} and \textsc{Rollback} actions instead of soft re-ranking; they require \emph{no annotation or training data}; and they come with \emph{structured diagnostics} (error messages, failing locations) that \methodabbr{} uses for repair---none of which is naturally available from a learned PRM.

\paragraph{Execution-guided and verifier-guided generation.}
One line of work uses deterministic verifiers (compilers, tests, formal solvers) to evaluate complete programs after generation, using the verdict for filtering, ranking, or single-shot repair~\citep{liCompetitionlevelCodeGeneration2022,chenCodeTCodeGeneration2022,niLEVERLearningVerify2023,chenTeachingLargeLanguage2024,liTestTimeScaling2025a}.
A more recent line moves the verifier \emph{inside} generation: \citet{lavonExecutionGuidedLinebyLine2025} run line-by-line Python execution feedback through Classifier-Free Guidance (EG-CFG); \citet{brandfonbrenerVerMCTSSynthesizingMultiStep2024} use MCTS guided by partial-program verification in Dafny and Coq (VerMCTS), with a token-budget metric (pass@T) closely related to ours; and \citet{aggarwalAlphaVerusBootstrappingFormally2024a} add verifier-driven tree-search refinement (Treefinement) for translating Dafny to Verus (AlphaVerus).
\methodabbr{} differs from these in-loop methods in five respects: (i) \emph{adaptive structural-boundary granularity} spanning statement, block, function, and program, rather than a single fixed scope (line in EG-CFG, function in VerMCTS), motivated by the fact that partial programs in static-typed mainstream languages admit only specific oracle-consumable scopes; (ii) \emph{hard state-machine control} with explicit \textsc{Commit}, \textsc{Rollback}, and \textsc{Feedback} actions, rather than soft probability interpolation, stochastic search, or whole-program restart; (iii) \emph{multi-oracle composition} (compiler diagnostics, type checker, differential testing) rather than a single signal type; (iv) \emph{decoding-time only} with a frozen pretrained model, while AlphaVerus involves a self-improving training loop; and (v) \emph{structured diagnostic-driven repair} via the explicit \textsc{Feedback} action and its patch-based retry mode, which consume oracle diagnostics and failing locations to construct targeted repair prompts, rather than passing raw execution traces as soft prompt context (EG-CFG) or discarding the failure content beyond a scalar pass/fail signal (VerMCTS, AlphaVerus).

\paragraph{ROCODE and source-conditioned repair.}
ROCODE~\citep{jiang2025rocode} establishes program-analysis-triggered backtracking during code synthesis. For a new located error it rolls back to the reported line; repeated or unlocated errors invoke a global maximum-token-entropy statement rule, while persistent syntax errors can trigger full restart. Its constrained regeneration stores failed paths in a trie and applies position-dependent probability decay along those paths. Analyzer diagnostics determine whether and where to backtrack, but their text is not added to the model's regeneration context.

\methodabbr{} shares the in-loop premise and addresses context preservation through explicit structural repair regions. A source program and retained target prefix continue to condition retries, so discouraging one token path does not itself specify the target-language constraint to repair. \methodabbr{} supplies diagnostic text and widens the editable region when local retry stalls. The ownership example in \S~\ref{sec:method} illustrates why a later diagnostic may require reopening an earlier decision; Appendix~\ref{app:rq3_regeneration} compares diagnostic conditioning with ROCODE path decay, and Appendix~\ref{app:rq3_localization} evaluates its entropy fallback on observed upstream edits.

The released ROCODE adapter makes newline-terminated Python prefixes analyzable by inserting \texttt{pass}/\texttt{break} and ignoring selected EOF failures. Its paper also reports C++ results, but the corresponding partial-artifact construction and Rust/TypeScript integrations are not provided in the released artifact. A cross-language port therefore requires additional rendering and compiler-driver choices. The mechanism analyses in Appendix~\ref{app:repair_guidance} use a common Rust renderer and oracle to isolate regeneration and localization choices from language-adapter differences.

Concurrent work, Hydra~\citep{du2026hydra}, targets the cost of repairing static errors in C/C++ code generation. It streams generated code into a modified Clang front end that checks prefixes asynchronously and checkpoints checker state by forking the compiler process, and it selects rollback points with a root-cause prior fitted on a training split of its benchmark. Relative to post-hoc repair, Hydra reduces latency and token consumption, while both reach near-complete static correctness.
\methodabbr{} differs in two respects. First, it requires no verifier modification. Hydra's retrofit relies on a checker that performs semantic analysis during parsing, as Clang does. \methodabbr{} instead runs unmodified verifiers on rendered artifacts, including non-compiler oracles such as ESLint. This matters for \texttt{rustc}, which parses the whole crate before type checking function bodies and borrow checking their MIR: a streaming front end would not reach these checks until the input is complete, whereas rendering exposes them at statement boundaries. Second, it targets pass rate under a fixed per-case token budget rather than the cost of reaching an outcome that post-hoc repair also attains. Its baselines remain far from saturated (with Qwen3-4B, na\"ive/self-refine reaches \rqOneCrustNaiveSrPass{} on C to Rust and \rqOneJstsNaiveSrPass{} on JS to TS), so in-loop verification yields higher pass rate at matched cost rather than only lower cost.

\paragraph{Constrained decoding for code.}
Constrained decoding enforces \emph{hard} token-level constraints derived from a recognizer (grammar, regex, finite-state machine, or type automaton) that operates over partial token sequences, masking tokens that would violate the recognizer at each sampling step~\citep{scholakPICARDParsingIncrementally2021,poesiaSynchromeshReliableCode2022,willardEfficientGuidedGeneration2023,beurer-kellnerPromptingProgrammingQuery2023,gengGrammarConstrainedDecodingStructured2023}.
Recent code-aware variants extend this in several directions: \citet{agrawalMonitorGuidedDecodingCode2023} drive decoding with repository-level static analysis; \citet{weiCopilotingCopilotsFusing2023} prune tokens via a completion engine; \citet{mundlerTypeConstrainedCodeGeneration2025} introduce prefix automata over inhabitable types; and \citet{banerjeeCRANEReasoningConstrained2025} augment the grammar to preserve reasoning capacity.
More recent framework work generalizes this with pluggable search strategies and composable constraints: \citet{princisTreeCoderSystematicExploration2025} unify sampling, beam search, MCTS, SMC, and ASAp under a tree-search abstraction that admits both token-level syntactic constraints and whole-program execution constraints (e.g., unit tests) via a product-of-experts formulation.
By construction, these methods enforce constraints either as recognizers over partial token sequences (typically syntax or local typing) or as post-hoc filters on completed programs; they cannot invoke semantic verifiers at intermediate program scopes during generation.
\methodabbr{} is orthogonal in two respects.
First, it accepts oracles as black boxes, including those that require structurally complete artifacts and therefore cannot be expressed as token-level recognizers (compilation against the full type and borrow checker, differential testing against the source program).
Second, it does not constrain the base sampler; verification failures are handled \emph{after the fact} by structure-aware rollback and diagnostic feedback, rather than prevented up front.
The two approaches compose: constrained decoding can enforce hard syntactic invariants while \methodabbr{} adds a semantic verification layer on top, a combination we leave to future work.

In summary, \methodabbr{} sits within test-time compute scaling and process-level supervision and shares the in-loop verification premise of ROCODE, EG-CFG, and VerMCTS. Its architectural commitments are deterministic verifier signals (vs.\ learned PRMs), adaptive structural-boundary granularity with explicit checkpoint rollback (vs.\ fixed-scope or whole-program in-loop methods), nested scope escalation with diagnostic-conditioned regeneration (vs.\ ROCODE's line- or entropy-selected rollback with failed-path decay), and a hard state-machine controller composing multiple oracles (vs.\ soft re-ranking, syntactic recognizers, or training-time verifier-feedback pipelines).

\section{Task and Oracle Setup}
\label{app:experimental_setup}

This appendix expands the per-task descriptions, dataset construction pipelines, and oracle configurations summarized in Section~\ref{sec:experiments}.

\paragraph{C to Rust.}
A representative unsafe-to-safe systems translation task. C programs routinely rely on manual memory management and undefined behavior that must be resolved into explicit Rust idioms (ownership, borrowing, lifetimes), making correctness verification non-trivial and the task widely studied in the program-analysis community~\citep{zhouSACTORLLMDrivenCorrect2025}.
The Rust compiler (\texttt{rustc}) provides a strong static oracle spanning syntactic and semantic analysis (types, borrows, lifetimes), and differential testing, executing both the C source and the Rust translation on the same inputs and comparing outputs, provides a behavioral equivalence oracle.
These oracles complement each other: \texttt{rustc} supplies static signals on partial programs, while differential testing supplies behavioral signals on complete programs.

\paragraph{JavaScript to TypeScript.}
A representative dynamic-to-static typing task. The translation is primarily about introducing sound type annotations: when an explicit annotation is missing and the compiler's inference cannot resolve a type, TypeScript silently falls back to \emph{implicit any}, which passes compilation but bypasses the type system and exposes the program to the type-related safety issues that migration was meant to eliminate~\citep{biermanUnderstandingTypeScript2014,hellendoornDeepLearningType2018}. Catching such degradation requires verifiers beyond the compiler alone.
Two verifiers are used: the TypeScript compiler (\texttt{tsc}) for compilation checks, and ESLint for detecting missing type annotations.
Unlike C to Rust, TypeWeaver packages do not ship with executable test suites, so differential testing is not available; the compiler and linter serve as the sole correctness oracles.

\paragraph{Component instantiation.}
Table~\ref{tab:instantiation} summarizes how the abstract \methodabbr{} components (Section~\ref{sec:method}) are instantiated for each task. The two tasks span complementary verification axes: unsafe-to-safe (C to Rust) versus dynamic-to-static (JS to TS), and compiler-plus-executable-test versus compiler-plus-linter oracle suites.

\begin{table}[h]
\centering
\caption{Instantiation of \methodabbr{} components for each translation task. The in-loop oracle drives \methodabbr{}'s in-generation verification, rollback, and feedback. The outer-loop oracle judges pass/fail at the end of each generation attempt for self-refine and best-of-$N$. The post-hoc oracle is used only for final evaluation.}
\label{tab:instantiation}
\small
\begin{tabular}{lll}
\toprule
\textbf{Component} & \textbf{C to Rust} & \textbf{JS to TS} \\
\midrule
Renderer          & Rust structural completion          & TS delimiter + context rules \\
In-loop oracle    & \texttt{rustc}                       & \texttt{tsc} (weakened) + ESLint \\
Outer-loop oracle & \texttt{rustc}                       & \texttt{tsc} (strict off) + ESLint \\
Post-hoc oracle   & \texttt{rustc}; differential testing & \texttt{tsc} (strict off) + ESLint \\
\bottomrule
\end{tabular}
\end{table}

\paragraph{C to Rust dataset and test-input generation.}
We construct the \numCRustSamples-sample evaluation set in three stages, all using fixed random seeds.
\emph{Stage 1: pool construction.} We filter Project CodeNet~\citep{puriCodeNetLargeScaleAI2021} to C submissions with status \texttt{Accepted} and retain only those whose source contains a standard-input call (\texttt{scanf}, \texttt{fgets}, \texttt{getchar}, etc.), since differential testing requires programs that consume external inputs. For each problem we select the submission with median code size as its representative, ensuring at most one program per problem.
\emph{Stage 2: stratified sampling.} We stratify the per-problem representatives by source byte size into three tiers (small ${<}300$ bytes, medium $300$--$1000$ bytes, large ${>}1000$ bytes) and sample proportionally across tiers with a minimum of $10$ representatives per tier, yielding $400$ candidate programs.
\emph{Stage 3: test-input generation.} CodeNet does not provide executable test inputs, so we generate them per candidate via a hybrid LLM-seeded AFL++ pipeline.
We first prompt GPT-5.4 for $25$ stdin test inputs per program, with instructions to include boundary values and ASCII-only content, and validate each candidate by executing the AFL-instrumented binary; seeds that exit non-zero or exceed a $5$-second timeout are discarded.
The validated seeds bootstrap AFL++ fuzzing for $90$ seconds in stdin mode.
The resulting AFL++ queue is minimized with \texttt{afl-cmin} and further filtered through AddressSanitizer, UndefinedBehaviorSanitizer, and Valgrind \texttt{memcheck} to drop inputs that trigger crashes, undefined behavior, or memory errors; surviving inputs form the final test corpus per program.
The pipeline succeeds on $306$ of the $400$ candidates; the remaining $94$ fail due to compilation errors, lack of valid LLM seeds after execution filtering, AFL++ producing no queue entries, or no inputs surviving the sanitizer filter.
We draw $\numCRustSamples$ cases uniformly at random from these $306$ as the final evaluation set.
Coverage is measured by re-executing each surviving input against a coverage-instrumented \texttt{clang} build and reported via \texttt{llvm-cov} source-based coverage.
The resulting fuzz corpora attain mean per-sample line coverage of $93.4\%$ (aggregate $85.1\%$, $11{,}721$ of $13{,}781$ lines) and mean per-sample branch coverage of $87.5\%$ (aggregate $78.9\%$, $6{,}838$ of $8{,}662$ branches).
The selected programs span $16$--$448$ lines of code, with median $46$ and mean $68.8$.

\paragraph{JavaScript to TypeScript dataset.}
We construct the \numJSTSSamples-sample evaluation set from TypeWeaver~\citep{yeeMachineLearningModels2023}.
We start from the two subsets \texttt{top1k-typed-nodeps-es6} and \texttt{top1k-untyped-nodeps-es6} and process each candidate package as follows.
For each package we produce a single bundled JavaScript file via \texttt{rollup} (ES module format on the package's main entry).
We discard packages whose bundled size falls outside $[30, 1000]$ lines of code, and we exclude packages that already pass \texttt{tsc} under strict mode without modification, since these provide no translation signal.
From the resulting pool we sample $\numJSTSSamples$ packages uniformly at random (seed $42$).
The selected packages span $30$--$741$ bundled lines of code, with median $91$ and mean $149.2$.

\paragraph{Oracle diagnostic filtering.}
All three in-generation verifiers (\texttt{rustc}, \texttt{tsc}, and ESLint) filter their raw diagnostic output before it drives rollback or feedback in \methodabbr{}. We distinguish two categories of filtering.

\emph{Partial-context filters.}
Partial-code checks can report incompleteness, such as missing definitions or closing syntax, before generation supplies the remaining context. The intermediate oracle drivers therefore apply the following filters:
\begin{itemize}[nosep,leftmargin=*]
\item \textbf{rustc (C to Rust):} suppresses unresolved type, value/function, and module diagnostics (E0412, E0425, E0433) in the partial-program oracle. It also suppresses E0277 trait-bound diagnostics with an attached replacement suggestion or an \texttt{impl} header, allowing later generated declarations to supply missing context.
\item \textbf{tsc (JS to TS):} suppresses selected missing-name, missing-module, and initialization diagnostics (TS2304, TS2552, TS2307, TS2564) while checking partial translations.
\item \textbf{ESLint (JS to TS):} drops diagnostics whose source position lies past the current prefix boundary, which can arise because our rendered artifact includes context-rule additions (e.g., a closing \texttt{\}}) beyond the actual generated prefix.
\end{itemize}

\emph{TypeScript settings and inner type-error filtering.}
The JS-to-TS instantiation uses the following compiler settings:
\begin{itemize}[nosep,leftmargin=*]
\item \textbf{Shared strictness setting:} both inner checks and the common final gate invoke tsc with \texttt{strict: false}, suppressing strictness-family diagnostics (including \texttt{noImplicitAny}).
\item \textbf{Type-correctness error filter:} TS2322 (assignment type mismatch), TS2339 and TS2551 (missing property, with or without a suggested alternative), and TS2345 (argument type mismatch) are dropped before the inner verdict is computed.
\end{itemize}
These errors are common in partial translations but are typically not actionable for in-loop retry: they often reflect missing context about external types or deliberate typing choices that become resolvable only once the full program is available. Retaining them would cause \methodabbr{} to spend retry budget on diagnostics the model cannot productively fix. We therefore filter them during generation only; the outer-loop and post-hoc tsc oracles preserve them and let them count against the verdict.

The shared final gate invokes the compiler and linter directly, retaining all of these error classes for every method. It uses \texttt{strict: false} and the ESLint \texttt{@typescript-eslint/typedef} rule to require explicit annotations. Annotation presence and strict type checking are complementary criteria; the latter also checks additional consistency and strictness properties. Appendix~\ref{app:rq1_strict} therefore reports an additional shared strict recheck of saved outputs.

The filters determine when the controller receives actionable process feedback; they do not relax the common final acceptance criterion. Final C-to-Rust acceptance similarly uses unfiltered \texttt{rustc}, with differential testing reserved for behavioral evaluation.

\paragraph{Baseline configuration details.}
\label{app:par_baseline_config}\label{app:rq1_s_star_deviations}
S\textsuperscript{*}~\citep{liTestTimeScaling2025a} is configured with $N{=}8$ parallel candidates, $R{=}3$ compile-feedback rounds per candidate, and compile-error argmin selection (the candidate whose final round compiles with the fewest errors is returned, breaking ties by sample order). Both choices deviate from the original defaults ($N{=}16$, $R{=}2$, adaptive input synthesis) for two reasons. First, $N{=}8, R{=}3$ replaces $N{=}16, R{=}2$ because $N{=}16$ would exceed the per-case token cap under which \methodabbr{} and na\"ive/BoN-32 operate; the resulting per-case generation budget of $N\times R = 24$ (versus the paper's default of $32$) aligns with the envelope shared by the other baselines. Second, compile-error argmin selection replaces adaptive input synthesis because the original mechanism prompts an auxiliary LLM to synthesize distinguishing test inputs and selects the candidate whose execution outputs align with the LLM's predicted behavior, a procedure that presupposes the availability of program-level test inputs in the task specification, which translation tasks (C-to-Rust, JS-to-TS) do not provide. We therefore adopt the ``Public Only'' selection variant reported in Table~3 of \citet{liTestTimeScaling2025a}, which selects the candidate whose final round compiles with the fewest errors and breaks ties by sample order. All methods share the final compiler/linter acceptance criterion; the intermediate diagnostic filtering described above is specific to \methodabbr{}'s search.
Because BoN-32 and S\textsuperscript{*} are substantially more computation-intensive than the head-to-head configurations, the BoN grid, S\textsuperscript{*} comparisons, and Gemma cross-model analysis (Appendix~\ref{app:rq2}) use $n=200$ for C to Rust and $n=100$ for JS to TS. Qwen3-30B-A3B uses the first 100 cases per task; the ablations and mechanism comparisons use the first 100 C-to-Rust cases. These subsets all come from the main evaluation pools.

\section{Model and Decoding Hyperparameters}
\label{app:models}

All experiments are run locally via the HuggingFace Transformers library.
We evaluate the following open-weight instruction-tuned models:
\begin{itemize}[nosep,leftmargin=*]
\item \textbf{Qwen3-4B-Instruct} (\texttt{Qwen/Qwen3-4B-Instruct-2507})~\citep{qwenteamQwen3TechnicalReport2025}.
\item \textbf{Gemma-4-E4B-it} (\texttt{google/gemma-4-E4B-it})~\citep{farabetGemma4Byte2026}.
\item \textbf{Qwen3-30B-A3B-Instruct} (\texttt{Qwen/Qwen3-30B-A3B-Instruct-2507}), a 30.5B-total-parameter mixture-of-experts model, at revision \texttt{0d7cf23991f47feeb3a57ecb4c9cee8ea4a17bfe}~\citep{qwenteamQwen3TechnicalReport2025}.
\end{itemize}
All models are used as black-box generators (no fine-tuning).
\methodabbr{} and the na\"ive baseline share the same model and decoding configuration within a given experiment, so any pass-rate difference isolates the effect of the decoding strategy rather than the underlying sampler.

\paragraph{Sampling.}
Between meta steps we use each model's default sampler as shipped in its \texttt{generation\_config.json}, without overriding sampling hyperparameters.
Table~\ref{tab:sampler_defaults} lists the values for all evaluated models.
A greedy decoding mode (\texttt{do\_sample=False}) is exposed as an ablation option but is not used in the main experiments.

\begin{table}[h]
\centering
\caption{Default sampler hyperparameters shipped with each model's \texttt{generation\_config.json}.}
\label{tab:sampler_defaults}
\small
\begin{tabular}{lcccc}
\toprule
\textbf{Model} & \texttt{do\_sample} & \texttt{temperature} & \texttt{top\_k} & \texttt{top\_p} \\
\midrule
Qwen3-4B-Instruct-2507 & true & 0.7 & 20 & 0.8 \\
Gemma-4-E4B-it & true & 1.0 & 64 & 0.95 \\
Qwen3-30B-A3B-Instruct-2507 & true & 0.7 & 20 & 0.8 \\
\bottomrule
\end{tabular}
\end{table}

\paragraph{Generation caps.}
The boundary-level generator requests at most 1024 new tokens per invocation, stopping earlier at a structural boundary or the remaining token budget. Inner controller actions and outer refinement rounds each have a safety cap of 2000; these count different operations. Na\"ive complete-program generation is bounded by the remaining case budget.
The per-case generation-token budget is set to $\budgetK\times$ the source token count (Section~\ref{sec:experiments}), consumed across all generation within a case (including tokens discarded by rollback).

\paragraph{Hardware.}
The Qwen3-4B and Gemma evaluation uses a single Nvidia A100 (40GB) or GH200 (96GB) GPU per run, with BF16 model weights. Qwen3-30B-A3B uses a 96GB H100 GPU with eight Grace ARM CPU cores. Compiler and linter subprocesses run on the CPU; generation runs on the GPU. The matched regeneration and entropy replays use an RTX 3090 with BF16 Qwen3-4B, PyTorch 2.10.0+cu126, and Transformers 5.5.4. Appendix~\ref{app:rq1_runtime} defines the reported timing interval.

\section{Per-Action Semantics}
\label{app:actions}

\begin{algorithm}[t]
\caption{\method{}. Given source $S$, oracles $\mathcal{O}$, and budget $B_{\mathrm{gen}}$, \methodabbr{} alternates boundary-aligned token generation with oracle verification. Failures trigger feedback-augmented rollback; success terminates at program-level oracle pass.}
\label{alg:dtv}
\begin{algorithmic}[1]
\REQUIRE source $S$, oracles $\mathcal{O}$, generation budget $B_{\mathrm{gen}}$
\ENSURE terminal artifact $A$, success flag $s \in \{\textsc{succ}, \textsc{fail}\}$
\STATE $Y \gets \emptyset$,\ \ $\mathcal{C} \gets \{\emptyset\}$,\ \ $\mathcal{F} \gets \emptyset$ \COMMENT{init prefix, checkpoints, feedback state}
\WHILE{$N_{\mathrm{gen}} < B_{\mathrm{gen}}$}
    \STATE $Y \gets \textsc{Generate}(Y,\ \mathsf{Augment}(S, \mathcal{F}))$ \COMMENT{sample to next structural boundary or EOS}
    \STATE $A \gets \mathsf{Render}(S, Y)$;\ \ $\{(q_j, d_j)\} \gets \textsc{Verify}(A, \mathcal{O})$
    \IF{$q_j = 1$ for all applicable $j$}
        \STATE $\mathcal{C} \gets \textsc{Commit}(\mathcal{C}, Y)$
        \IF{$g(A) = \texttt{program}$}
            \STATE \textbf{return} $(A,\ \textsc{succ})$
        \ENDIF
    \ELSE
        \STATE $\mathcal{F} \gets \textsc{Update}(\mathcal{F},\ \{(q_j, d_j)\})$
        \STATE $(\textit{action},\ \ell) \gets \textsc{Decide}(\mathcal{F})$
        \IF{$\textit{action} = \textsc{bailout}$}
            \STATE \textbf{return} $(A,\ \textsc{fail})$
        \ENDIF
        \STATE $Y \gets \textsc{Rollback}(\mathcal{C}, \ell)$
    \ENDIF
\ENDWHILE
\STATE \textbf{return} $(\mathsf{Render}(S, Y),\ \textsc{fail})$ \COMMENT{budget exhausted}
\end{algorithmic}
\end{algorithm}

The controller's action set is implemented as follows; the scope, retry-mode, and bailout policies are described in Appendix~\ref{app:impl}.
\begin{itemize}[nosep,leftmargin=*]
\item \textsc{Generate}: invoke the base sampler to extend the current prefix by one or more tokens, stopping at a structural boundary (e.g., end of statement) or when the generation budget is exhausted.
\item \textsc{Verify}: render the current prefix into an artifact, select applicable oracles by scope, and record their verdicts and diagnostics into the feedback state.
\item \textsc{Commit}: mark the current prefix as a trusted checkpoint, saving the group stack state from the renderer for later structure-aware rollback.
\item \textsc{Rollback}: restore a checkpoint at scope $\ell \in \mathcal{G}$; statement rollback restores the last committed checkpoint, block/function rollback truncates to the group start and drops affected frames, and program rollback clears all checkpoints.
\item \textsc{Feedback}: construct a retry prompt from the current feedback state and ask the model to address the diagnostics; supports two realizations described in Section~\ref{sec:method}.
\item \textsc{Terminate}: end the decoding loop; triggered normally at end-of-sequence, or early in the case of a \emph{bailout} when the same diagnostic persists across scope escalations.
\end{itemize}

\section{Implementation Details}
\label{app:impl}

This appendix specifies the concrete data structures and policies that instantiate the \methodabbr{} algorithm of Section~\ref{sec:method}. Other instantiations are possible; we describe the one used in our experiments.

\noindent{\bf Scope map implementation.}
We implement the scope map $g: \mathcal{A} \to \mathcal{G}$ via a \emph{group stack} maintained by the renderer: as tokens stream in, the renderer pushes frames onto the stack at block and function openings, and pops them at the corresponding closings. This mirrors the scope-tracking stack standard parsers maintain when reading source code, with the difference that it operates on the partial generation rather than on a complete file. The renderer is paired with a token-level stopping criterion that halts the base sampler when the current prefix reaches a structural boundary (with awareness of strings, comments, and nested delimiters), handing control back to the controller to trigger the next meta step.

The renderer supplies the oracle-consumable artifact referenced by the legal prefix invariant (Section~\ref{sec:method}): it emits only when the current prefix ends at a structural boundary and the group stack is in a consistent state. The semantic side of the invariant (absence of oracle violations at the current scope) is decided by \textsc{Verify} and \textsc{Commit} in Algorithm~\ref{alg:dtv}.

\noindent{\bf Checkpoints and repair scheduling.}
Each \textsc{Commit} checkpoints the current prefix and synchronizes the renderer's group stack. Statement rollback restores the last committed checkpoint without deleting it. Block/function rollback truncates to the corresponding group start and drops the invalidated frames. The manager also supports program rollback, which clears the checkpoints, but the evaluated inner policy caps rollback at function scope and delegates whole-program retries to the outer loop. Retry has a \emph{scope} $\ell \in \{\texttt{stmt}, \texttt{block}, \texttt{func}\}$ and a \emph{mode} (Appendix~\ref{app:feedback_examples}):
\begin{itemize}[nosep,leftmargin=*]
\item \emph{Inline continuation}: the model resumes generation from the rolled-back prefix with the diagnostic summary appended as an inline comment.
\item \emph{Patch-based repair}: instead of continuing inline, the controller closes the current assistant turn and asks the model in a new user turn for a minimal patch addressing the diagnostics. The patch is constrained to the rollback scope; outputs that violate the constraint are rejected and handled by the recovery policy.
\end{itemize}
The scope and mode schedules use distinct counters. Scope escalation is a fixed heuristic over repeated local diagnostic stalls: rollback starts at statement scope, and when the same diagnostic persists beyond a fixed retry count, rollback widens to the nearest active block, or function if no block is active. Oracles may also request a coarser scope directly, subject to the function-scope cap; the in-loop oracles in our experiments do not, so rollback in all reported runs starts at statement scope. A successful or inconclusive verification resets the statement-stall counter.

Feedback scheduling is keyed by the restored prefix and repair scope. A new key starts with inline continuation. Persistent failure switches to patch repair; function-scope failures or a high diagnostic-error count can trigger this switch earlier, after an inline attempt. Once patch repair is selected, it stays selected for that key. Repeated responses without a usable patch invoke rollback and its scope-selection policy. A changed prefix/scope creates a new repair key, so mode can return to inline after widening.

\noindent{\bf Bailout.}
The controller counts verified failures at each diagnostic anchor (enclosing function, primary error code, primary message), independent of the current rollback scope. At the configured visit limit, it issues \textsc{Terminate} with bailout flag $\beta_k=1$ before the next rollback, returning the current artifact and diagnostics. Counters reset for a new outer attempt. Budget exhaustion and candidate closure also end the inner run, and the common final gate evaluates the returned artifact. Thus bailout can occur before every repair mode has been exhausted. Recovery settings were tuned on a small C-to-Rust sample and then held fixed for all reported experiments, including JS to TS and the other generators; the ablations establish the value of recovery and escalation, not the optimality of these settings.

\paragraph{Reusable core and adapters.}
The controller, rollback, feedback, shared types, and oracle interfaces form the reusable core. Language integration implements prefix rendering, context rules, and feedback configuration; verifier integration implements drivers and diagnostic handling; model-family backends implement decoding and chat-format control.

\section{Feedback Prompt Formats}
\label{app:feedback_examples}

\begin{figure}[t]
\centering
\includegraphics[width=\linewidth]{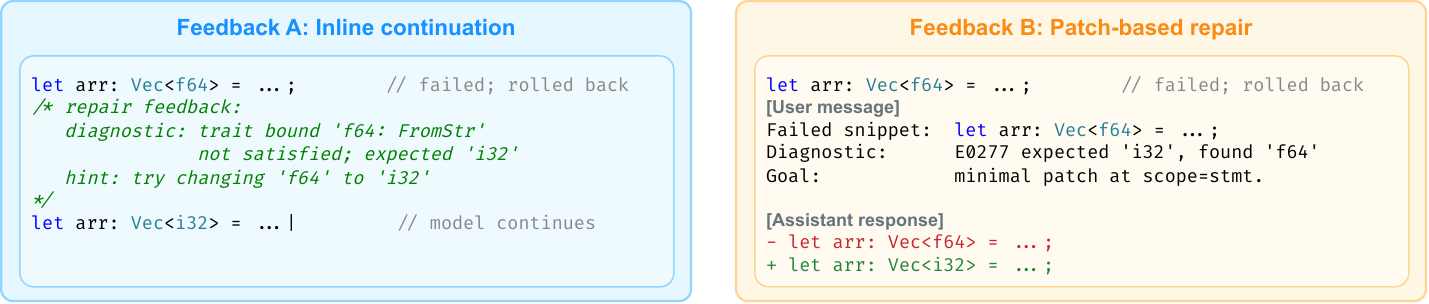}
\caption{The two repair modes (Section~\ref{sec:method}) on the same \texttt{rustc} E0277 failure. \textbf{(left)} \emph{Inline continuation}: diagnostic appended as a comment, model continues in the same turn. \textbf{(right)} \emph{Patch-based repair}: structured user turn, model returns a diff patch in a fresh assistant turn.}
\label{fig:feedback_examples}
\end{figure}

The two retry modes from Section~\ref{sec:method} differ in how the diagnostic-augmented prompt is delivered to the model and what the model is asked to return. Figure~\ref{fig:feedback_examples} shows the same C-to-Rust failure (a \texttt{rustc} E0277 trait-bound error: the model bound \texttt{Vec<f64>} where the surrounding context required \texttt{Vec<i32>}) handled by each mode.

\section{RQ1 detailed results}
\label{app:rq1}

\providecommand{\rqOnePctFmt}[1]{\fpeval{round(#1*100,1)}\%}
\providecommand{\rqOnePpFmt}[1]{%
  \edef\rqOnetmpval{\fpeval{round(#1*100,1)}}%
  \ifdim\rqOnetmpval pt>0pt+\fi\rqOnetmpval%
}

This appendix complements RQ1 (Section~\ref{sec:rq1}) with full results, paired tests, the BoN comparison, measured runtime, and strict rechecking.

\subsection{Full-set head-to-head comparison}
\label{app:rq1_full_set}

\begin{figure}[h]
  \centering
  \includegraphics[width=\linewidth]{figures/rq1_legend.pdf}\\[-0.4em]
  \begin{subfigure}[t]{0.49\linewidth}
    \centering
    \includegraphics[width=\linewidth]{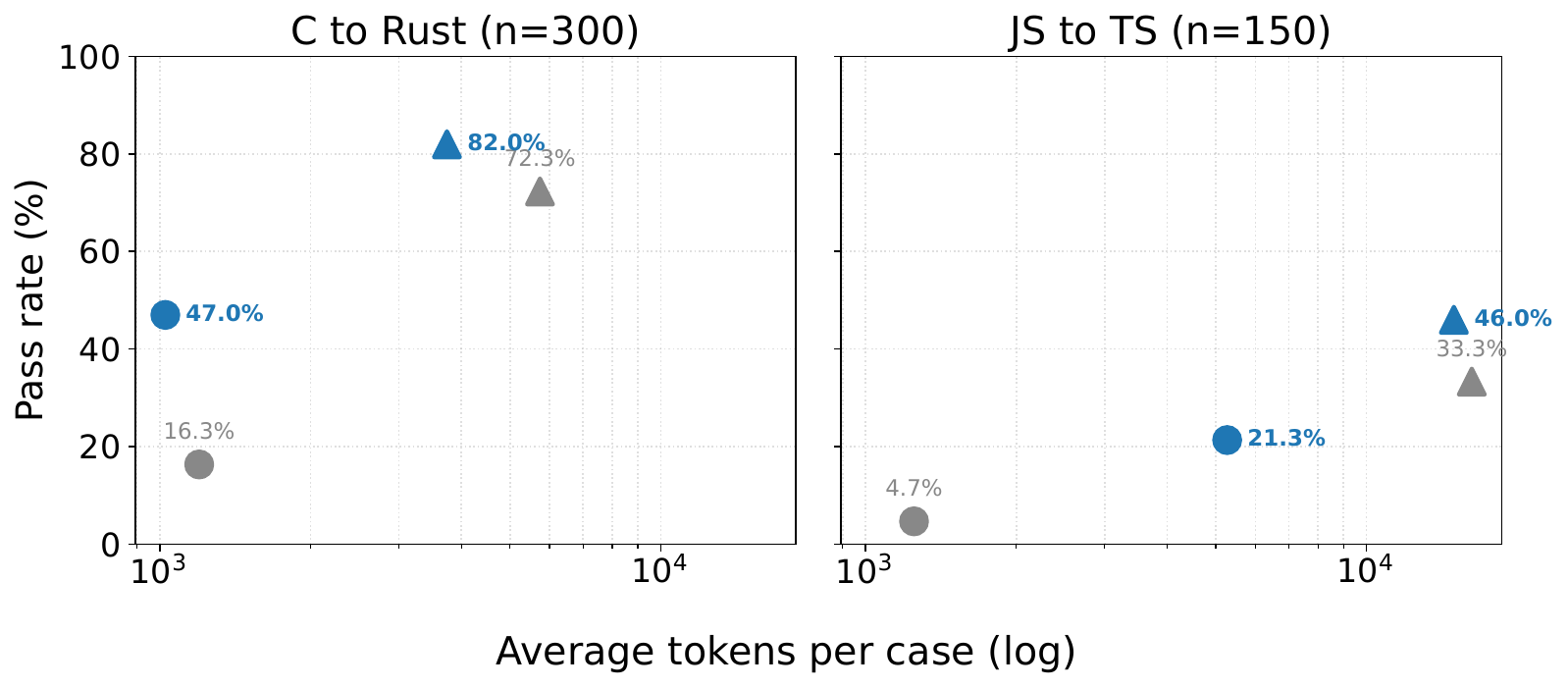}
    \caption{Per-case avg tokens (log scale) vs.\ compile pass rate.}
  \end{subfigure}
  \hfill
  \begin{subfigure}[t]{0.49\linewidth}
    \centering
    \includegraphics[width=\linewidth]{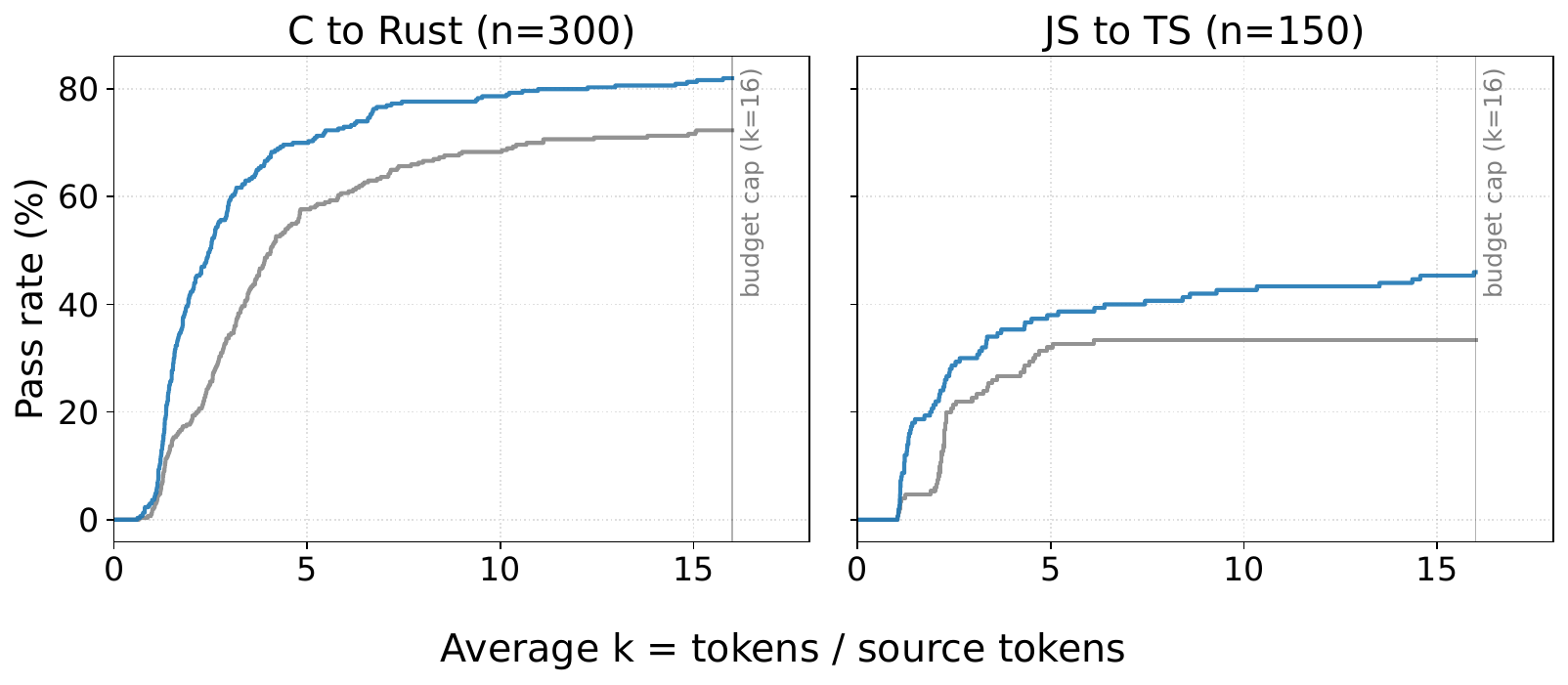}
    \caption{Cumulative pass rate vs.\ per-case cost $k$.}
  \end{subfigure}
  \caption{Full-set head-to-head comparison on C to Rust ($n=\numCRustSamples$) and JS to TS ($n=\numJSTSSamples$), restricted to the four head-to-head configurations (na\"ive vs.\ \methodabbr{} with one-shot vs.\ self-refine); the legend is shared with main-body Figure~\ref{fig:rq1} (BoN-$N$ and S\textsuperscript{*} entries do not appear in this figure). \textbf{(a)} Cost-rate scatter: \methodabbr{}/self-refine occupies the upper-left region (low cost, high pass rate) on both tasks. \textbf{(b)} Cumulative pass rate vs.\ per-case cost: \methodabbr{} exhibits the early-rise effect on both tasks and a continued-rise pattern on JS to TS that distinguishes it from na\"ive's plateau (Section~\ref{sec:rq1}).}
  \label{fig:rq1_pareto_full}
\end{figure}

Figure~\ref{fig:rq1_pareto_full} reproduces main-body Figure~\ref{fig:rq1} on the full evaluation set ($n=\numCRustSamples$ for C to Rust, $n=\numJSTSSamples$ for JS to TS), restricted to the four head-to-head configurations referenced in Section~\ref{sec:rq1} (na\"ive vs.\ \methodabbr{} with one-shot vs.\ self-refine). The cost-matched subset version, jointly comparing these head-to-head configurations against the BoN-$N$ and S\textsuperscript{*} baselines on the $n=200$ C-to-Rust and $n=100$ JS-to-TS subsets, is reported in main-body Figure~\ref{fig:rq1}; the head-to-head pass rates on the full set quoted in Section~\ref{sec:rq1} are taken directly from the data underlying this appendix figure.

\subsection{Pass rate breakdown}
\label{app:rq1_main_table}

Table~\ref{tab:rq1_main} gives the full pass-rate and token-cost breakdown underlying the compact RQ1 summary in Section~\ref{sec:rq1}.

\begin{table}[h]
  \centering
  \caption{Pass rate, average tokens-per-case ratio $k$, and tokens-per-pass (mean over passing cases, with 95\% bootstrap CI) for all four configurations on both tasks. Per-case paired token-cost statistics on the both-pass subset are reported separately in Appendix~\ref{app:rq4_bothpass_tokens}.}
  \label{tab:rq1_main}
  \small
  \rqOneMainResultsTable
\end{table}

\subsection{Runtime and verifier invocations}
\label{app:rq1_runtime}

Table~\ref{tab:runtime} reports recorded per-case elapsed time for the matched self-refine configurations, starting after model loading and source preparation and ending before output persistence. Table~\ref{tab:verifier_calls} separates outer verification rounds from \methodabbr{}'s inner verification calls. An inner call is a boundary-level verification event; on JS to TS it invokes the compiler and linter serially. These counts are not interchangeable with the controller's action count. In-loop repair trades more intermediate checks for fewer outer attempts and less continuation of failing prefixes.

\begin{table}[ht]
  \centering
  \caption{Pass rate and mean elapsed time under the shared $k=\budgetK$ cap. The task-specific cohorts are shown explicitly. Time includes generation and verification.}
  \label{tab:runtime}
  \small
  \setlength{\tabcolsep}{3pt}
  \begin{adjustbox}{max width=\linewidth}
  \rqOneRuntimeTable
  \end{adjustbox}
\end{table}

\begin{table}[ht]
  \centering
  \caption{Mean outer verification rounds and \methodabbr{} inner calls per case for the cohorts in Table~\ref{tab:runtime}.}
  \label{tab:verifier_calls}
  \small
  \rqOneVerifierCallsTable
\end{table}

The measured mean latency is \rqOneRustVerifierLatency{} seconds for a native \texttt{rustc} check and \rqOneTypeScriptVerifierLatency{} seconds for the serial TypeScript/ESLint pair, which starts separate Node.js processes. These are CPU-side costs, whereas model decoding uses the GPU (hardware in Appendix~\ref{app:models}). The Qwen3-30B-A3B experiment reduces the DTV/na\"ive elapsed ratio on both tasks while retaining higher pass rates. Since the TypeScript and ESLint checks are independent, parallel execution would reduce the pair latency toward the slower check, approximately \rqOneParallelTypeScriptLatencyEstimate{} seconds. This is a latency estimate; Table~\ref{tab:runtime} reports the measured serial implementation.

\paragraph{Takeaway.}
\methodabbr{} turns its token-budgeted success gains into wall-clock savings when verifier calls are inexpensive. With Qwen3-30B-A3B, its relative time savings grow on C to Rust and its latency premium narrows on JS to TS, while retaining higher pass rates on both tasks. This lower relative overhead strengthens the practical advantage of in-loop verification with a stronger generator.

\subsection{Shared strict TypeScript recheck}
\label{app:rq1_strict}

The inner diagnostic filter affects \methodabbr{}'s search, while every method's completed output faces the same unfiltered final compiler/linter gate (Appendix~\ref{app:experimental_setup}). We additionally recheck the saved JS-to-TS outputs with TypeScript 5.9.3 under \texttt{strict: true}, holding the setting identical across methods. Compiler options are \texttt{noEmit: true}, \texttt{skipLibCheck: true}, target ES2020, and the ES2020 and DOM libraries, with the same installed type declarations. A strict success must satisfy both the original final criterion and this extra compiler check; the denominator remains the entire cohort.

Table~\ref{tab:strict_recheck} shows that \methodabbr{} retains its pass-rate advantage across all evaluated generators, although the stricter criterion reduces absolute pass rates.

\begin{table}[ht]
  \centering
  \caption{JS-to-TS pass rates under the common final compiler/linter criterion and its conjunction with \texttt{tsc --strict}. All rows compare na\"ive/self-refine with \methodabbr{}/self-refine.}
  \label{tab:strict_recheck}
  \small
  \strictRecheckTable
\end{table}

\paragraph{Takeaway.}
\methodabbr{}/self-refine yields more outputs that pass both the original final gate and the strict recheck than na\"ive/self-refine across all evaluated generators. It therefore delivers higher verified success under the same token budget even when completed translations must satisfy stronger type requirements.

\subsection{Cost-matched baseline grid}
\label{app:rq1_baseline_table}

Table~\ref{tab:rq1_baseline} reports the full pass-rate and per-case token-cost breakdown for the cost-matched baseline grid summarized in Section~\ref{sec:rq1}. The grid compares the head-to-head configurations with best-of-$N$ for $N \in \{1, 2, 4, 8, 16, 32\}$ and S\textsuperscript{*} on the cost-matched subsets ($n=200$ for C to Rust, $n=100$ for JS to TS); na\"ive/self-refine and \methodabbr{}/self-refine are repeated here at the subset $n$ for direct comparison against the BoN and S\textsuperscript{*} columns.

\begin{table}[h]
  \centering
  \caption{Pass rate, average token cost per case, tokens per successful pass, and per-case token cost relative to \methodabbr{}/self-refine for the cost-matched baseline grid on the $n{=}200$ C-to-Rust and $n{=}100$ JS-to-TS subsets. The $\times$ ratio is per-case tokens divided by \methodabbr{}/self-refine's per-case tokens on the same task. na\"ive/self-refine and \methodabbr{}/self-refine are repeated here at the subset $n$ for direct comparison.}
  \label{tab:rq1_baseline}
  \small
  \resizebox{\linewidth}{!}{
\begin{tabular}{llrrrrr}
\toprule
\textbf{Task} & \textbf{Configuration} & \textbf{Pass} & \textbf{Pass rate} & \textbf{Tok/case} & \textbf{Tok/pass} & \textbf{vs \methodabbr{} (tok/case)} \\
\midrule
C $\to$ Rust ($n=$200) & naive/self-refine & 152/200 & 76.0\% & 5266 & 1798 & 1.51$\times$ \\
 & \methodabbr{}/self-refine & 169/200 & 84.5\% & 3489 & 1805 & 1.00$\times$ \\
 & naive/BoN-1 & 31/200 & 15.5\% & 737 & 377 & 0.21$\times$ \\
 & naive/BoN-2 & 42/200 & 21.0\% & 1437 & 476 & 0.41$\times$ \\
 & naive/BoN-4 & 58/200 & 29.0\% & 2772 & 928 & 0.79$\times$ \\
 & naive/BoN-8 & 73/200 & 36.5\% & 5154 & 1880 & 1.48$\times$ \\
 & naive/BoN-16 & 84/200 & 42.0\% & 9574 & 2573 & 2.74$\times$ \\
 & naive/BoN-32 & 89/200 & 44.5\% & 17866 & 3445 & 5.12$\times$ \\
 & S*/n8r3 & 151/200 & 75.5\% & 16563 & 11551 & 4.75$\times$ \\
\midrule
JS $\to$ TS ($n=$100) & naive/self-refine & 37/100 & 37.0\% & 15105 & 1740 & 1.09$\times$ \\
 & \methodabbr{}/self-refine & 50/100 & 50.0\% & 13815 & 2185 & 1.00$\times$ \\
 & naive/BoN-1 & 7/100 & 7.0\% & 1186 & 435 & 0.09$\times$ \\
 & naive/BoN-2 & 7/100 & 7.0\% & 2377 & 435 & 0.17$\times$ \\
 & naive/BoN-4 & 7/100 & 7.0\% & 4727 & 435 & 0.34$\times$ \\
 & naive/BoN-8 & 7/100 & 7.0\% & 9509 & 435 & 0.69$\times$ \\
 & naive/BoN-16 & 7/100 & 7.0\% & 18871 & 435 & 1.37$\times$ \\
 & naive/BoN-32 & 7/100 & 7.0\% & 37664 & 435 & 2.73$\times$ \\
 & S*/n8r3 & 33/100 & 33.0\% & 27388 & 9751 & 1.98$\times$ \\
\bottomrule
\end{tabular}
}
\end{table}

\subsection{Paired comparisons (compile pass)}
\label{app:rq1_pairs}

Table~\ref{tab:rq1_pairs} reports the paired compile-pass comparisons used to separate \methodabbr{}'s gains from case-level difficulty variation.

\begin{table}[h]
  \centering
  \caption{Paired McNemar tests on per-case compile pass for the four canonical comparisons on both tasks. The Only-A and Only-B columns report the discordant counts. All comparisons in the main text are paired (same case identifiers under the same dataset and budget).}
  \label{tab:rq1_pairs}
  \small
  \rqOnePairTable
\end{table}

\subsection{Functional guardrail (C to Rust)}
\label{app:rq1_functional}

The pass-rate metric in the main body uses compilation as the primary criterion because compilation is the dense in-loop verifier signal that the \methodabbr{} controller consumes. To rule out the possibility that \methodabbr{}'s compile-rate gains are achieved at the expense of behavioral correctness (for example, by patching function bodies with stubs that compile but fail differential tests), we run a paired functional analysis on the matched outer-loop setting: na\"ive/self-refine versus \methodabbr{}/self-refine. A case is a full success only if the Rust translation compiles and matches the C source on all differential test inputs; the per-case test pass rate is the fraction of differential tests passed and is set to zero on compilation failure. Table~\ref{tab:rq1_func} reports the marginal outcomes per configuration, with the conditional column giving full-success rate among cases that compile.

\begin{table}[h]
  \centering
  \caption{Marginal functional outcomes for na\"ive/self-refine and \methodabbr{}/self-refine on C to Rust ($n=\numCRustSamples$). Compile is the share of cases whose Rust translation compiles, the average test pass rate is the per-case fraction of differential tests passed (zero on compilation failure) averaged over all $n$ cases, and the conditional column gives the full-success rate among the cases that compile.}
  \label{tab:rq1_func}
  \small
  \begin{tabular}{lrrr}
  \toprule
  \textbf{Configuration} & \textbf{Compile} & \textbf{Avg.\ test pass rate} & \textbf{Full $\mid$ compile} \\
  \midrule
  na\"ive/self-refine & \rqOneCrustNaiveSrNumPass/\numCRustSamples{} (\rqOneCrustNaiveSrPass{}) & \rqOnePctFmt{\rqOneCrustNaiveSrFuncNormMean} & \rqOneCrustNaiveSrFuncCondFullPass{} \\
  \methodabbr{}/self-refine & \rqOneCrustDtvSrNumPass/\numCRustSamples{} (\rqOneCrustDtvSrPass{}) & \rqOnePctFmt{\rqOneCrustDtvSrFuncNormMean} & \rqOneCrustDtvSrFuncCondFullPass{} \\
  \bottomrule
  \end{tabular}
\end{table}

The marginal view shows \methodabbr{}'s conditional full-success rate (\rqOneCrustDtvSrFuncCondFullPass{}) below na\"ive's (\rqOneCrustNaiveSrFuncCondFullPass{}), but this conditional comparison conditions on different sets of cases: \methodabbr{} compiles \rqOneCrustDtvSrNumPass{} of \numCRustSamples{} cases versus na\"ive's \rqOneCrustNaiveSrNumPass{}, so \methodabbr{}'s denominator includes harder cases that na\"ive cannot compile at all. The cleanest way to remove this selection effect is to restrict to the cases where both configurations produce a compiling translation. On this matched compile set (Table~\ref{tab:rq1_func_both_compile}, $n=\rqOneCrustNaiveSrVsDtvSrFuncBcN$), \methodabbr{}'s full-success rate is \rqOneCrustNaiveSrVsDtvSrFuncBcBFullRate{} versus na\"ive's \rqOneCrustNaiveSrVsDtvSrFuncBcAFullRate{} (\rqOneCrustNaiveSrVsDtvSrFuncBcFullDiffPp{}~pp; McNemar $p\rqOneCrustNaiveSrVsDtvSrFuncBcFullMcnemarP$), and its average per-case differential test rate is \rqOneCrustNaiveSrVsDtvSrFuncBcBAvgRate{} versus na\"ive's \rqOneCrustNaiveSrVsDtvSrFuncBcAAvgRate{} (\rqOneCrustNaiveSrVsDtvSrFuncBcAvgDiff{}~pp, 95\% bootstrap CI $[\rqOneCrustNaiveSrVsDtvSrFuncBcAvgDiffCiLo, \rqOneCrustNaiveSrVsDtvSrFuncBcAvgDiffCiHi]$). Neither difference reaches significance and the bootstrap CI straddles zero, so on cases where both methods produce a compiling target, \methodabbr{}'s behavioral correctness is statistically indistinguishable from na\"ive's.

\begin{table}[h]
  \centering
  \caption{Both-compile paired comparison on C to Rust, restricted to cases where both na\"ive/self-refine and \methodabbr{}/self-refine produce a compiling translation ($n=\rqOneCrustNaiveSrVsDtvSrFuncBcN$). The full-success columns report the share of these cases that pass all differential tests; the average rate column averages the per-case fraction of differential tests passed, with a 95\% paired-bootstrap CI on the difference.}
  \label{tab:rq1_func_both_compile}
  \small
  \begin{tabular}{rrrrrl}
  \toprule
  \textbf{$n$} & \textbf{na\"ive full} & \textbf{\methodabbr{} full} & \textbf{Full diff (pp)} & \textbf{McNemar} & \textbf{Avg.\ rate diff [95\% CI]} \\
  \midrule
  \rqOneCrustNaiveSrVsDtvSrFuncBcN{} & \rqOneCrustNaiveSrVsDtvSrFuncBcAFullRate{} & \rqOneCrustNaiveSrVsDtvSrFuncBcBFullRate{} & \rqOneCrustNaiveSrVsDtvSrFuncBcFullDiffPp{} & $p\rqOneCrustNaiveSrVsDtvSrFuncBcFullMcnemarP$ & \rqOneCrustNaiveSrVsDtvSrFuncBcAvgDiff{}~pp [$\rqOneCrustNaiveSrVsDtvSrFuncBcAvgDiffCiLo$, $\rqOneCrustNaiveSrVsDtvSrFuncBcAvgDiffCiHi$] \\
  \bottomrule
  \end{tabular}
\end{table}

The full-sample paired view (Table~\ref{tab:rq1_func_pairs}) confirms this conclusion across all \numCRustSamples{} cases. The discordant counts are \rqOneCrustNaiveSrVsDtvSrFuncOnlyA{} (only na\"ive) versus \rqOneCrustNaiveSrVsDtvSrFuncOnlyB{} (only \methodabbr{}), giving a difference of \rqOneCrustNaiveSrVsDtvSrFuncDiffPp{}~pp on full success (McNemar $p\rqOneCrustNaiveSrVsDtvSrFuncMcnemarP$); the paired test pass rate difference is \rqOnePpFmt{\rqOneCrustNaiveSrVsDtvSrFuncNormDiff}~pp with a 95\% bootstrap CI of $[\rqOnePpFmt{\rqOneCrustNaiveSrVsDtvSrFuncNormDiffCiLo}, \rqOnePpFmt{\rqOneCrustNaiveSrVsDtvSrFuncNormDiffCiHi}]$~pp. Together, the both-compile and full-sample paired analyses rule out a hidden behavioral regression: \methodabbr{}'s compile-rate gains are not paired with worse differential correctness.

\begin{table}[h]
  \centering
  \caption{Full-sample paired comparison on full-functional success for na\"ive/self-refine versus \methodabbr{}/self-refine across all \numCRustSamples{} C-to-Rust cases. ``Only na\"ive'' and ``Only \methodabbr{}'' count cases where only the named method achieves full success. The test rate diff is the paired difference in average test pass rate (in percentage points) with a 95\% bootstrap CI.}
  \label{tab:rq1_func_pairs}
  \small
  \begin{tabular}{rrrrl}
  \toprule
  \textbf{Only na\"ive} & \textbf{Only \methodabbr{}} & \textbf{Diff (pp)} & \textbf{McNemar} & \textbf{Test rate diff (pp) [95\% CI]} \\
  \midrule
  \rqOneCrustNaiveSrVsDtvSrFuncOnlyA{} & \rqOneCrustNaiveSrVsDtvSrFuncOnlyB{} & \rqOneCrustNaiveSrVsDtvSrFuncDiffPp{} & $p\rqOneCrustNaiveSrVsDtvSrFuncMcnemarP$ & \rqOnePpFmt{\rqOneCrustNaiveSrVsDtvSrFuncNormDiff} [$\rqOnePpFmt{\rqOneCrustNaiveSrVsDtvSrFuncNormDiffCiLo}$, $\rqOnePpFmt{\rqOneCrustNaiveSrVsDtvSrFuncNormDiffCiHi}$] \\
  \bottomrule
  \end{tabular}
\end{table}

\section{Cross-model robustness}
\label{app:rq2}

\begin{figure}[t]
  \centering
  \includegraphics[width=0.85\linewidth]{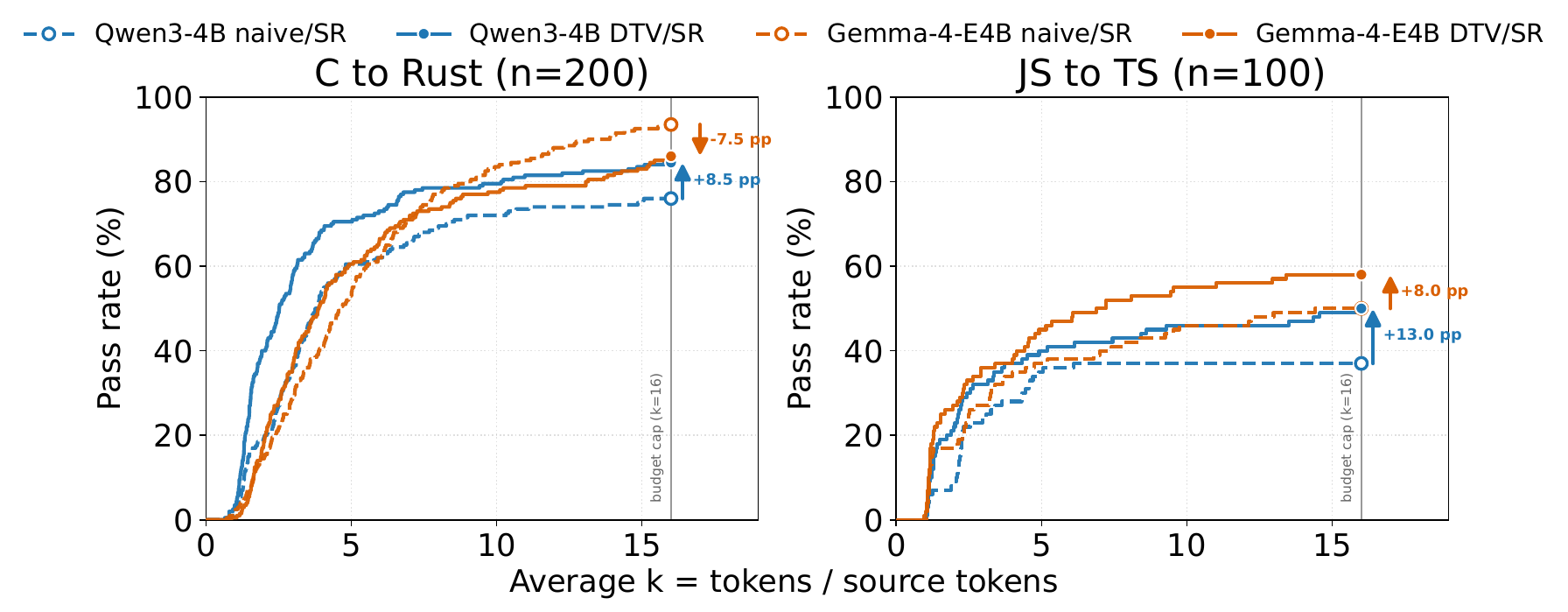}
  \caption{Cross-model cumulative pass rate vs.\ normalized per-case cost $k$ for na\"ive/self-refine and \methodabbr{}/self-refine on both tasks. \methodabbr{} exceeds na\"ive across the lower budget range on all four pairs; at the budget cap, \methodabbr{} remains ahead on three pairs, while on \rqTwoGemmaName{} C to Rust the curves cross near $k\approx 7$ and na\"ive overtakes (behavioral correctness preserved; Appendix~\ref{app:rq2_functional}).}
  \label{fig:rq2}
\end{figure}

This appendix presents the cross-model analysis referenced in Section~\ref{sec:rq1}. Figure~\ref{fig:rq2} compares \rqTwoQwenName{} and \rqTwoGemmaName{} on matched task subsets; Figure~\ref{fig:qwen30b_scaling} adds Qwen3-30B-A3B on the first 100 cases per task. The metric hierarchy remains pass rate and token cost, with C-to-Rust differential testing as a guardrail. In the Qwen3-4B/Gemma comparison, \methodabbr{} exceeds na\"ive at the budget cap on three pairs (\rqTwoQwenCrustGain{}, \rqTwoQwenJstsGain{}, \rqTwoGemmaJstsGain{}) and reduces normalized cost on three (\rqTwoQwenCrustKDelta{}, \rqTwoQwenJstsKDelta{}, \rqTwoGemmaJstsKDelta{}). Gemma C to Rust exhibits the budget-dependent crossover discussed below.

On \rqTwoGemmaName{} C to Rust, the comparison is budget-dependent (Figure~\ref{fig:rq2}, left panel). \methodabbr{}/self-refine reaches a higher pass rate than na\"ive/self-refine across the lower-to-mid budget range, with a peak lead of $+10.5$\,pp at $k\approx 4$; the curves cross near $k\approx 7$ and na\"ive/self-refine continues to climb past \methodabbr{} through the cap, ending at \rqTwoGemmaCrustNaivePass{} versus \rqTwoGemmaCrustDtvPass{}. We attribute the erosion of \methodabbr{}'s early-rise advantage to the high baseline pass rate on this pair: with na\"ive/self-refine already passing \rqTwoGemmaCrustNaivePass{} of cases at the cap, the residual hard cases provide fewer opportunities for \methodabbr{}'s verify-and-rollback to convert into marginal passes than na\"ive's outer-loop retry produces over the same total budget. The per-case failure modes that drive the cap-budget gap, and whether they fall outside the scope of \methodabbr{}'s structure-aware rollback, are part of the per-case analysis in Section~\ref{sec:rq4}. Behavioral correctness on the differential-testing guardrail is unchanged on this pair (\rqTwoGemmaCrustNaiveFuncFull{} vs \rqTwoGemmaCrustDtvFuncFull{} in functional pass rate; Appendix~\ref{app:rq2_functional}).

On JS to TS, both model families show \methodabbr{} pass-rate gains at the budget cap (\rqTwoQwenJstsGain{} on \rqTwoQwenName{}, \rqTwoGemmaJstsGain{} on \rqTwoGemmaName{}) at reduced normalized cost (\rqTwoQwenJstsKDelta{} and \rqTwoGemmaJstsKDelta{} respectively). Across the four pairs, \methodabbr{}'s gains on both pass rate and per-case cost generalize across model families: on three pairs \methodabbr{} exceeds na\"ive at the budget cap in both metrics, and across the lower portion of the budget range \methodabbr{} exceeds na\"ive in pass rate on all four. The single cap-budget exception is concentrated on the pair where the baseline is highest (\rqTwoGemmaName{} C to Rust at \rqTwoGemmaCrustNaivePass{}).

\subsection{Qwen3-30B-A3B scaling and one-shot decomposition}
\label{app:rq2_large}

\begin{figure}[ht]
  \centering
  \includegraphics[width=0.85\linewidth]{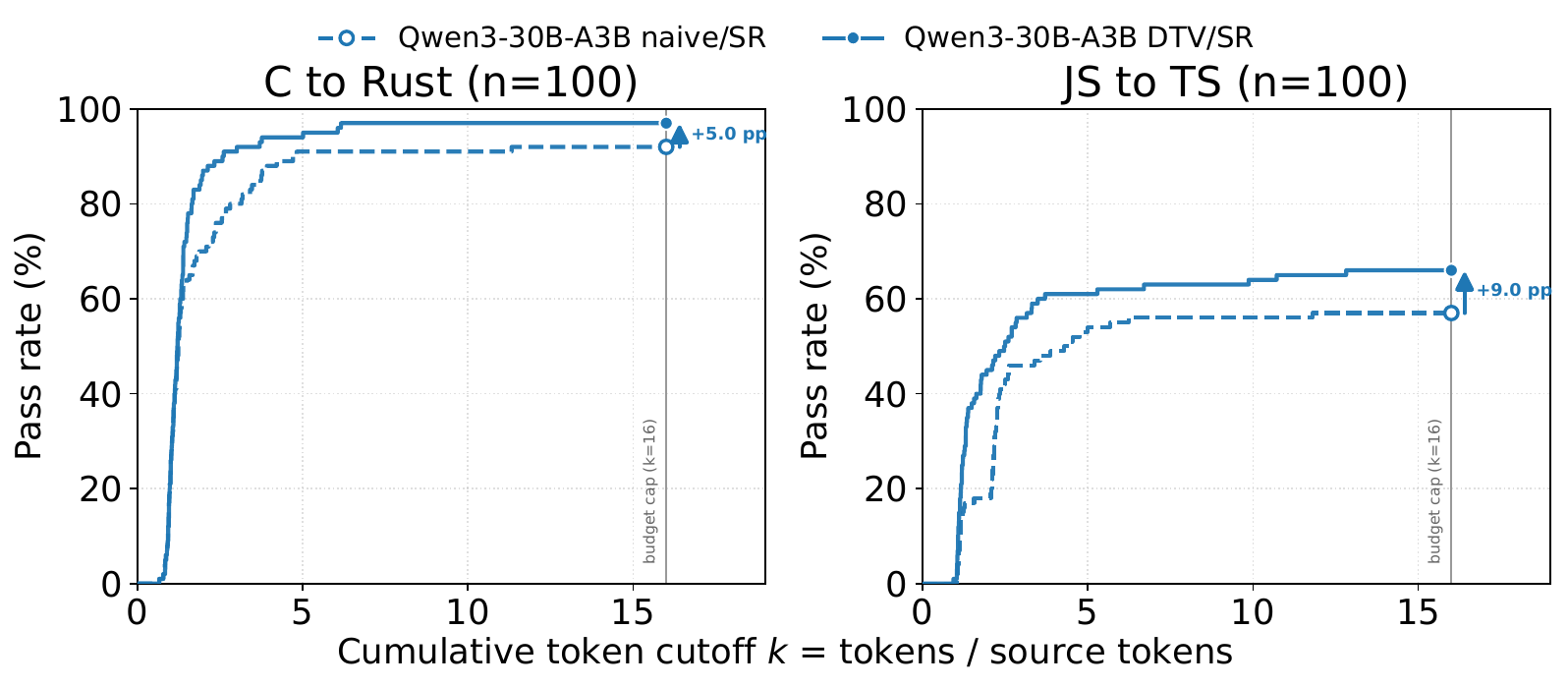}
  \caption{Qwen3-30B-A3B cumulative pass rates versus normalized per-case cost, reconstructed from raw $k=16$ batch records on 100 cases per task. Steps mark actual successful-case costs. Dashed: na\"ive/self-refine; solid: \methodabbr{}/self-refine.}
  \label{fig:qwen30b_scaling}
\end{figure}

We evaluate \texttt{Qwen/Qwen3-30B-A3B-Instruct-2507}, a mixture-of-experts generator with 30.5B total parameters, on the first 100 cases of each randomly sampled main-evaluation cohort. These are prefixes of the same pools used above, rather than replacements for the 200-case Gemma C-to-Rust comparison. Both methods use the same $k=\budgetK$ cap, prompts, and final verifiers. Figure~\ref{fig:qwen30b_scaling} reconstructs cumulative pass rates from the raw batch records, placing each step at a successful case's actual generated-token cost normalized by its source-token count. The curves use full-budget runs rather than independent reruns at each cutoff. Crash and missing-output cases count as failures at every cutoff, including two JS-to-TS \methodabbr{} crash records and one missing case.

At the integer cutoffs from $k=2$ through $k=16$, \methodabbr{} leads on C to Rust, ending at \rqTwoQwenThirtyBCrustDtvPass{} versus \rqTwoQwenThirtyBCrustNaivePass{}. On JS to TS it leads at the integer cutoffs from $k=1$ through $k=16$ and ends at \rqTwoQwenThirtyBJstsDtvPass{} versus \rqTwoQwenThirtyBJstsNaivePass{}. The low-budget advantages are pronounced at $k=2$: \rqTwoQwenThirtyBCrustDtvBudgetTwoPass{} versus \rqTwoQwenThirtyBCrustNaiveBudgetTwoPass{}, and \rqTwoQwenThirtyBJstsDtvBudgetTwoPass{} versus \rqTwoQwenThirtyBJstsNaiveBudgetTwoPass{}.

Table~\ref{tab:qwen30b_oneshot} isolates the first complete-program attempt, excluding outer self-refinement. \methodabbr{} improves its pass rate from \rqTwoQwenThirtyBCrustNaiveOneshotPass{} to \rqTwoQwenThirtyBCrustDtvOneshotPass{} on C to Rust and from \rqTwoQwenThirtyBJstsNaiveOneshotPass{} to \rqTwoQwenThirtyBJstsDtvOneshotPass{} on JS to TS. The reported mean $k$ values condition on cases with a recorded first outer verification. One-shot is an operating point defined by the first attempt, not by the CDF cutoff $k=1$; a first attempt may itself consume more than one source-length of tokens.

\begin{table}[ht]
  \centering
  \caption{First-attempt Qwen3-30B-A3B results. Pass rates use the full 100-case cohorts; mean normalized cost $k$ uses cases with a recorded first outer verification.}
  \label{tab:qwen30b_oneshot}
  \small
  \rqTwoQwenThirtyBOneshotTable
\end{table}

\paragraph{Takeaway.}
With Qwen3-30B-A3B, \methodabbr{} yields more successful translations than the na\"ive baseline both on the first complete-program attempt and at the shared token-budget cap. It therefore improves a stronger generator's first-attempt reliability as well as its success under a fixed overall budget, demonstrating benefits beyond additional outer retries.

\subsection{Marginal stats}
\label{app:rq2_marginal}

Table~\ref{tab:rq2_main} reports per-cell pass rate, average per-case $k$, and tokens per pass for na\"ive/self-refine and \methodabbr{}/self-refine on both tasks. Each Gemma cell uses the corresponding RQ1 sample list ($n=200$ for C to Rust, $n=100$ for JS to TS); Qwen cells are restricted to the same prefixes for paired cross-model comparison, so the matched case set within each (model, task) cell has the same case identifiers across configurations.

\begin{table}[h]
  \centering
  \caption{Per-cell pass rate, average per-case $k$, and tokens per pass for na\"ive/self-refine and \methodabbr{}/self-refine on both tasks. $k$ is normalized in each model's own tokenizer. Costs use recorded token consumption for both passing and failing cases; missing cases are assigned $k=\budgetK$.}
  \label{tab:rq2_main}
  \small
  \resizebox{\linewidth}{!}{\rqTwoMainResultsTable}
\end{table}

\subsection{Paired comparisons}
\label{app:rq2_paired}

Within each cell, both methods see the same case identifiers under the same per-case budget cap, so paired McNemar tests on per-case pass directly isolate method-attributable gain from case-level difficulty variation (Table~\ref{tab:rq2_pairs}). To separate per-success efficiency from marginal pass-rate composition differences, Table~\ref{tab:rq2_paired_k} restricts to the subset of cases where both methods produce a passing translation and compares average $k$ on that matched both-pass subset.

\begin{table}[h]
  \centering
  \caption{Paired McNemar tests on per-case pass within each cell (same case identifiers under na\"ive vs.\ \methodabbr{}). Discordant counts: ``Only na\"ive'' = na\"ive passes and \methodabbr{} does not; ``Only \methodabbr{}'' = \methodabbr{} passes and na\"ive does not.}
  \label{tab:rq2_pairs}
  \small
  \rqTwoPairTable
\end{table}

\begin{table}[h]
  \centering
  \caption{Both-pass subset analysis: average $k$ on the cases where both na\"ive/self-refine and \methodabbr{}/self-refine produce a passing translation. This view isolates \methodabbr{}'s per-success efficiency from the marginal pass-rate composition difference reported in Table~\ref{tab:rq2_main}. $\Delta k$ is the percentage difference (\methodabbr{} minus na\"ive), normalized by na\"ive's mean $k$; negative values favor \methodabbr{}.}
  \label{tab:rq2_paired_k}
  \small
  \rqTwoPairedKTable
\end{table}

\subsection{Functional guardrail (C to Rust)}
\label{app:rq2_functional}

Mirroring RQ1's functional guardrail (Appendix~\ref{app:rq1_functional}), Table~\ref{tab:rq2_func} reports differential-testing outcomes on both C-to-Rust cells to check whether \methodabbr{}'s pass-rate change accompanies a behavioral-correctness loss. The Qwen functional values here use the Gemma-matched 200 cases; the full-300 Qwen view is in Appendix~\ref{app:rq1_functional}.

\begin{table}[h]
  \centering
  \caption{Functional outcomes for na\"ive/self-refine and \methodabbr{}/self-refine on C to Rust under the matched cross-model case set ($n=200$ per model). Same metrics as RQ1's functional guardrail (Appendix~\ref{app:rq1_functional}); the Qwen functional values here differ from the RQ1 appendix because RQ1 uses the full $n=300$ set while RQ2 restricts to the Gemma-matched first 200.}
  \label{tab:rq2_func}
  \small
  \resizebox{\linewidth}{!}{\rqTwoCrustFunctionalTable}
\end{table}

On the matched 200 ids, na\"ive/self-refine on \rqTwoQwenName{} produces \rqTwoQwenCrustNaiveFuncFull{} full-functional success and \methodabbr{}/self-refine \rqTwoQwenCrustDtvFuncFull{} (\rqTwoQwenCrustFuncDelta{}; McNemar $p\rqTwoQwenCrustFuncMcnemarP{}$). The marginal difference is small in absolute terms (4 cases out of 200) and consistent with the n.s.\ RQ1 result on the full 300 set; we therefore do not interpret it as a substantive behavioral-correctness loss.

On \rqTwoGemmaName{} C to Rust, the marginal full-functional success rate is identical between na\"ive/self-refine and \methodabbr{}/self-refine (\rqTwoGemmaCrustNaiveFuncFull{} versus \rqTwoGemmaCrustDtvFuncFull{}, exact tie; McNemar $p\rqTwoGemmaCrustFuncMcnemarP{}$). \methodabbr{}'s compiling pool is smaller than na\"ive's by 15 cases (170 versus 185), but the conditional full-success rate within \methodabbr{}'s compiling pool is higher than na\"ive's (\rqTwoGemmaCrustDtvFuncCondFullPass{} versus \rqTwoGemmaCrustNaiveFuncCondFullPass{}), suggesting that the cases \methodabbr{} does compile are more likely to also pass differential tests. Aggregating both views, \methodabbr{}'s pass-rate regression on this near-saturation cell does not coincide with a behavioral-correctness regression; on the cases \methodabbr{} does compile, behavioral pass-through is in fact slightly higher than na\"ive's, indicating that the lost compiles concentrate among translations whose compile-pass-but-test-fail status would not have been detectable by the in-loop oracle alone.

\section{RQ2 detailed results}
\label{app:rq3}

This appendix complements RQ2 (Section~\ref{sec:rq3}) with ablation costs, paired tests, outer-loop rescue accounting, and escalation traces.

\subsection{Cost breakdown}
\label{app:rq3_main_table}

Table~\ref{tab:rq3_full} reports the full per-configuration cost breakdown underlying the compact RQ2 summary in Section~\ref{sec:rq3}. The Tokens/pass column reports the mean tokens consumed on cases that compile-pass with a 95\% bootstrap CI, mirroring the reporting convention of Appendix~\ref{app:rq1_main_table}. \methodabbr{}-full's per-success cost (\rqThreeCrustDtvSrTokensPerPass{}) is higher than each of the three ablations' (\rqThreeCrustNoFeedbackTokensPerPass{}, \rqThreeCrustNoEscalationTokensPerPass{}, \rqThreeCrustDetectAbortTokensPerPass{}); the gap reflects the additional in-loop verification and structured-recovery work \methodabbr{}-full performs on cases that succeed. At the per-case granularity at which the budget is actually allocated, \methodabbr{}-full nevertheless converts more of the budget into passing translations (the Avg.\ tokens column) than two of the three ablations.

\begin{table}[h]
  \centering
  \caption{Per-configuration pass rate, average tokens per case, and Tokens/pass with 95\% bootstrap CI on the first \rqThreeCrustDtvSrNumTotal{} C-to-Rust cases. The Avg.\ tokens column is the per-case mean over all cases (failures hit the budget cap); the Tokens/pass column averages tokens over compile-passing cases, with a bootstrap CI on this mean.}
  \label{tab:rq3_full}
  \small
  \rqThreeMainResultsTable
\end{table}

\subsection{Paired ablation comparisons}
\label{app:rq3_paired}

Table~\ref{tab:rq3_pairs} reports paired McNemar tests for each ablation against the \methodabbr{}-full baseline on three metrics. Within each row, both configurations see the same case identifiers, so the discordant counts (Only-A: baseline-only pass; Only-B: ablation-only pass) directly attribute outcome differences to the ablation rather than to case-level difficulty variation.

\begin{table}[h]
  \centering
  \caption{Paired McNemar tests on per-case compile, inner-1shot, and functional pass for each ablation against the \methodabbr{}-full baseline ($n=\rqThreeCrustDtvSrNumTotal{}$ paired cases). Only-A: baseline-only pass; Only-B: ablation-only pass. With $n=100$, McNemar can reliably detect effects of approximately 10\,pp or larger; differences below this floor with non-significant $p$-values should be read as point estimates rather than null effects.}
  \label{tab:rq3_pairs}
  \small
  \resizebox{\linewidth}{!}{\rqThreePairTable}
\end{table}

The compile-rate drops for detect-and-abort (\rqThreeCrustDtvSrVsDetectAbortDiffPp{}\,pp, $p\rqThreeCrustDtvSrVsDetectAbortMcnemarP{}$) and no-feedback (\rqThreeCrustDtvSrVsNoFeedbackDiffPp{}\,pp, $p\rqThreeCrustDtvSrVsNoFeedbackMcnemarP{}$) are paired-significant. The no-escalation drop (\rqThreeCrustDtvSrVsNoEscalationDiffPp{}\,pp, $p\rqThreeCrustDtvSrVsNoEscalationMcnemarP{}$) is directionally consistent with the other two but falls below the detectability floor at $n=\rqThreeCrustDtvSrNumTotal{}$. Inner-1shot drops for detect-and-abort (\rqThreeCrustDtvSrVsDetectAbortInnerOneDiffPp{}\,pp) and no-feedback (\rqThreeCrustDtvSrVsNoFeedbackInnerOneDiffPp{}\,pp) are large enough to be detected at this sample size ($p\rqThreeCrustDtvSrVsDetectAbortInnerOneMcnemarP{}$ and $p\rqThreeCrustDtvSrVsNoFeedbackInnerOneMcnemarP{}$ respectively); no-escalation leaves inner-1shot unchanged ($p\rqThreeCrustDtvSrVsNoEscalationInnerOneMcnemarP{}$). All three functional differences fall below the detectability floor and should be read alongside the descriptive ordering reported in Section~\ref{sec:rq3}.

\subsection{Outer self-refine rescue contribution}
\label{app:rq3_rescue}

Outer self-refine compensates for inner-loop ablations by rerunning \methodabbr{} with accumulated diagnostic feedback in the prompt. Table~\ref{tab:rq3_rescue} reports the rescue rate (cases recovered by outer retry as a fraction of inner-1shot failures) for each configuration. Rescue rates fall in a 62--73\% band across all four configurations, indicating outer self-refine's rescue capacity is roughly comparable across inner-loop ablations. Consequently, the gap between inner-1shot and final compile rates (Table~\ref{tab:rq3_rescue}) reflects what outer self-refine can compensate for rather than how the ablation interacts with the outer loop.

\begin{table}[h]
  \centering
  \caption{Outer self-refine rescue accounting: rescued cases (final compile $-$ inner-1shot compile) as a fraction of inner-1shot failures, for each configuration. The rescue rate band (62.5--73.2\%) indicates that outer self-refine's rescue capacity is roughly independent of which inner mechanism is ablated.}
  \label{tab:rq3_rescue}
  \small
  \rqThreeRescueTable
\end{table}

\subsection{Trace walk-throughs: when escalation helps and hurts}
\label{app:rq3_traces}

We examine two paired cases illustrating the mechanism underlying the \rqThreeCrustDtvSrVsNoEscalationDiffPp{}\,pp final-pass-rate gap between \methodabbr{}-full and no-escalation. The runs use temperature sampling rather than greedy decoding, so paired-trace differences are mechanistic rather than deterministic counterfactual.

\paragraph{Escalation rescues a non-local mutability error.}
Case \texttt{s368372837} (only-A: \methodabbr{}-full passes, no-escalation fails) is a non-local error: rustc reports \texttt{cannot assign twice to immutable variable sold} at a mutation site and suggests making an earlier binding mutable. \methodabbr{}-full attempts three statement-scope rollbacks to the same prefix length, each producing the same diagnostic; the statement-stall counter then triggers escalation to block scope, which truncates the committed prefix back through the binding region. Regeneration produces a passing verify and commit, and the case finishes in 652 generated tokens. No-escalation cycles through statement repair until the per-case budget cap of 3600 tokens, without reaching outer-loop verification. Across the no-escalation cohort, 10 cases exhaust their budget within inner-loop activity (zero recorded outer rounds), consistent with statement-only rollback failing to make progress on these diagnostics.

\paragraph{Escalation discards a useful prefix.}
Case \texttt{s126370263} (only-B: no-escalation passes, \methodabbr{}-full fails) shows the opposite failure mode. \methodabbr{}-full repeatedly attempts patch-based repair on a local \texttt{if}/\texttt{else} expression; after several stalls the controller escalates to function-scope rollback, which truncates the committed prefix to length zero and forces regeneration from scratch. The regenerated trajectory does not converge within the budget. No-escalation on the same case applies a single statement-scope rollback to fix an unrelated macro syntax error, commits, and passes on the first outer round.

\paragraph{Aggregate caveat.}
These two cases illustrate the canonical positive and negative interactions between rollback scope and the regeneration trajectory, but they do not generalize mechanically to all 20 paired identity flips. Of the 13 only-A cases, \methodabbr{}-full uses non-statement rollback on 7; the remaining 6 succeed without scope widening in \methodabbr{}-full and fail under no-escalation, indicating that escalation policy also affects generation trajectories indirectly (e.g., function-scope rollback resets the feedback mechanism from patch-based to inline continuation, exposing the model to a different repair-prompt format). Of the 7 only-B cases, 5 show \methodabbr{}-full taking non-statement rollback (consistent with over-aggressive scope widening discarding useful prefix); the other 2 show no scope-widening at all, suggesting stochastic trajectory variation under temperature sampling.

\section{RQ3 detailed results}
\label{app:rq4}

This appendix complements the main-body RQ3 results (Section~\ref{sec:rq4}) with the fix-shape classifier algorithm, the full fix-shape breakdown including UNKNOWN counts and the alternative R\textsubscript{final} classification, the per-error-code rescue rate, the shared-code ratio distribution underlying the on-rescued mechanism claim, and the fix-shape transfer of the rescued cohort from one-shot to self-refine.

\subsection{Fix-shape classifier algorithm}
\label{app:rq4_classifier}

The fix-shape classifier introduced in Section~\ref{sec:rq4} labels each na\"ive R1-failure case by the shape of the edit that na\"ive's R2 (its repair attempt) actually attempted, computed mechanically from the line-level diff between R1 and R2. No human labels enter the classification. These diagnostic-line overlap classes do not determine whether the edited region is reachable by block/function rollback. The upstream-edit analysis in Appendix~\ref{app:rq3_localization} supplies that complementary motivation for scope widening, using final passing programs as an endpoint proxy.

For each na\"ive case where round 1 fails, let $R_1$ denote na\"ive's first-attempt code and $R_2$ its second-attempt code. Let $E$ denote the set of $R_1$ line numbers reported as errors in the diagnostic block fed back to round 2; for C to Rust the line numbers are extracted from \texttt{rustc}'s primary span markers (\texttt{- ->}~\texttt{program.rs:L:C}), and for JS to TS from \texttt{tsc}'s and ESLint's primary error markers (\texttt{- LL:C: error:}). Let $C$ denote the set of $R_1$ line numbers that $R_2$ modified, computed from \texttt{difflib.SequenceMatcher} opcodes: replace and delete chunks mark every $R_1$ line in the chunk; insert chunks mark the $R_1$ line immediately after the insertion point (or line~1 if the insertion is at the top). Let $n_{R_1}$ denote the number of $R_1$ lines.

The four-class label and a sub-reason for transparency are assigned by Algorithm~\ref{alg:fixshape}, with the first matching rule winning.

\begin{algorithm}[h]
\caption{Fix-shape classifier (mode A). Given the modified line set $C$, the error line set $E$, and the $R_1$ length $n_{R_1}$, label the case as LOCAL, NONLOCAL, MIXED, or UNKNOWN.}
\label{alg:fixshape}
\begin{algorithmic}[1]
\REQUIRE $n_{R_1}$, $E$, $C$
\IF{$n_{R_1} = 0$}
  \STATE \textbf{return} UNKNOWN \COMMENT{\textsc{extract\_fail}}
\ELSIF{$C = \emptyset$}
  \STATE \textbf{return} UNKNOWN \COMMENT{\textsc{no\_op}: $R_2$ produced no diff against $R_1$}
\ELSIF{$|C| / n_{R_1} > 0.5$}
  \STATE \textbf{return} UNKNOWN \COMMENT{\textsc{rewrite}: $R_2$ rewrote the majority of $R_1$}
\ELSIF{$E = \emptyset$}
  \STATE \textbf{return} UNKNOWN \COMMENT{\textsc{no\_error\_lines}: no extractable line numbers}
\ELSIF{$C \setminus E = \emptyset$}
  \STATE \textbf{return} LOCAL \COMMENT{every $R_2$ edit lands on an $R_1$ error line}
\ELSIF{$C \cap E = \emptyset$}
  \STATE \textbf{return} NONLOCAL \COMMENT{no $R_2$ edit lands on an error line}
\ELSE
  \STATE \textbf{return} MIXED \COMMENT{partial overlap}
\ENDIF
\end{algorithmic}
\end{algorithm}

The 0.5 rewrite threshold is fixed across both tasks and across model families. UNKNOWN is reported separately and excluded from hierarchy claims because the fix shape is not determined by the diff alone in those sub-cases; the headline ordering in Section~\ref{sec:rq4} (LOCAL $\geq$ MIXED $\geq$ NONLOCAL) is read from the three resolved classes.

The primary classification (mode A) uses $R_2$ as defined above. An alternative classification (mode B) uses $R_{\textrm{final}}$, the round of na\"ive's outer self-refine trace that ultimately compiled, in place of $R_2$, and is restricted to cases where na\"ive eventually passes; mode B is reported alongside mode A in Table~\ref{tab:rq4_fixshape}. The two modes agree on \rqFourCrustAgreementExact{} of mode-B-eligible cases on C to Rust (loose: \rqFourCrustAgreementLoose{}, n=\rqFourCrustAgreementEligibleN{}) and \rqFourJstsAgreementExact{} on JS to TS (loose: \rqFourJstsAgreementLoose{}, n=\rqFourJstsAgreementEligibleN{}); the loose agreement collapses LOCAL and MIXED into a single ANCHORED bucket.

\subsection{Fix-shape full breakdown}
\label{app:rq4_fixshape}

Table~\ref{tab:rq4_fixshape_full} reports the complete fix-shape rescue rate including the UNKNOWN bucket and the alternative R\textsubscript{final} classification (mode B), complementing the slim main-body view in Table~\ref{tab:rq4_fixshape}. The universe is na\"ive R1-failure cases (rescued plus both-fail; $n=\rqFourCrustOneshotUniverseN$ on C to Rust under mode A, $n=\rqFourJstsOneshotUniverseN$ on JS to TS under mode A). The UNKNOWN row sits at \rqFourCrustOptAUnknownRate{} on C to Rust ($n=\rqFourCrustOptAUnknownDenom$) and \rqFourJstsOptAUnknownRate{} on JS to TS ($n=\rqFourJstsOptAUnknownDenom$); we exclude UNKNOWN from the LOCAL $\geq$ MIXED $\geq$ NONLOCAL hierarchy claim because its constituent sub-reasons (no-op and rewrite) describe undetermined fix shapes rather than a coherent shape class.

\begin{table}[h]
  \centering
  \caption{Full fix-shape rescue rate on na\"ive R1-failure cases. Columns A use R\textsubscript{2} as the repair anchor (primary classification, matching the main-body Table~\ref{tab:rq4_fixshape}); columns B use R\textsubscript{final} from na\"ive-eventually-pass cases (alternative anchor). Cells report rescue rate (rate-eligible $n$); $*$ marks bins below the underpowered threshold of 5 cases.}
  \label{tab:rq4_fixshape_full}
  \small
  \rqFourFixshapeTableFull
\end{table}

The NONLOCAL bucket on JS to TS is underpowered. Mode A counts $\rqFourJstsOptANonlocalShare$ NONLOCAL cases out of $\rqFourJstsOneshotUniverseN$ R1-failure cases on JS to TS, with $\rqFourJstsOptANonlocalDenom$ cases entering the rescue-rate denominator; mode B counts $\rqFourJstsOptBNonlocalShare$ NONLOCAL cases ($n=\rqFourJstsOptBNonlocalDenom$). Below the underpowered threshold of 5 cases, the rescue-rate point estimate is reported with a $*$ marker in Table~\ref{tab:rq4_fixshape} and is not used to claim a NONLOCAL gap on JS to TS in Section~\ref{sec:rq4}; the LOCAL versus MIXED gap on JS to TS (\rqFourJstsOptALocalRate{} vs \rqFourJstsOptAMixedRate{}, mode A) is well-powered ($n=\rqFourJstsOptALocalDenom$ and $n=\rqFourJstsOptAMixedDenom$ respectively) and supports the LOCAL $\geq$ MIXED edge of the hierarchy on this task.

\subsection{Per-error-code rescue rate}
\label{app:rq4_percode}

The fix-shape hierarchy in Section~\ref{sec:rq4} categorizes cases by the shape of na\"ive's repair attempt; we additionally report per-error-code rescue rate as a complementary view organized by the diagnostic that broke na\"ive's R1 (Table~\ref{tab:rq4_percode}). For each error code that appears on at least 5 na\"ive R1-failure cases, we report the number of such cases, the number where \methodabbr{}/one-shot rescues, and the rescue rate. A case enters the denominator for every distinct error code that appears on its R1 diagnostic block.

\begin{table}[h]
  \centering
  \caption{Per-error-code rescue rate on na\"ive R1-failure cases, top 6 codes by R1-failure prevalence on each task. A case may enter the denominator for multiple codes when its R1 diagnostic block reports several. Codes with fewer than 5 R1-failure cases are excluded.}
  \label{tab:rq4_percode}
  \small
  \resizebox{\linewidth}{!}{\rqFourPercodeTable}
\end{table}

On C to Rust, the highest rescue rates concentrate on borrow-check and type-mismatch families (\texttt{E0599} method resolution at 54.4\%, \texttt{E0384} immutable-binding violations at 50.0\%) where the diagnostic anchors a concrete fix on the error line. The lower-rescue families (\texttt{E0425} unresolved name at 30.6\%, \texttt{E0433} unresolved trait at 28.6\%) typically require a non-local edit at an import or declaration upstream of the error line. On JS to TS, the dominant code is \texttt{@typescript-eslint/typedef} (the missing-type-annotation rule under our zero-typedef policy) at 19.5\%; the lower-rescue codes \texttt{TS2339}, \texttt{TS2554}, \texttt{TS1192}, \texttt{TS1259} are concentrated on cases where the correct annotation depends on cross-program usage rather than on a fact local to the variable declaration. The per-code ordering is consistent with the fix-shape hierarchy: codes whose canonical fix is local-cause and local-fix rescue at higher rates, and codes whose canonical fix lives upstream rescue at lower rates.

\subsection{Shared-code edit-concentration ratio}
\label{app:rq4_mechanism}

\begin{figure}[h]
  \centering
  \includegraphics[width=0.75\linewidth]{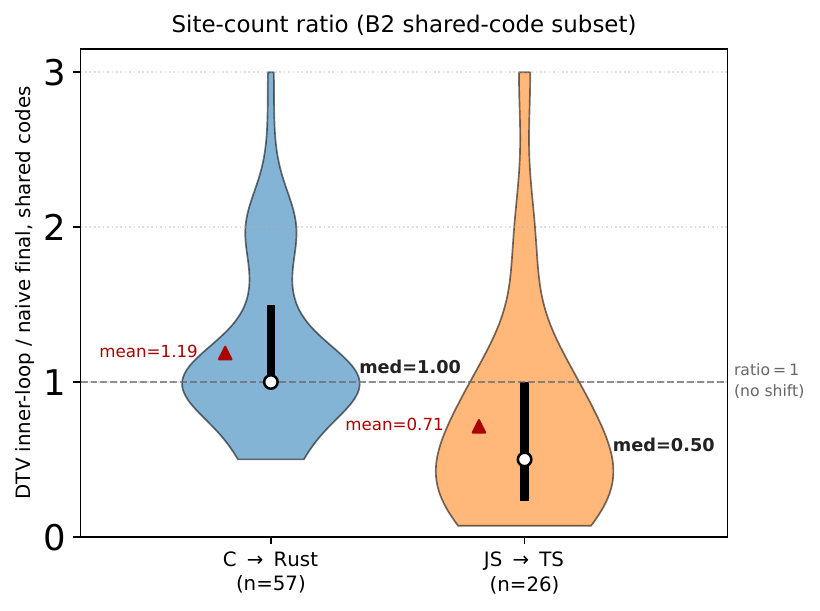}
  \caption{Per-case distribution of the shared-code edit-concentration ratio on rescued cases where \methodabbr{}'s inner verifier fired and the \methodabbr{}/na\"ive code sets intersect. $n=\rqFourCrustSharedRatioN$ on C to Rust and $n=\rqFourJstsSharedRatioN$ on JS to TS. Ratios at or below 1 indicate \methodabbr{} reaches rescue with at most as many per-shared-code verify-and-rollback cycles as na\"ive accumulated as R1 errors on the same codes.}
  \label{fig:rq4_mechanism}
\end{figure}

Figure~\ref{fig:rq4_mechanism} reports the per-case distribution of the shared-code edit-concentration ratio used in the Section~\ref{sec:rq4} mechanism analysis on rescued cases. The universe is the subset of rescued cases where \methodabbr{}'s inner verifier fired and structure-aware rollback intervened (\methodabbr{}/one-shot passes after at least one inner-loop rollback), intersected with the shared-code subset (\methodabbr{} inner-loop error codes and na\"ive R1 error codes have non-empty intersection). For each case in this universe, the ratio is computed as the number of distinct \methodabbr{} inner-loop diagnostic emissions on the shared codes divided by the number of distinct na\"ive R1 diagnostic emissions on the same codes; a ratio below 1 indicates \methodabbr{} emits fewer per-shared-code verify diagnostics than na\"ive accumulated as R1 errors on the same codes.

The C-to-Rust distribution is centered at parity (median \rqFourCrustSharedRatioMedian{}, with first and third quartiles $Q_1$=\rqFourCrustSharedRatioQOne{} and $Q_3$=\rqFourCrustSharedRatioQThree{}, mean \rqFourCrustSharedRatioMean{}); the JS-to-TS distribution is shifted left of parity (median \rqFourJstsSharedRatioMedian{}, $Q_1$=\rqFourJstsSharedRatioQOne{}, $Q_3$=\rqFourJstsSharedRatioQThree{}, mean \rqFourJstsSharedRatioMean{}). Both distributions sit at or below 1 at the median, supporting the Section~\ref{sec:rq4} claim that \methodabbr{}'s incremental work is concentrated on the same error codes na\"ive's R1 hit rather than redirected to a different error class. The complementary both-fail comparison reinforces this view: \rqFourCrustBothFailSameCodePct{} of C-to-Rust both-fail cases ($n=\rqFourCrustBothFailSameCodeN$ of $\rqFourCrustBothFailN$) and \rqFourJstsBothFailSameCodePct{} of JS-to-TS both-fail cases ($n=\rqFourJstsBothFailSameCodeN$ of $\rqFourJstsBothFailN$) have \methodabbr{}'s inner-loop diagnostics intersecting na\"ive's R1 diagnostics, indicating that when \methodabbr{} does not pass, it is most often stuck on the same error codes that broke na\"ive.

\subsection{Paired round-of-pass statistics (self-refine)}
\label{app:rq4_paired_rounds}

Table~\ref{tab:rq4_paired_rounds} reports the paired round-of-pass statistics summarized in the Section~\ref{sec:rq4} round-savings claim. The both-pass subset includes only cases where both na\"ive/self-refine and \methodabbr{}/self-refine reach pass on the matched case identifiers. The med ratio and mean ratio columns are over the per-case na\"ive\_rounds / \methodabbr{}\_rounds; the e/s/l column reports the count of cases where \methodabbr{} reaches pass earlier, in the same outer round, or later than na\"ive; $\Delta$ is the per-case difference (na\"ive $-$ \methodabbr{}) in outer rounds; sign $p$ uses the two-sided exact sign test on the discordant pairs. The right-shifted ratio distributions ($Q_1$=\rqFourCrustSrPairedRatioQOne, $Q_3$=\rqFourCrustSrPairedRatioQThree on C to Rust; $Q_1$=\rqFourJstsSrPairedRatioQOne, $Q_3$=\rqFourJstsSrPairedRatioQThree on JS to TS) confirm the round-savings effect is broad rather than driven by tail outliers; both sign tests reject equality and both bootstrap CIs on the mean ratio exclude 1.0 (Table~\ref{tab:rq4_paired_rounds}).

\begin{table}[h]
  \centering
  \caption{Paired round-count statistics on the both-pass subset (matched na\"ive/self-refine and \methodabbr{}/self-refine, same case identifiers). The e/s/l column counts cases where \methodabbr{} passes in an earlier, the same, or a later outer round than na\"ive.}
  \label{tab:rq4_paired_rounds}
  \small
  \resizebox{\linewidth}{!}{\rqFourPairedRoundsTable}
\end{table}

\subsection{Both-pass token-cost analysis (self-refine)}
\label{app:rq4_bothpass_tokens}

The compile-pass paired analysis in Appendix~\ref{app:rq1_pairs} establishes that \methodabbr{} improves the binary outcome (more cases compile). This subsection addresses the orthogonal question used in the Section~\ref{sec:rq4} round-savings analysis: on the cases where both methods succeed, does \methodabbr{} use more or fewer tokens? Restricted to the both-pass subset of the matched na\"ive/self-refine versus \methodabbr{}/self-refine comparison, Table~\ref{tab:rq4_bothpass_tokens} reports the per-case token-difference distribution ($\Delta$tok = \methodabbr{} $-$ na\"ive in tokens; negative favors \methodabbr{}).

\begin{table}[h]
  \centering
  \caption{Per-case token-difference distribution on the both-pass subsets of the matched na\"ive/self-refine versus \methodabbr{}/self-refine comparison ($\Delta$tok = \methodabbr{} tokens $-$ na\"ive tokens; negative favors \methodabbr{}). The \methodabbr{} $<$ / $=$ / $>$ na\"ive columns report the number of cases where \methodabbr{} uses fewer, equal, or more tokens (the equal column captures cases where both methods pass on first attempt with identical token counts). The trimmed mean drops the top 10\% and bottom 10\% of cases (the largest \methodabbr{}-better and \methodabbr{}-worse outliers symmetrically) to expose tail-driven shifts in the arithmetic mean.}
  \label{tab:rq4_bothpass_tokens}
  \small
  \resizebox{\linewidth}{!}{\rqOneBothPassTokenTable}
\end{table}

\paragraph{Both-pass medians favor \methodabbr{} on both tasks.}
On C to Rust ($n=\rqOneCrustNaiveSrVsDtvSrBpN$), \methodabbr{} uses fewer tokens than na\"ive on \rqOneCrustNaiveSrVsDtvSrBpDtvBetter{} of the both-pass cases (\rqOneCrustNaiveSrVsDtvSrBpDtvBetterPct{}), na\"ive uses fewer on \rqOneCrustNaiveSrVsDtvSrBpNaiveBetter{} (\rqOneCrustNaiveSrVsDtvSrBpNaiveBetterPct{}), and the median paired difference is \rqOneCrustNaiveSrVsDtvSrBpMedianTok{} tokens (sign test $p\rqOneCrustNaiveSrVsDtvSrBpSignP$). On JS to TS ($n=\rqOneJstsNaiveSrVsDtvSrBpN$), \methodabbr{} is the winner on \rqOneJstsNaiveSrVsDtvSrBpDtvBetter{} cases (\rqOneJstsNaiveSrVsDtvSrBpDtvBetterPct{}) versus \rqOneJstsNaiveSrVsDtvSrBpNaiveBetter{} for na\"ive (\rqOneJstsNaiveSrVsDtvSrBpNaiveBetterPct{}), with a median paired difference of \rqOneJstsNaiveSrVsDtvSrBpMedianTok{} tokens (sign test $p\rqOneJstsNaiveSrVsDtvSrBpSignP$, underpowered at this sample size). Both medians point in the same direction: on the typical case where both methods pass, \methodabbr{} uses fewer tokens.

\paragraph{The mean disagrees with the median on JS to TS, driven by an asymmetric tail.}
On C to Rust the arithmetic mean (\rqOneCrustNaiveSrVsDtvSrBpMeanTok{} tokens) agrees with the median direction; on JS to TS the mean is \rqOneJstsNaiveSrVsDtvSrBpMeanTok{} tokens, opposite of the median. Inspecting the tails explains the discrepancy. Among the JS-to-TS both-pass cases, the top \rqOneJstsNaiveSrVsDtvSrBpTailTopN{} (10\%) where \methodabbr{} uses most extra tokens contribute a cumulative \rqOneJstsNaiveSrVsDtvSrBpTailDtvBadSum{} tokens, while the bottom \rqOneJstsNaiveSrVsDtvSrBpTailTopN{} where \methodabbr{} saves most contribute only \rqOneJstsNaiveSrVsDtvSrBpTailDtvGoodSum{} tokens (a \rqOneJstsNaiveSrVsDtvSrBpTailRatio{}$\times$ asymmetry favoring the \methodabbr{}-bad tail). After symmetrically trimming \rqOneJstsNaiveSrVsDtvSrBpTailTopN{} cases from each tail, the mean over the remaining \rqOneJstsNaiveSrVsDtvSrBpTrimN{} cases swings to \rqOneJstsNaiveSrVsDtvSrBpTrimMeanTok{} tokens, recovering the median direction. The C-to-Rust both-pass distribution has the opposite tail shape: top-10\% \methodabbr{}-good cases cumulatively save \rqOneCrustNaiveSrVsDtvSrBpTailDtvGoodSum{} tokens versus top-10\% \methodabbr{}-bad cases costing \rqOneCrustNaiveSrVsDtvSrBpTailDtvBadSum{} tokens (\rqOneCrustNaiveSrVsDtvSrBpTailRatio{}$\times$ in the opposite direction), so the C-to-Rust mean is robust to trimming and stays negative (\rqOneCrustNaiveSrVsDtvSrBpTrimMeanTok{} tokens after the same symmetric 10\%/10\% trim).

\paragraph{Mechanism behind the JS-to-TS \methodabbr{}-bad tail.}
The structure-aware rollback mechanism underlying this asymmetric tail is described in Section~\ref{sec:rq4}: it ties back to the same non-local fix-shape boundary that lowers \methodabbr{}'s per-case rescue rate.

\section{Diagnostic guidance and repair scope}
\label{app:repair_guidance}

We examine how to guide regeneration after a failure and how much earlier code to make editable. We use ROCODE~\citep{jiang2025rocode} as a representative implementation of the mechanism-level baselines: failed-path suppression for regeneration and entropy-based statement selection for rollback. The analyses test whether diagnostic feedback improves local repair success and whether an uncertainty-selected statement exposes observed upstream edits.

\subsection{Matched regeneration-signal comparison}
\label{app:rq3_regeneration}

We isolate the regeneration signal by replaying the \regenerationSignalCaseCount{} initial compiler failures among the first 100 C-to-Rust cases evaluated with Qwen3-4B. For each failure, both arms resume from the same statement checkpoint with identical source context, sampling settings, and generation budgets, and are checked at the next structural boundary.

The diagnostic arm uses inline continuation with the \texttt{rustc} diagnostic appended as a block comment, supplying information about the constraint to repair. The path-decay arm discourages failed continuations without supplying diagnostic text. It reuses ROCODE's released trie and position-dependent decay~\citep{jiang2025rocode} with factor 0.9, followed by its top-$k$/top-$p$ sampling logic.\footnote{ROCODE implementation: \url{https://github.com/jiangxxxue/ROCODE} (MIT license).}

The outcome requires reaching the shared boundary, producing a renderable artifact, and passing the same in-loop \texttt{rustc} oracle. All pairs reach the boundary; an unavailable rendered artifact counts as failure. Table~\ref{tab:matched_regeneration} reports local one-repair pass rates, and Table~\ref{tab:matched_regeneration_pairs} reports the paired outcomes. Diagnostic conditioning improves pass rate by \regenerationSignalDiff{}~pp (exact two-sided McNemar $p=\regenerationSignalMcnemarP$; paired-bootstrap 95\% CI $[\regenerationSignalCiLo,\regenerationSignalCiHi]$~pp, 20,000 resamples, seed 20260726). This outcome measures local verifier admissibility, not complete-program correctness.

\begin{table}[ht]
  \centering
  \caption{One-repair verifier pass under a shared failure state, statement checkpoint, renderer, and oracle.}
  \label{tab:matched_regeneration}
  \small
  \regenerationSignalPassTable
\end{table}

\begin{table}[ht]
  \centering
  \caption{Paired outcomes for the matched regeneration comparison ($n=\regenerationSignalCaseCount$).}
  \label{tab:matched_regeneration_pairs}
  \small
  \regenerationSignalPairsTable
\end{table}

\subsection{Observed upstream edits and entropy localization}
\label{app:rq3_localization}

We examine the \rollbackLocalizationCompilingCaseCount{} cases in the initial-failure cohort whose full \methodabbr{} trajectory eventually compiles. For every failure-time artifact, we compare its generated-code tokens with the final passing program using a Tree-sitter terminal-token diff that excludes comments and formatting. A case is positive if at least one failure contains an observed edited line $L_e$ before the first error-level diagnostic's primary line $L_d$. This identifies \rollbackLocalizationUpstreamCount/\rollbackLocalizationCompilingCaseCount{} cases (\rollbackLocalizationUpstreamPct{}). As in the fix-shape analysis of Appendix~\ref{app:rq4_classifier}, the successful-program diff is a proxy for repair location, not evidence that each edit is necessary or unique. The aggregate and its classifications were manually checked.

For each positive case, we select its earliest qualifying failure, yielding one observation per case (44 inner failures and one outer failure). We teacher-force the artifact under the original source-translation prompt without diagnostic feedback. Following ROCODE's ordering, base-2 full-vocabulary entropy is computed after temperature scaling and before path decay or top-$k$/top-$p$ filtering. We select the global maximum-entropy generated token and its containing statement. Let $L_H$ be the selected token's line: exact-line hits satisfy $L_H=L_e$, while a rollback \emph{exposes} the observed edit if the selected statement starts at or before $L_e$.

Table~\ref{tab:entropy_localization} reports the all-token analysis and the variant excluding comments and whitespace. Both locate the exact edit line in \rollbackLocalizationEntropyExact/\rollbackLocalizationUpstreamCount{} cases, expose the edit in \rollbackLocalizationEntropyExposes/\rollbackLocalizationUpstreamCount{}, and leave it frozen in \rollbackLocalizationEntropyLate/\rollbackLocalizationUpstreamCount{}; median signed distance is \rollbackLocalizationEntropyDistance{} lines. The selected cohort tests access to observed upstream edits; it does not evaluate complete repair trajectories.

\begin{table}[H]
  \centering
  \caption{Entropy localization at the earliest qualifying failure of each observed-upstream-edit case ($n=\rollbackLocalizationUpstreamCount$). Distance is in generated-code lines; ``Code only'' excludes comments and whitespace. ``Too late'' means the selected statement starts after the observed edit line, so rollback leaves that edit frozen. ``Exact line'' cases are included in ``Exposes edit''.}
  \label{tab:entropy_localization}
  \small
  \rollbackLocalizationEntropyTable
\end{table}

\paragraph{Takeaway.}
Diagnostic-conditioned regeneration improves local repair success over failed-path suppression, while structural escalation can reopen earlier code left inaccessible by a too-late statement checkpoint. Together, these mechanisms give \methodabbr{} more directed regeneration and greater flexibility to repair earlier decisions.

\end{document}